\documentclass[11pt]{article}
\usepackage[utf8]{inputenc}
\usepackage[english]{babel}
\usepackage{authblk}
\usepackage{enumerate}
\usepackage{color}
\usepackage{graphicx}
\usepackage{amssymb, amsmath, amsthm}
\usepackage{amsfonts}
\usepackage{bm}
\usepackage{calc}
\usepackage{caption}
\usepackage{subcaption}
\usepackage{url}
\usepackage{epstopdf}
\usepackage{float}

\usepackage{geometry}
\usepackage[title]{appendix}
\usepackage{cite}
 
\numberwithin{equation}{section}

\theoremstyle{remark}
\newtheorem{remark}{Remark}[section]

\theoremstyle{definition}

\theoremstyle{definition}
\newtheorem*{definition*}{Definition}

\DeclareMathOperator*{\totder}{\mathrm{d}}

\author{Jens Berg\thanks{jens.berg@math.uu.se}\ }
\author{Kaj Nystr\"{o}m\thanks{kaj.nystrom@math.uu.se}} 
\affil{Department of Mathematics, Uppsala University\\
S-751 06 Uppsala, Sweden}

\title{Neural network augmented inverse problems for PDEs}
\date{}

\begin{document}
\maketitle

\begin{abstract}
In this paper we show how to augment classical methods for inverse problems with artificial neural networks. The neural network acts as a prior for the coefficient to be estimated from noisy data. Neural networks are global, smooth function approximators and as such they do not require explicit regularization of the error functional to recover smooth solutions and coefficients. We give detailed examples using the Poisson equation in 1, 2, and 3 space dimensions and show that the neural network augmentation is robust with respect to noisy and incomplete data, mesh, and geometry.
\end{abstract}

\setcounter{equation}{0} \setcounter{theorem}{0}

\section{Introduction}
In this paper we study the classical coefficient approximation problem for partial differential equations (PDEs). The inverse problem consists of determining the coefficient(s) of a PDE given more or less noisy measurements of its solution. A typical example is the heat distribution in a material with unknown thermal conductivity. Given measurements of the temperature at certain locations, we are to estimate the thermal conductivity of the material by solving the inverse problem for the stationary heat equation.

The problem is of both practical and theoretical interest. From a practical point of view, the governing equation for some physical process is often known, but the material, electrical, or other properties are not. From a theoretical point of view, the inverse problem is a challenging often ill-posed problem in the sense of Hadamard. Inverse problems have been studied for a long time, starting with \cite{levenberg} in the 40's, \cite{arquardt} and \cite{drumshape} in the 60's, to being popularized by \cite{tikhonov} in the 70's by their work on regularization techniques. Today there is a vast amount of literature available on inverse, and PDE constrained optimization problems in general, and we refer the reader to the many survey papers and textbooks, and the references therein, for an overview. See for example \cite{inversesurvey, stepguide, pdeoptbook, coeffinversebook}.

As the PDE cannot be solved analytically in general, a numerical method is required. A common choice is the finite element method (FEM) which have been frequently used with success for a large class of inverse problems. See for example \cite{feminverseperkkele, feminverseelliptic, feminversetrap, feminversemodel}. Recently, there has been a lot of activity in a branch of numerical analysis known as probabilistic numerics, with successful applications in inverse problems and PDE constrained optimization in general. The probabilistic numerical methods are similar, yet different, to the approach taken in this paper. In the probabilistic setting, one places a statistical prior, typically a Gaussian prior, on the unknown coefficient and uses Bayesian inference to compute statistics of the unknown coefficient. As more data is used in the inference, the variance of the coefficient is reduced and one ends up with a confidence interval for the true coefficient. See for example \cite{2017arXiv170203673C, HenOsbGirRSPA2015, StochasticNewton, hennig13_quasi_newton_method, gausspde1, gausspde2, 2016arXiv160507811C, 2014arXiv14071517B}.

In this work, we use the finite element method as supplied by the software package \texttt{FEniCS} \cite{AlnaesBlechta2015a, LoggMardalEtAl2012a} to solve the underlying PDE. To compute the gradient we need for the PDE constrained optimization, we use the software \texttt{dolfin-adjoint} \cite{dolfinadjoint} which works together with \texttt{FEniCS} to automatically derive and solve the adjoint PDE. Finally, we augment \texttt{FEniCS} and \texttt{dolfin-adjoint} with an artificial neural network to represent the unknown coefficient in the PDE. The neural networks we consider in this paper are simple feedforward neural networks with sigmoid activation functions in the hidden layers, and linear activations in the output layer. Such a neural network defines a smooth mapping $\mathbb{R}^N \to \mathbb{R}$ which can approximate any continuous function to arbitrary accuracy \cite{Hornik1990551}. The similarity with the probabilistic method lies in that a model for the coefficient is selected a priori, and as more data is analyzed, the model improves. The probabilistic method uses a statistical non-parametric model, and here we use neural networks which is a deterministic parametric prior.

The model problem we are working with in this paper is the scalar Poisson equation in a subset $\Omega \subset \mathbb{R}^N$, $N = 1, 2, 3$, given by
\begin{equation}
-\nabla \cdot (q(x) \nabla u(x)) = f(x),
\end{equation}
subject to Dirichlet boundary conditions. The inverse problem is to compute the coefficient $q$, given a series of more or less noisy measurements $\hat{u}_i$ of $u_i$, for $i = 0, \ldots, M$. It is clear that the inverse problem is ill-posed whenever $\nabla u = 0$ on an open set, as then there is no information available about the coefficient $q$. Moreover, well-posedness of the inverse problem depends on the norm under consideration. For example, consider the Poisson equation in one space dimension, $x \in (0, \pi)$, $q = 1/2$, $u = x^2$, and a sequence of coefficients and observations $\hat{q}_N = (2 + \cos(Nx))^{-1}$, $\hat{u}_N = u + \delta_N$ with
\begin{equation}
\delta_N = \frac{x}{N} \sin(Nx) + \frac{1}{N^2} \cos(Nx).
\end{equation}
In the $L^{\infty}$-norm given by
\begin{equation}
||u||_{\infty} = \max_{x \in \Omega} |u(x)|
\end{equation}
we have
\begin{equation}
\begin{aligned}
||\hat{u}_N - u||_{\infty} \leq \frac{c}{N} \to 0, && ||\hat{q}_N - q||_{\infty} = \frac{1}{2} \not\to 0,
\end{aligned}
\end{equation}
and we can clearly see that $q$ is not a continuous function of the data in the sense of Hadamard \cite{Kohn1988}. One can, however, show that this problem is well-posed in the Sobolev norm $H^1$. However, as PDE constrained optimization in the $H^1$-norm is in general not practically applicable, since one does not have measurements of $\nabla \hat{u}$ in general, we will consistently use the standard $L^2$-norm.

\section{The general inverse problem}
We consider stationary partial differential equations of the general form
\begin{equation}
\begin{aligned}
F(u, q) &= 0, && x \in \Omega, \\
Bu &= g, && x \in \partial \Omega,
\label{generalpde}
\end{aligned}
\end{equation}
where $\Omega \subset \mathbb{R}^N$ is a bounded domain and $\partial \Omega$ its boundary. Moreover, $u = u(x)$ is the solution variable, $B$ is a boundary operator, $g$ is the boundary data, and $q = q(x)$ is an a priori unknown parameter.

The deviation between $u$ and $\hat{u}$ is defined by the error functional
\begin{equation}
J(u, q) = \frac{1}{2}||u - \hat{u}||^2 = \frac{1}{2}\int|u - \hat{u}|^2dx,
\label{errfunc}
\end{equation}
and the goal is to compute $q^*$ such that
\begin{equation}
q^* = \min_q J(u, q)
\label{qmin}
\end{equation}
subject to $u$ satisfying \eqref{generalpde}. To efficiently solve the minimization problem \eqref{qmin}, we require the gradient of the error functional \eqref{errfunc} with respect to $q$. To compute the gradients, we will use the adjoint approach.

\subsection{The adjoint equations}
Note that the parameter $q$ defines the mapping
\begin{equation}
q \mapsto u(q),
\label{qmap}
\end{equation}
which can be computed by solving \eqref{generalpde} by analytical or numerical means. This turns \eqref{generalpde}, and \eqref{errfunc} into pure functions of $q$,
\begin{equation}
\begin{aligned}
F(u, q) &\mapsto F(u(q), q), \\
J(u, q) &\mapsto J(u(q), q),
\end{aligned}
\end{equation}
and the PDE constrained problem \eqref{qmin} becomes an unconstrained problem as the constraint is built-in to the solution of \eqref{generalpde}. The total derivative of the error functional can be computed as
\begin{equation}
\frac{\totder J(u(q), q)}{\totder q} = \frac{\partial J(u(q), q)}{\partial u(q)} \frac{\partial u(q)}{\partial q} + \frac{\partial J(u(q), q)}{\partial q},
\label{jgrad}
\end{equation}
where $\partial J(u(q), q)/\partial u(q)$ and $\partial J(u(q), q)/\partial q$ are in general easily computed. To compute the troublesome solution Jacobian $\partial u(q)/\partial q$, we formally differentiate the PDE itself to obtain
\begin{equation}
\frac{\totder F(u(q), q)}{\totder q} = \frac{\partial F(u(q), q)}{\partial u(q)} \frac{\partial u(q)}{\partial q} + \frac{\partial F(u(q), q)}{\partial q},
\end{equation}
or equivalently
\begin{equation}
-\frac{\partial F(u(q), q)}{\partial u(q)} \frac{\partial u(q)}{\partial q} = \frac{\partial F(u(q), q)}{\partial q}.
\label{ftangent}
\end{equation}
The relation \eqref{ftangent} is known as the tangent linear system related to the functional \eqref{errfunc}. In the rest of the derivations, we will write $F(u(q), q) = F$, $J(u(q), q) = J$ and $u(q) = u$ to ease the notation. The tangent linear operator $\partial F/\partial u$ is the linearization of the differential operator around the solution $u$ which acts on the solution Jacobian $\partial u/\partial q$. In the case of a linear PDE, the differential operator in \eqref{generalpde} and the tangent linear operator coincides. Assuming that the tangent linear system is invertible, we can use \eqref{generalpde} and \eqref{ftangent} to solve
\begin{equation}
\frac{\partial u}{\partial q} = -\left( \frac{\partial F}{\partial u} \right)^{-1} \frac{\partial F}{\partial q}.
\label{dudq}
\end{equation}
Substituting \eqref{dudq} into \eqref{jgrad} and taking the Hermitian transpose gives
\begin{equation}
\frac{\totder J^*}{\totder q} = -\frac{\partial F^*}{\partial q} \left( \frac{\partial F^*}{\partial u} \right)^{-1} \frac{\partial J^*}{\partial u} + \frac{\partial J^*}{\partial q}.
\end{equation}
We define the \textit{adjoint} variable $\lambda$ by
\begin{equation}
\lambda = -\left( \frac{\partial F^*}{\partial u} \right)^{-1} \frac{\partial J^*}{\partial u},
\end{equation}
or equivalently
\begin{equation}
-\frac{\partial F^*}{\partial u} \lambda = \frac{\partial J^*}{\partial u}.
\label{adjoint}
\end{equation}
The above equation \eqref{adjoint} is the \textit{adjoint}, or \textit{dual}, equation associated with the \textit{forward} problem \eqref{generalpde} and the error functional \eqref{errfunc}. By solving the adjoint equation \eqref{adjoint} and substituting into \eqref{jgrad}, we can compute the total derivative of the error functional as
\begin{equation}
\frac{\totder J}{\totder q} = \lambda^* \frac{\partial F}{\partial q} + \frac{\partial J}{\partial q}.
\label{jgradadjoint}
\end{equation}
With the total derivative of the error functional given by \eqref{jgradadjoint}, any gradient based optimization method can be used to solve the minimization problem \eqref{qmin}.

\section{The numerical method}
In general, the PDE \eqref{generalpde} cannot be solved analytically and hence neither the mapping \eqref{qmap} nor the gradient \eqref{jgradadjoint} can be computed. In practice, a numerical method is used to solve both the PDE \eqref{generalpde} and the adjoint equation \eqref{adjoint}. A common method is to use FEM. To effectively use FEM, one selects a finite element space $V$ for the solution, and another space $Q$ for the coefficient. Typically, $Q$ is the space of piecewise constants to reduce the number of degrees of freedom. As the space $V$ consist of local basis functions, one needs regularization of the error functional to reconstruct a smooth coefficient. Typically, one adds Tikhonov regularization to the error functional of the form
\begin{equation}
\tilde{J} = \frac{1}{2} \int |u - \hat{u}|^2 dx + \frac{\alpha}{2} \int |q|^2 dx,
\end{equation}
where $\alpha > 0$ is a regularization parameter. If, for some reason, an estimate of the coefficient and noise level is known a priori, one can further improve the results by adding generalized Tikhonov regularization of the form
\begin{equation}
\tilde{J} = \frac{1}{2} \int |u - \hat{u}|^2 dx + \frac{\alpha^{\delta}}{2} \int |q - q_*|^2 dx,
\label{genregfunc}
\end{equation}
where $q_*$ is the a priori estimated coefficient and the regularization parameter $\alpha^{\delta}$ depends on the noise level. The actual value of regularization parameters $\alpha$ and $\alpha^{\delta}$ depends on the problem at hand, and finding an optimal value is a non-trivial task. The optimal generalized Tikhonov regularization parameter $\alpha^{\delta}$ can be computed by using, for example, the Morozov discrepancy principle or heuristic $L$-curves. These methods are, however, of limited practical use in a PDE constrained optimization context as they become extremely expensive. We will, however, discuss optimal regularization of the 1D example in appendix \ref{optreg}.

In this section, we do not use any regularization in the error functional \eqref{errfunc}. We will use only the most basic settings and avoid any fine tuning of hyperparameters, or any other parameters, to get as much of a black box solution strategy as possible. This means that the comparisons we make with pure FEM are not entirely fair as finding the optimal function spaces and regularizations are beyond the scope of this paper. The main purpose of this paper is to present neural networks as possible augmentations to classical methods for inverse problems. Adding explicit regularization will improve the neural network augmentation as well as can be seen in appendix~\ref{optreg}.

\subsection{The finite element method and automatic differentiation}
In this work, we have used the FEM software \texttt{FEniCS} to solve the PDE \eqref{generalpde}, together with \texttt{dolfin-adjoint} to automatically derive and solve the adjoint equation \eqref{adjoint}. The automatic differentiation in \texttt{dolfin-adjoint} works by overloading the solver functions in \texttt{FEniCS} and recording a graph of solver operations. Each node in the graph has an associated gradient operation and the total derivative can be computed by reversed mode differentiation, i.e. backpropagation. The reversed mode differentiation has the benefit that the gradient computation is exact with respect to the FEM discretization, as long as it is smooth enough. The only source of error in the computation of \eqref{jgradadjoint} is thus in the numerical solution of the forward problem\footnote{At least as long as the adjoint equation can be solved by a direct method. For large enough systems, an iterative method must be used which introduces additional errors. These are, however, usually small compared to the discretization errors of the forward problem.}.

\subsection{Neural network representation of the coefficient}
Rather than representing the unknown coefficient in a finite element space, we can represent the coefficient by a feedforward artificial neural network. That is, we let $q = q(x; W, b)$ be parameterized by the weights and biases of a neural network, here denoted by $W$ and $b$. This approach yield some immediate benefits. A neural networks is a global, smooth, function approximator which can approximate any continuous function to arbitrary accuracy by having a sufficient amount of parameters \cite{Hornik1990551, Li1996327}. Since it is a global function, it can be cheaply evaluated anywhere in the domain without first searching for the correct mesh element and performing interpolation, which is an expensive operation.

The coefficient $q$ is computed by feed forwarding
\begin{equation}
\begin{aligned}
q(x) &= \sigma(z^L) \\
z^L &= W^L\sigma_{L-1}(z^{L-1}) + b^L \\
z^{L-1} &= W^{L-1}\sigma_{L-2}(z^{L-2}) + b^{L-1} \\
&\mathrel{\makebox[\widthof{=}]{\vdots}} \\
z^2 &= W^2\sigma_1(z^1) + b^2 \\
z^1 &= W^1x + b^1,
\end{aligned}
\end{equation}
where $W^l_{ij} = w^l_{ij}$ are the weight matrices connecting layers $l$ and $l-1$ and $b^l$ are the vectors of biases for each layer, and $\sigma_l$ is the activation function of layer $l$. To determine weights and biases, we seek $W^*$, $b^*$ such that
\begin{equation}
W^*, b^* = \min_{W, b} J.
\label{qminnet}
\end{equation}
As there is one extra layer of parameters, we need to apply the chain rule once more to compute the total derivative of the error functional. Let $p$ denote any of the weight or biases parameters, then
\begin{equation}
\frac{\totder J}{\totder p} = \frac{\partial J}{\partial u} \frac{\partial u}{\partial q} \frac{\partial q}{\partial p} + \frac{\partial J}{\partial q} \frac{\partial q}{\partial p} = \left( \frac{\partial J}{\partial u} \frac{\partial u}{\partial q} + \frac{\partial J}{\partial q} \right) \frac{\partial q}{\partial p}.
\label{jgradnet}
\end{equation}
The first factor on the right hand side of \eqref{jgradnet} is computed as before by using \texttt{FEniCS} and \texttt{dolfin-adjoint}. The last factor is the gradient of the network output with respect to the network parameters, which can be computed exactly by using backpropagation or automatic differentiation.

Note that the neural network augmentation is not restricted to FEM. Any numerical method which efficiently can solve the forward problem \eqref{generalpde}, and the adjoint problem \eqref{adjoint}, can be used in combination. The neural network is simply a prior for the coefficient with good function approximation properties, and efficient means for computing its gradient. The application of the chain rule in \eqref{jgradnet} factors the discretization of the forward and backward problems from the computation of the gradient of the network. As a side effect, the discretization of the PDE can be refined without increasing the number of parameters in the neural network. In the simplest FEM case, the coefficient is represented as a constant on each mesh element. The number of optimization parameters is then equal to the number of mesh elements, which quickly become infeasible if the mesh is refined.

In the numerical examples which follow, we have consistently used a neural network design with as few parameters as possible. A low number of parameters is beneficial both for the numerical optimization as well as for the the implicit regularization. The designs are still, however, based on heuristics and experience. We found during this work that surprisingly small networks perform better than larger networks as long as the number of parameters is sufficient to approximate the coefficient. Once a sufficient design has been found, the results start to deteriorate with increasing network complexity due to overfitting and less efficient implicit regularization.

\section{Moderately ill-posed examples} \label{moderately}
In the numerical examples, we have used an exact solution and coefficient to compute a forcing function in \eqref{generalpde}. The exact solution is used to generate the data in the error functional. In the case of added noise, we add a normally distributed value $\delta r$, $r \in \mathcal{N}(0, 1)$, for some noise level $\delta$, to each of the interior data points. The data on the boundary is always exact.

The results differ somewhat depending on the noise and random initialization of the network's weights. In the figures and tables, we use \texttt{Numpy}'s random number generators with \texttt{numpy.random.seed(2)}, for reproducibility, to add noise and initialize the weights. We use the same initial guess for the coefficient and settings for the optimizer in both the FEM and network cases.

The equation we use in the examples is the Poisson equation. The equation requires that the coefficient to be estimated is positive. To ensure that the coefficient is initially positive, we initialize the network's weights from a uniform distribution $\mathcal{U}[0, 1)$ and all the biases are initialized to zero. Note that the endpoint is excluded by default in \texttt{numpy}'s random number generators. It is of no practical importance.

The examples which follow are called moderately ill-posed since the coefficients to be estimated are smooth and the measurement data is available in the whole domain. The purpose of the examples is to show the applicability and ease of use of the method due to the implicit regularization by the neural network prior.

\subsection{One-dimensional Poisson equation} \label{secheat1d}
The one-dimensional Poisson equation with Dirichlet boundary conditions is given by
\begin{equation}
\begin{aligned}
-(qu_x)_x &= f, && x \in (0, 1), \\
u(0) &= g_0, && u(1) = g_1,
\label{heat1d}
\end{aligned}
\end{equation}
where $f$, $g_{0,1}$ are known, and $q = q(x) : \mathbb{R} \to \mathbb{R}_+$ is the coefficient to be estimated. We discretize the domain $\Omega$ into $M=101$ continuous piecewise linear elements where the solutions are represented. We solve the minimization problem using both standard FEM, with the coefficient represented in the solution space, as well as represented by a neural network. Here, we use the BFGS \cite{bfgs} optimizer from \texttt{SciPy} \cite{scipy} as the number of optimization parameters is rather small. We iterate until the norm of the gradient of the error functional is less than $10^{-6}$.

\begin{figure}[htp]
\centering
\begin{subfigure}[t]{0.45\textwidth}
\centering
\includegraphics[width=\textwidth]{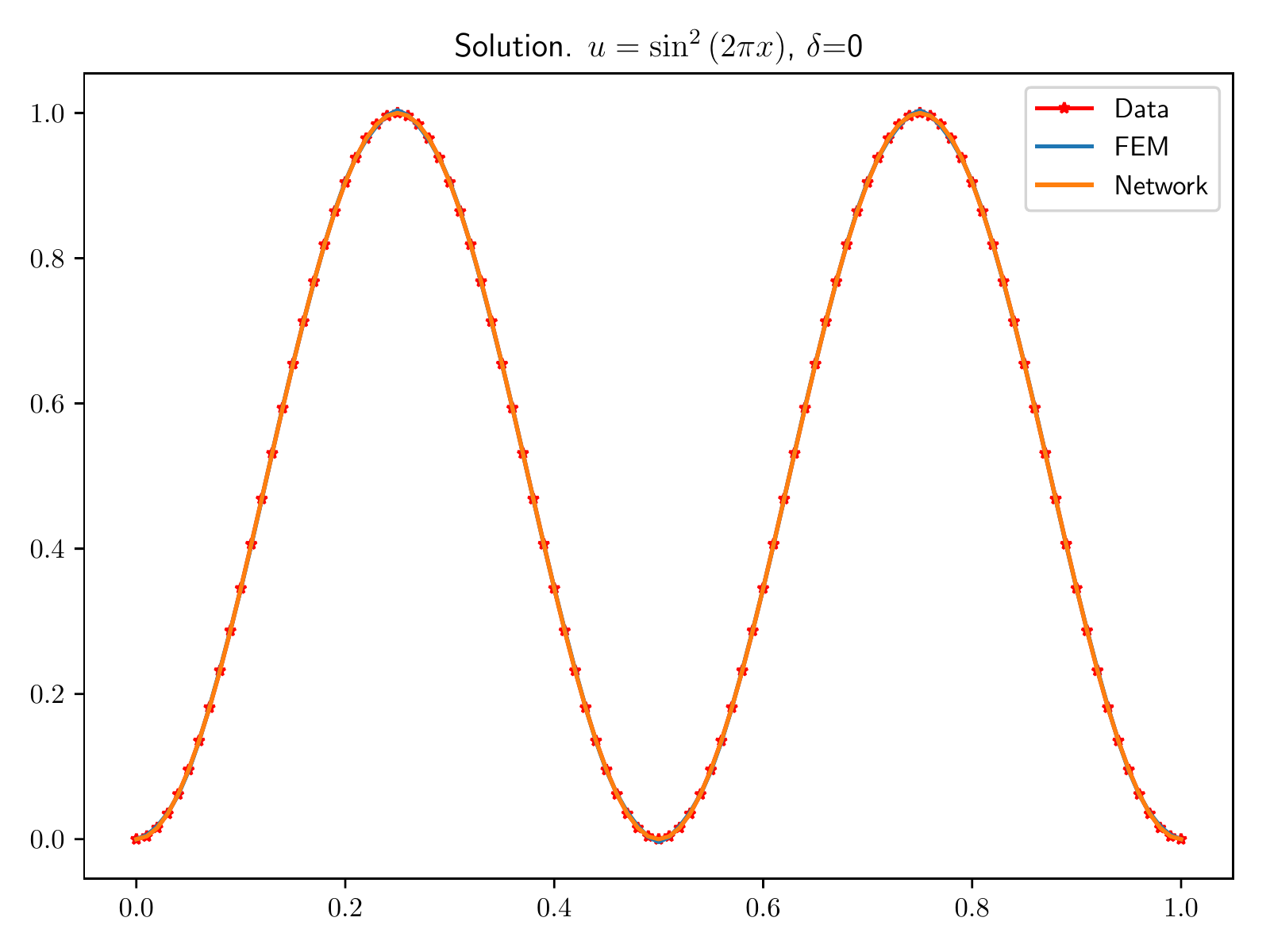}
\caption{Solutions with the optimized coefficients and no noise.}
\end{subfigure}
\begin{subfigure}[t]{0.45\textwidth}
\centering
\includegraphics[width=\textwidth]{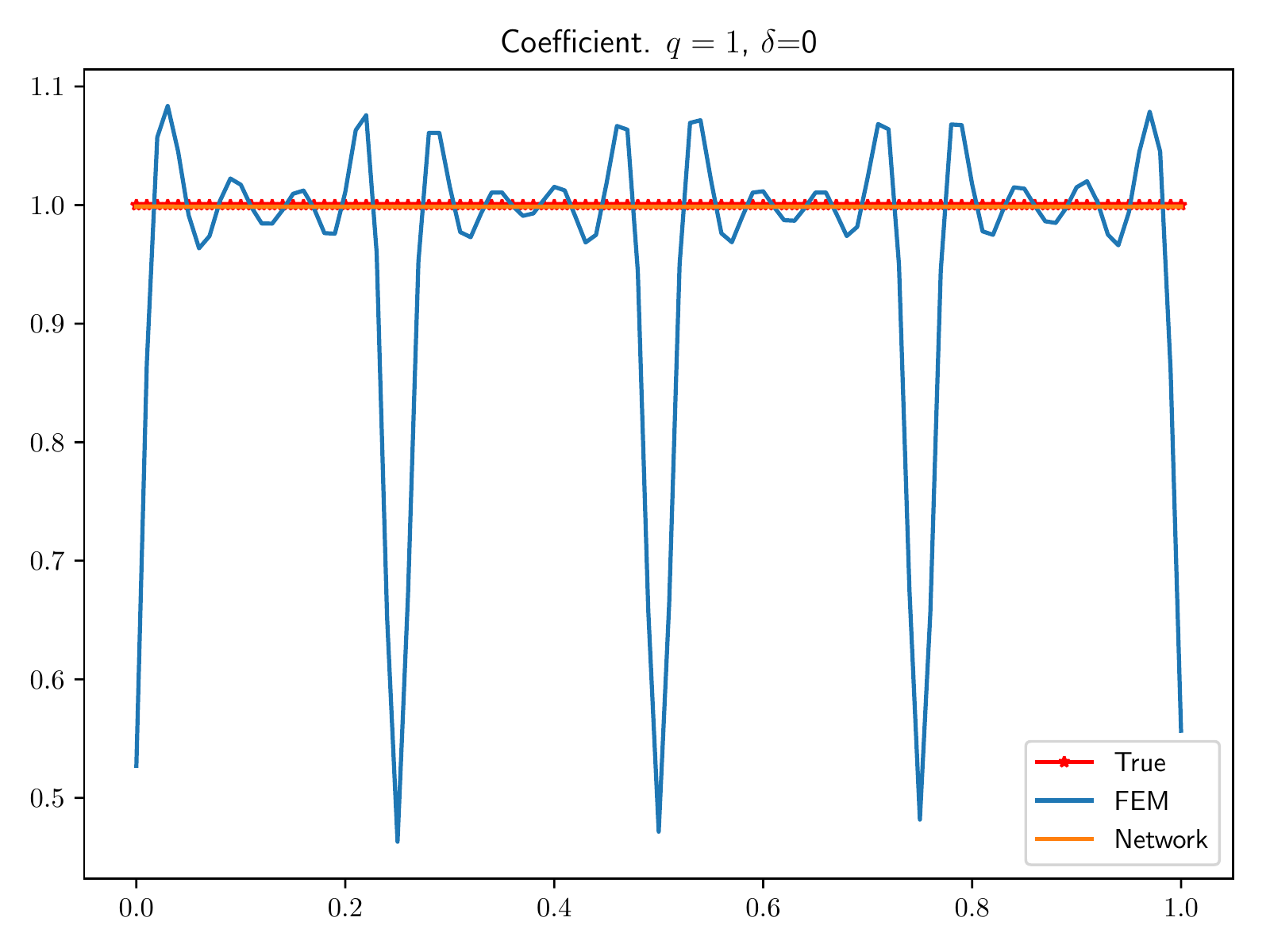}
\caption{The optimized coefficients without noise.}
\end{subfigure}
\begin{subfigure}[t]{0.45\textwidth}
\centering
\includegraphics[width=\textwidth]{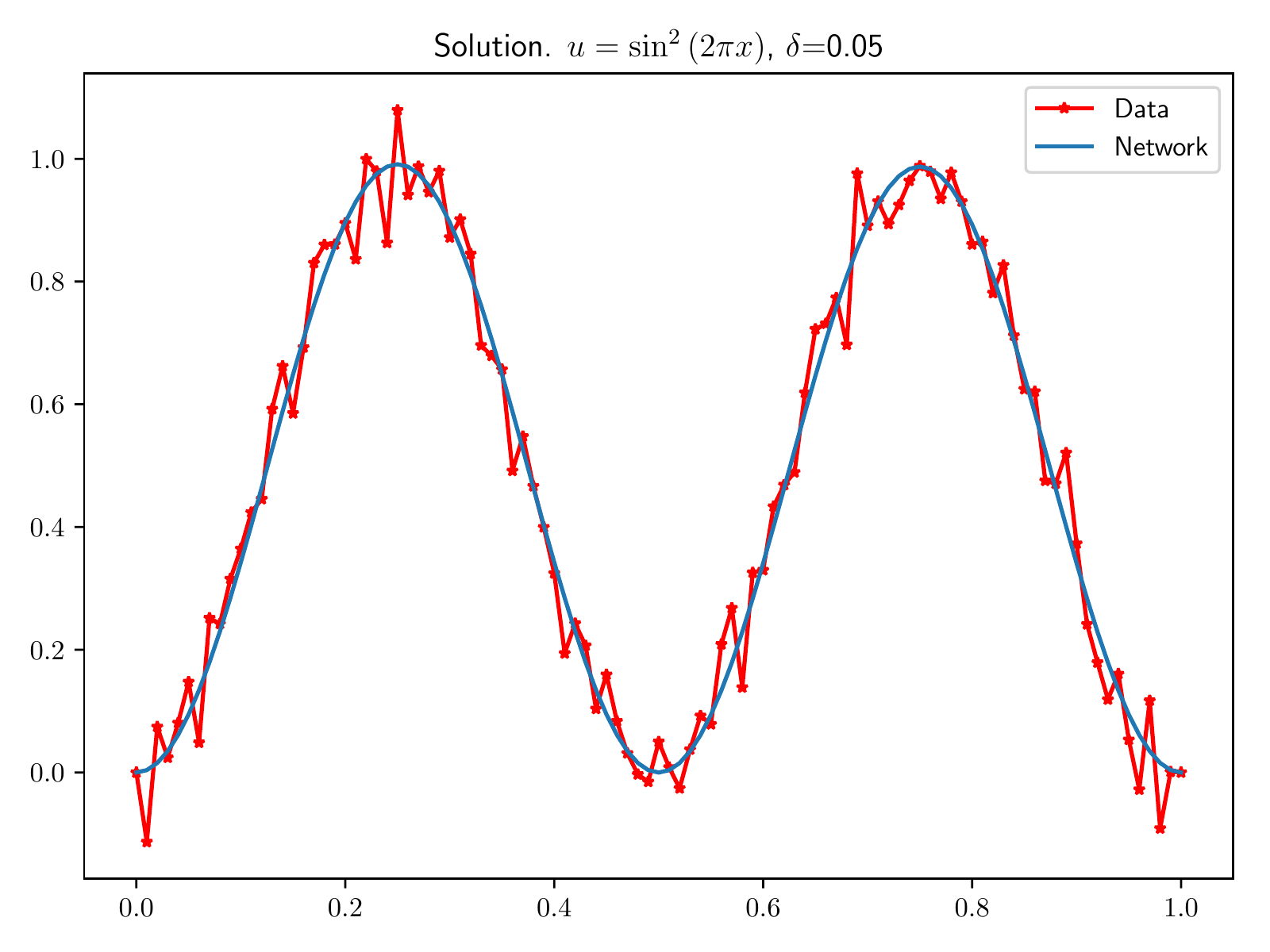}
\caption{Network solution with the optimized coefficient and 5\% noise level.}
\end{subfigure}
\begin{subfigure}[t]{0.45\textwidth}
\centering
\includegraphics[width=\textwidth]{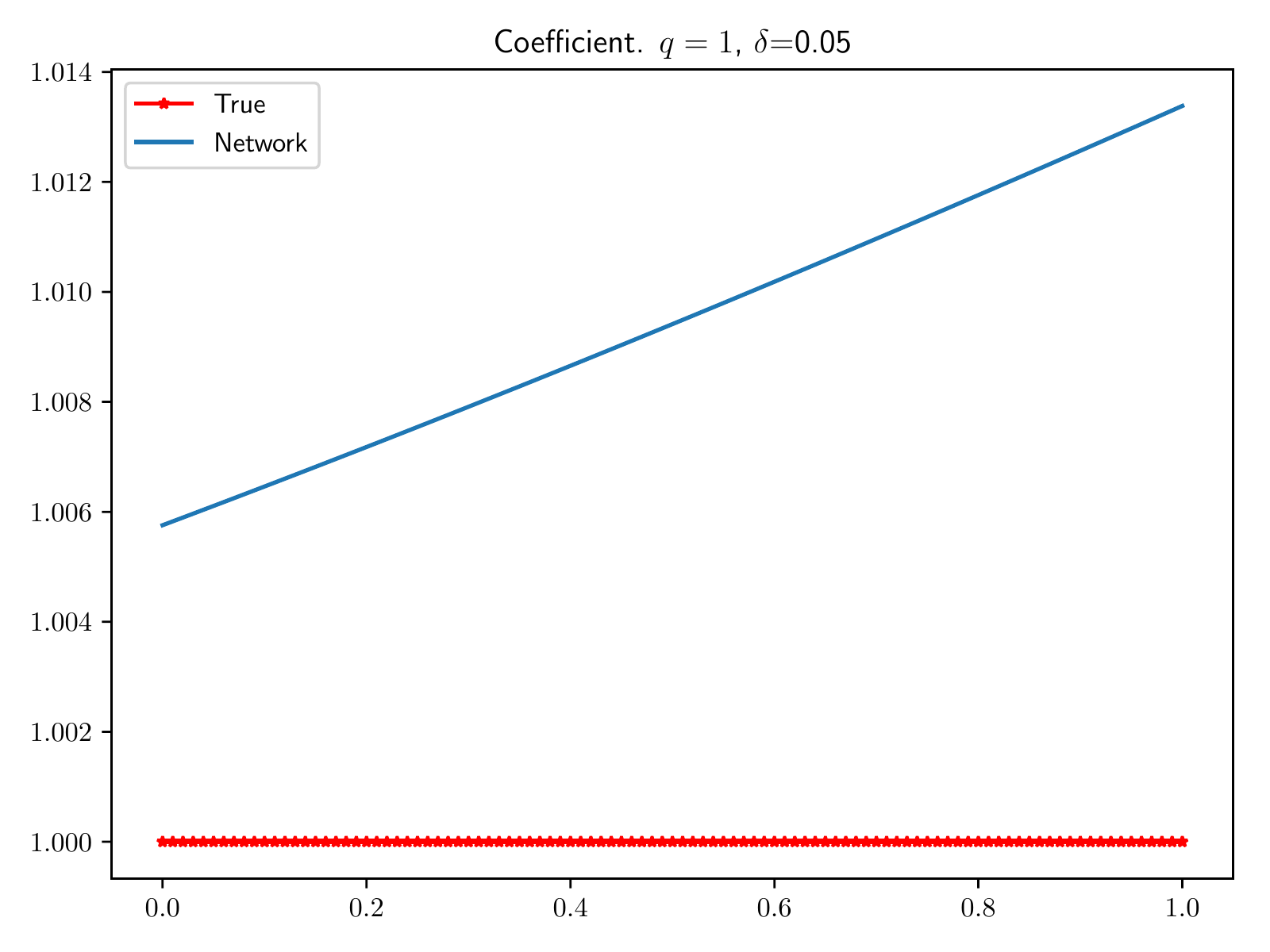}
\caption{The optimized network coefficient with some noise.}
\end{subfigure}
\begin{subfigure}[t]{0.45\textwidth}
\centering
\includegraphics[width=\textwidth]{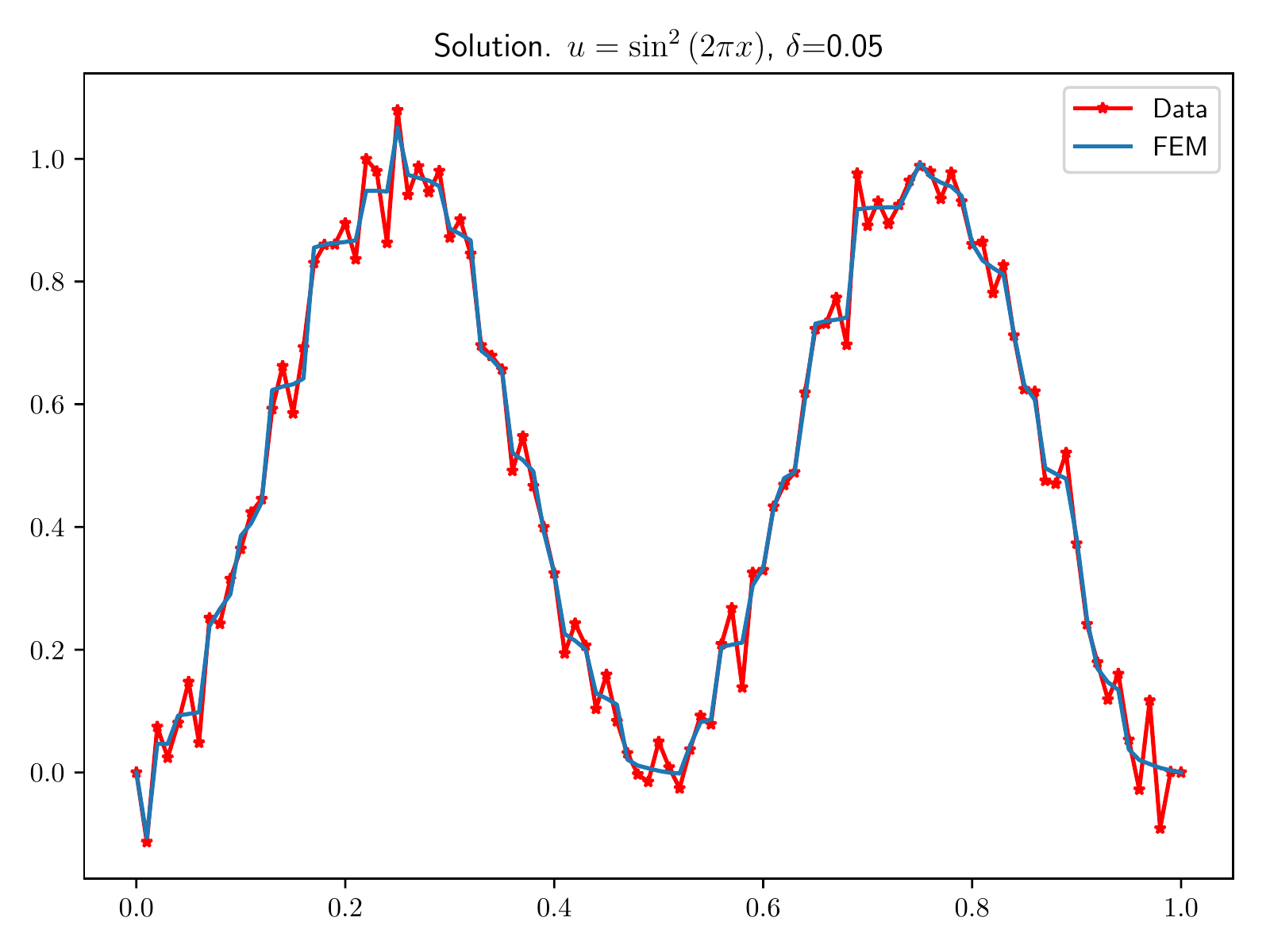}
\caption{FEM solution with the optimized coefficient and 5\% noise level.}
\end{subfigure}
\begin{subfigure}[t]{0.45\textwidth}
\centering
\includegraphics[width=\textwidth]{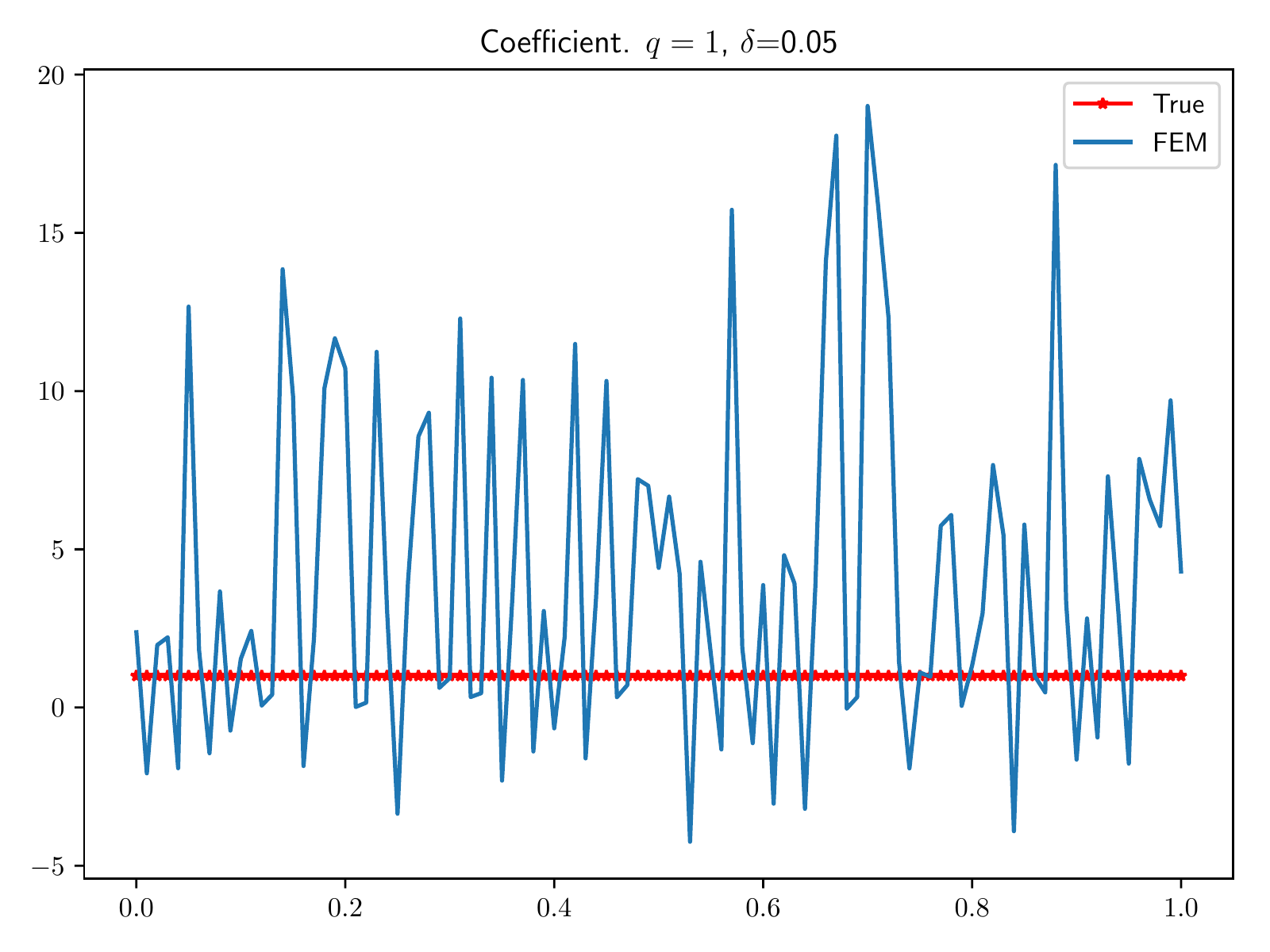}
\caption{The optimized FEM coefficient with some noise.}
\end{subfigure}
\caption{Comparison between FEM and a neural network with constant coefficient $\hat{q}=1$.}
\label{heat1dqconstfigs}
\end{figure}

\begin{figure}[htp]
\centering
\begin{subfigure}[t]{0.45\textwidth}
\centering
\includegraphics[width=\textwidth]{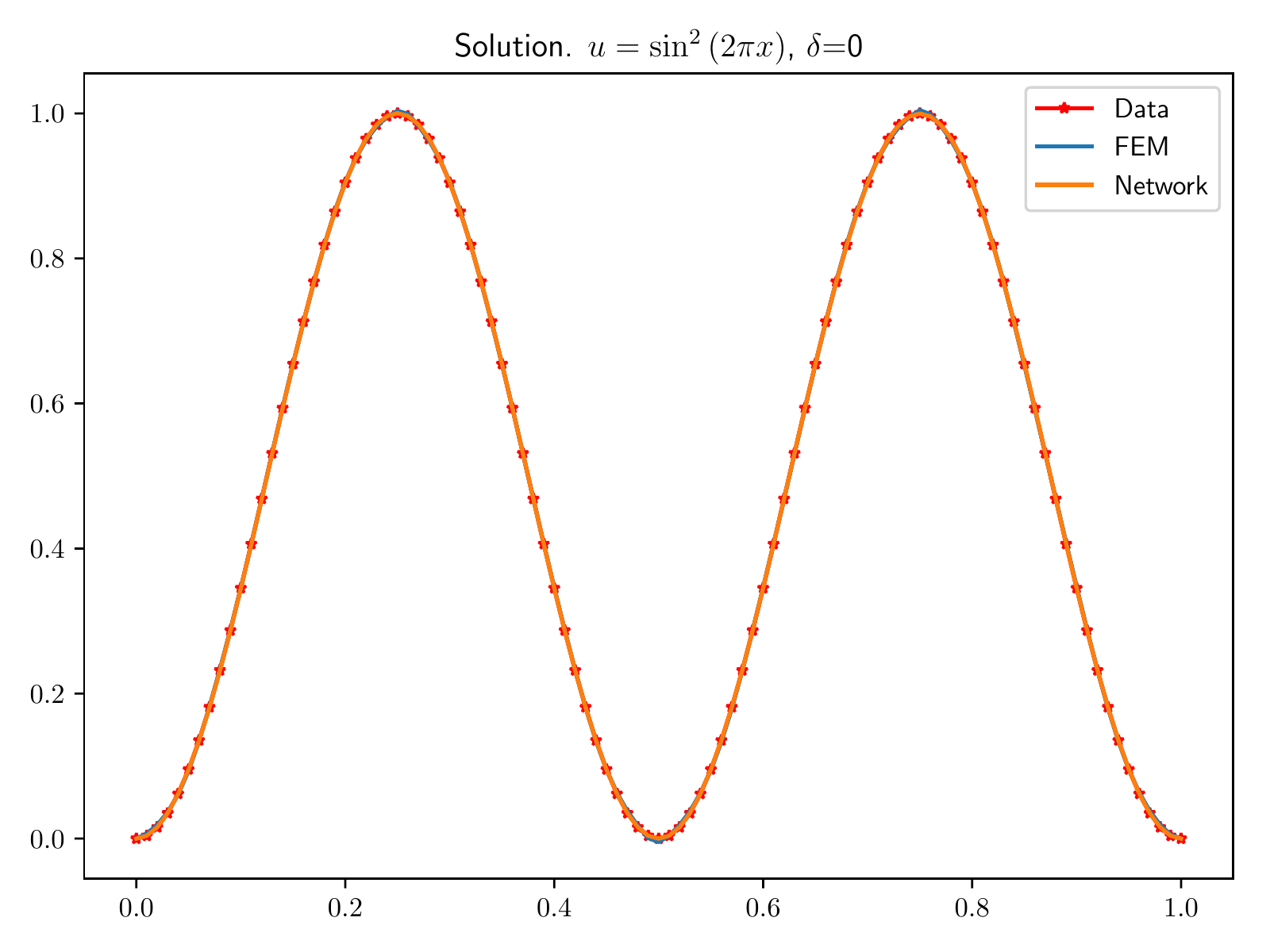}
\caption{Solutions with the optimized coefficients and no noise.}
\end{subfigure}
\begin{subfigure}[t]{0.45\textwidth}
\centering
\includegraphics[width=\textwidth]{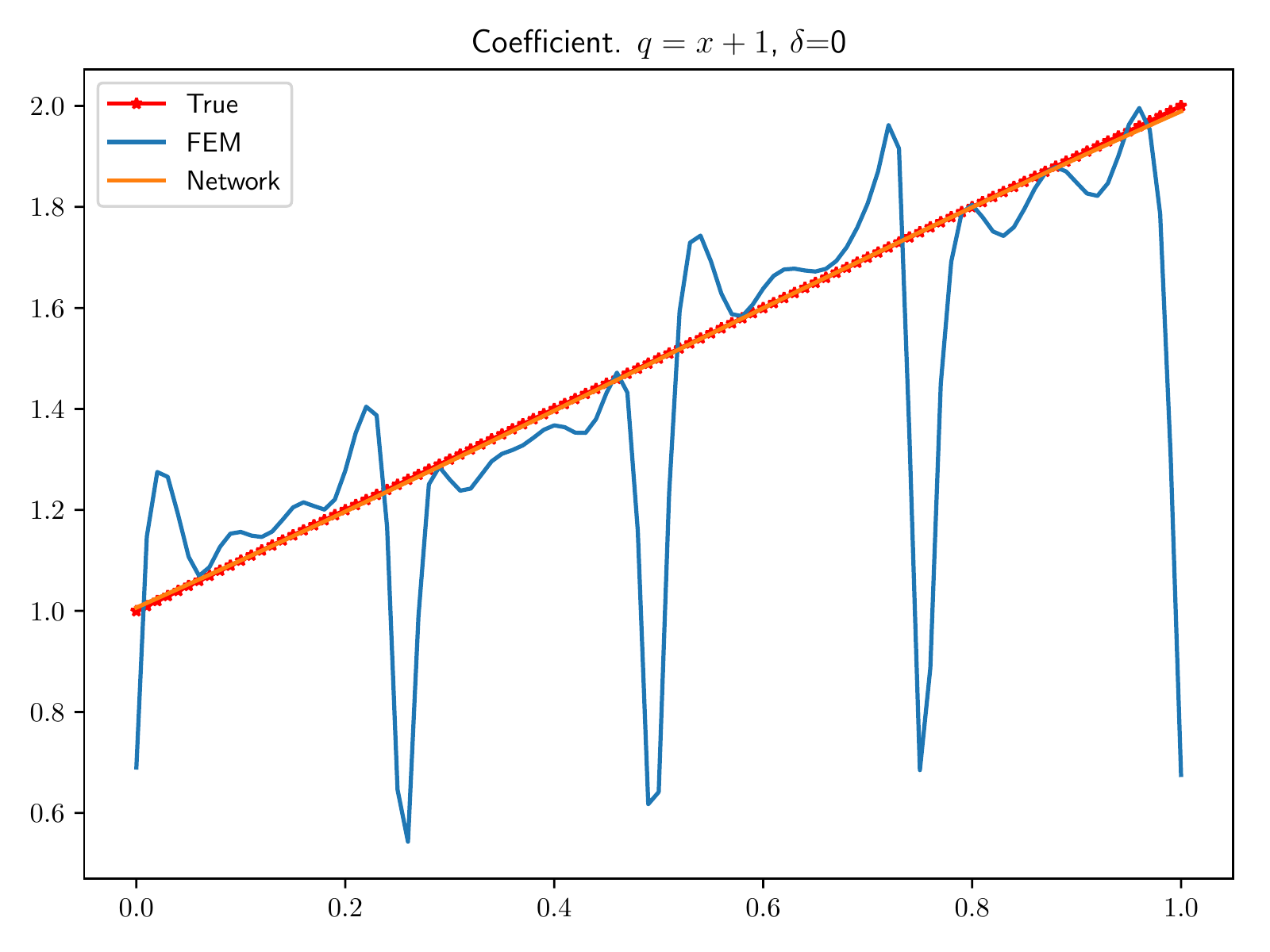}
\caption{The optimized coefficients without noise.}
\end{subfigure}
\begin{subfigure}[t]{0.45\textwidth}
\centering
\includegraphics[width=\textwidth]{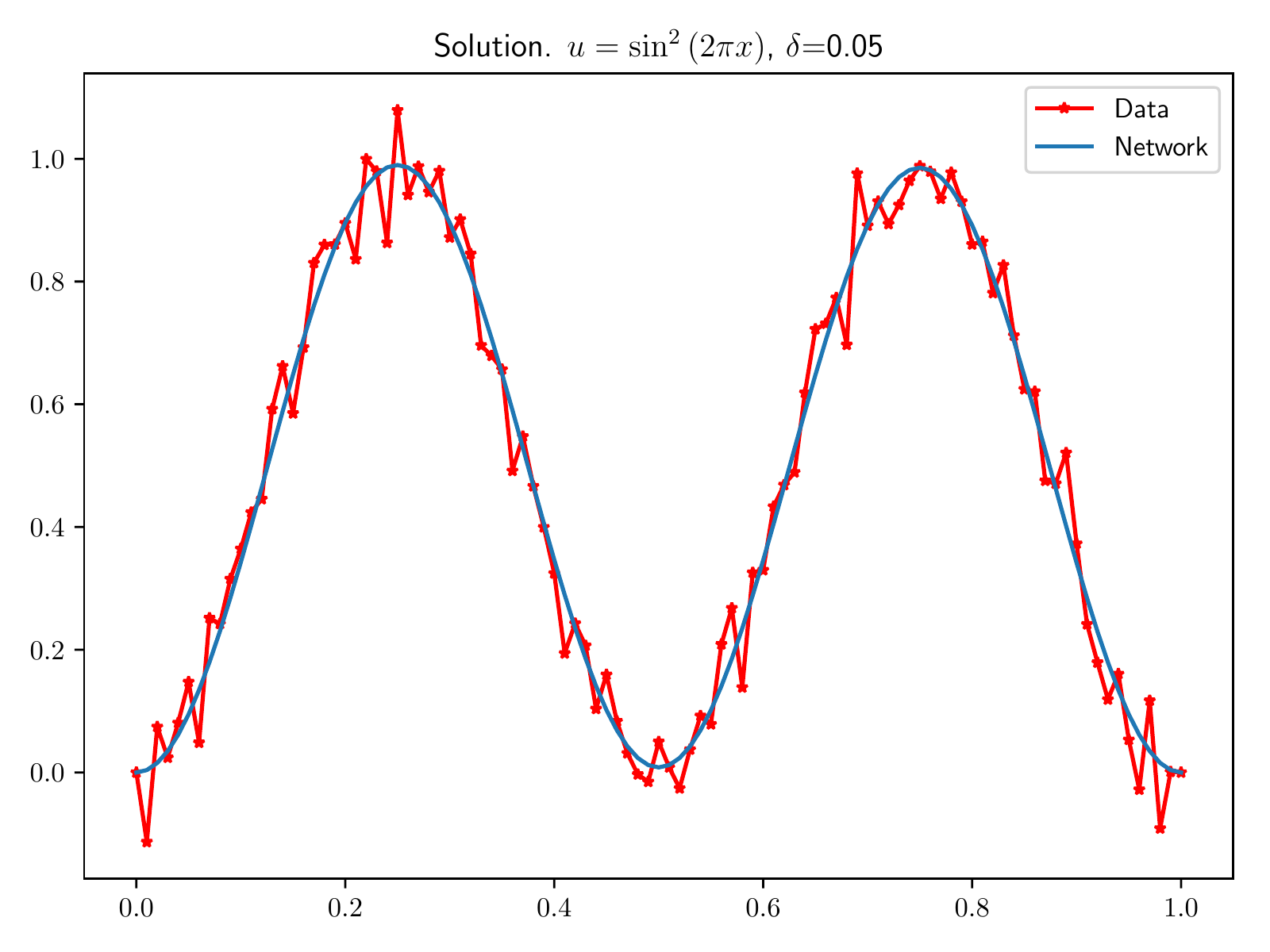}
\caption{Network solution with the optimized coefficient and 5\% noise level.}
\end{subfigure}
\begin{subfigure}[t]{0.45\textwidth}
\centering
\includegraphics[width=\textwidth]{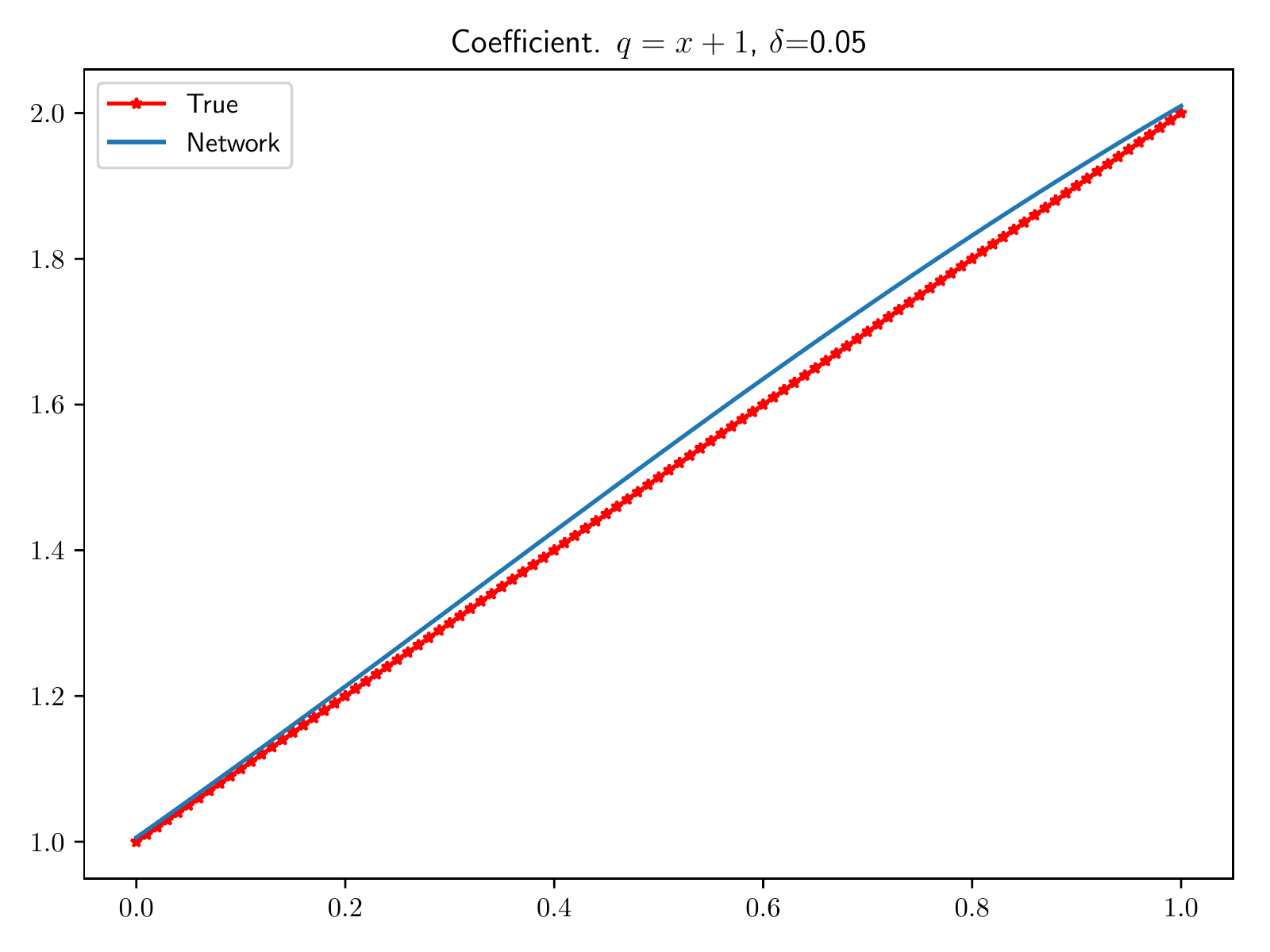}
\caption{The optimized network coefficient with some noise.}
\end{subfigure}
\begin{subfigure}[t]{0.45\textwidth}
\centering
\includegraphics[width=\textwidth]{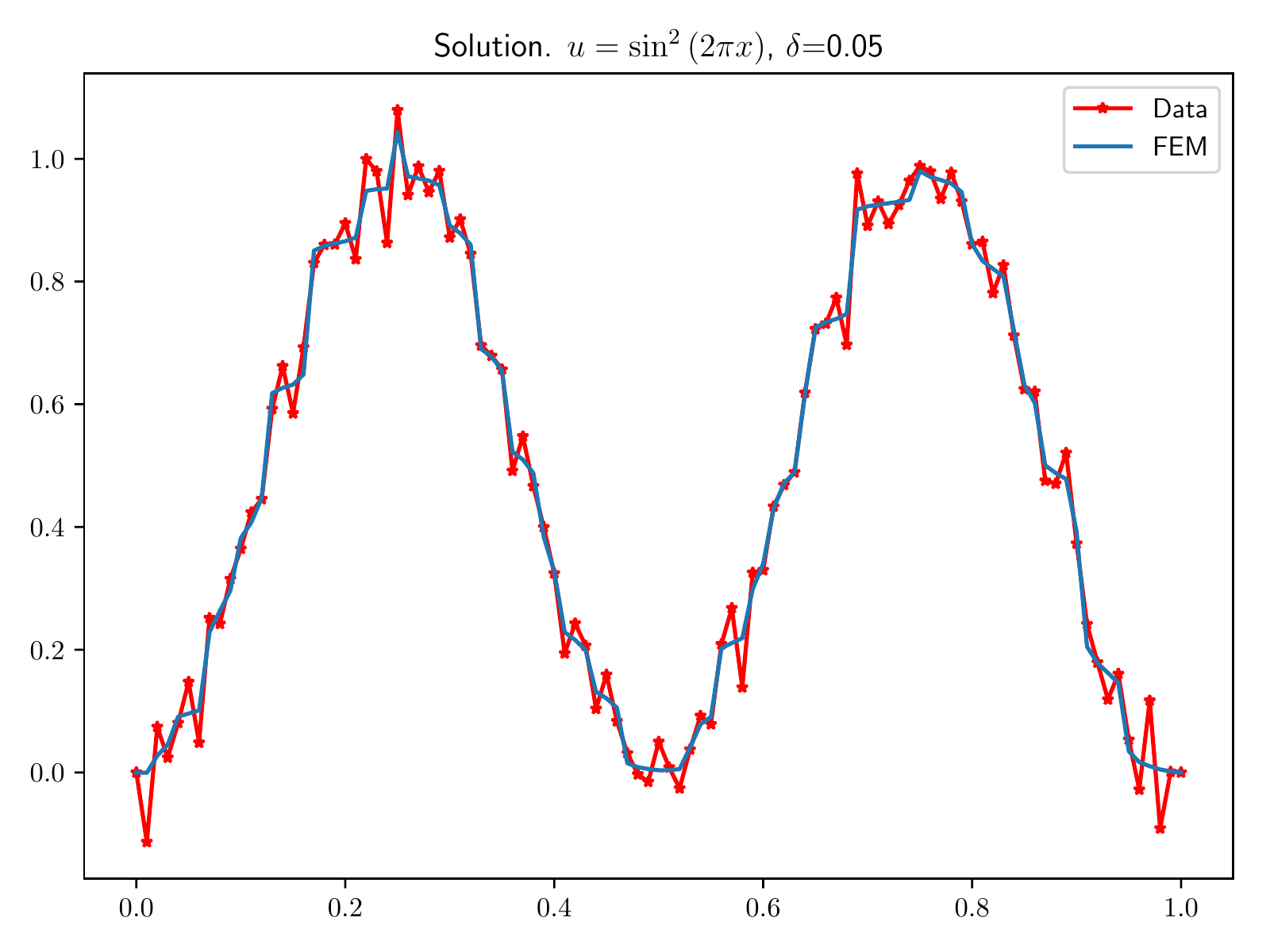}
\caption{FEM solution with the optimized coefficient and 5\% noise level.}
\end{subfigure}
\begin{subfigure}[t]{0.45\textwidth}
\centering
\includegraphics[width=\textwidth]{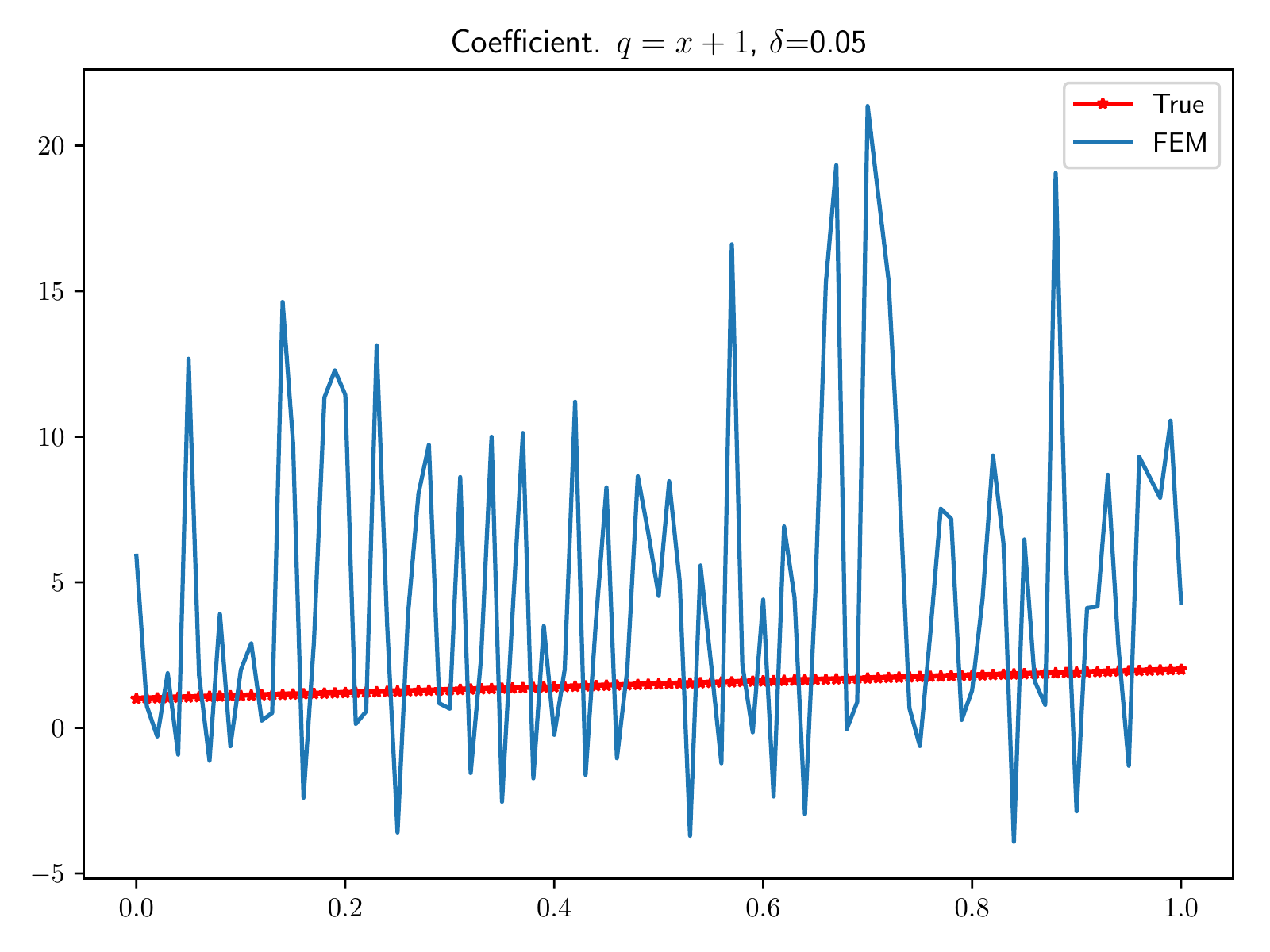}
\caption{The optimized FEM coefficient with some noise.}
\end{subfigure}
\caption{Comparison between FEM and a neural network with linear coefficient $\hat{q}=1+x$.}
\label{heat1dqlinefigs}
\end{figure}

\begin{figure}[htp]
\centering
\begin{subfigure}[t]{0.45\textwidth}
\centering
\includegraphics[width=\textwidth]{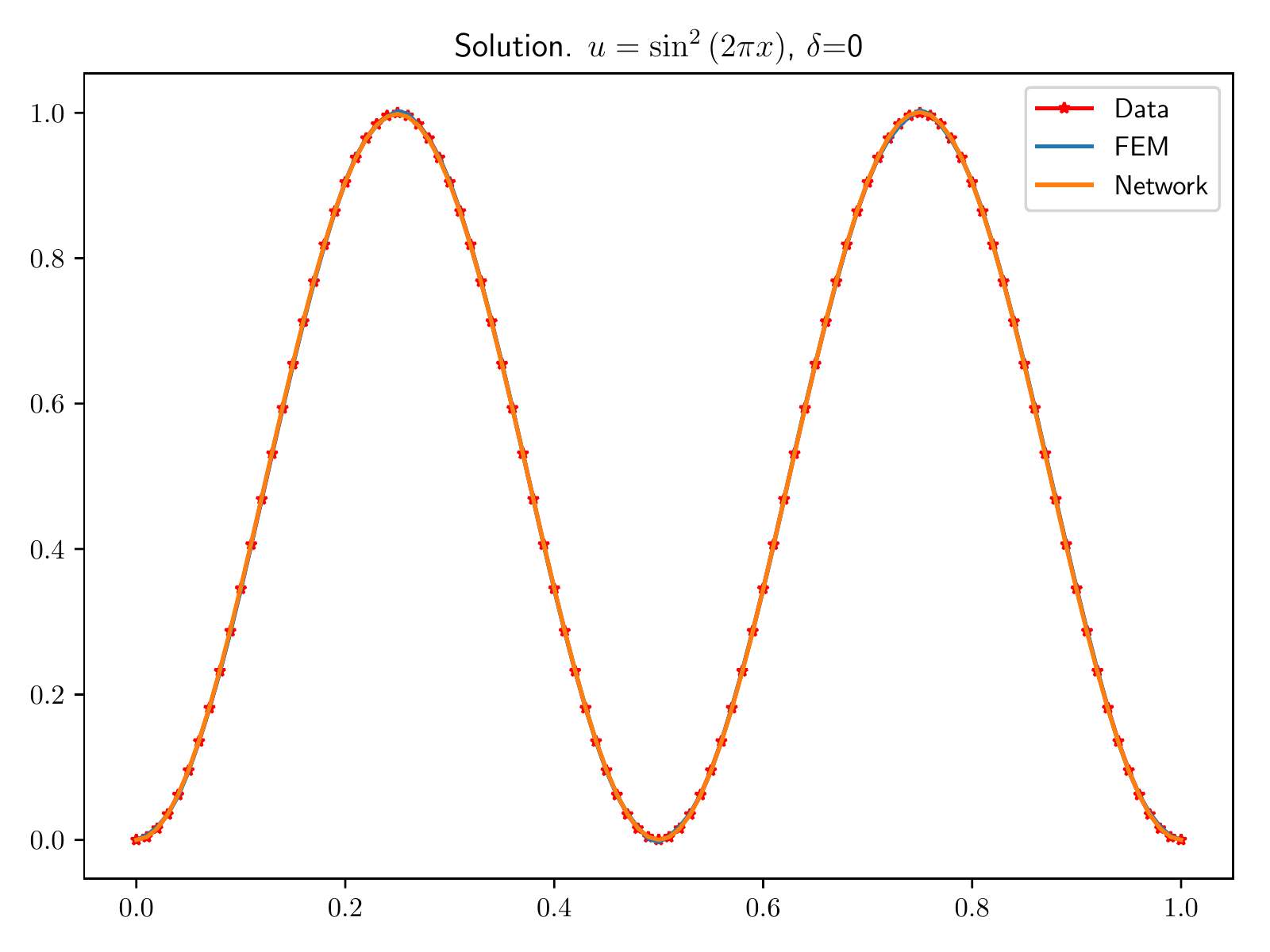}
\caption{Solutions with the optimized coefficients and no noise.}
\end{subfigure}
\begin{subfigure}[t]{0.45\textwidth}
\centering
\includegraphics[width=\textwidth]{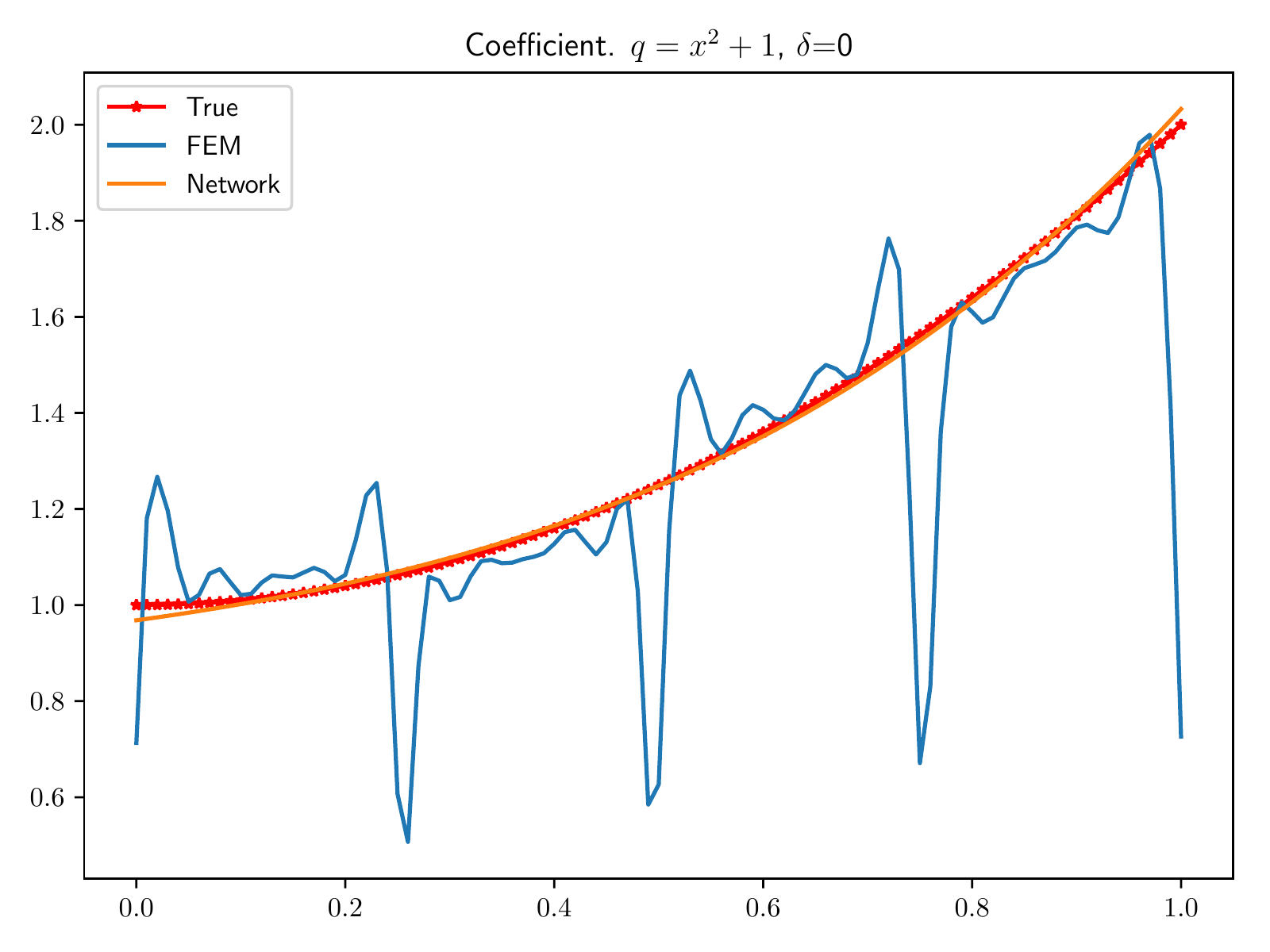}
\caption{The optimized coefficients without noise.}
\end{subfigure}
\begin{subfigure}[t]{0.45\textwidth}
\centering
\includegraphics[width=\textwidth]{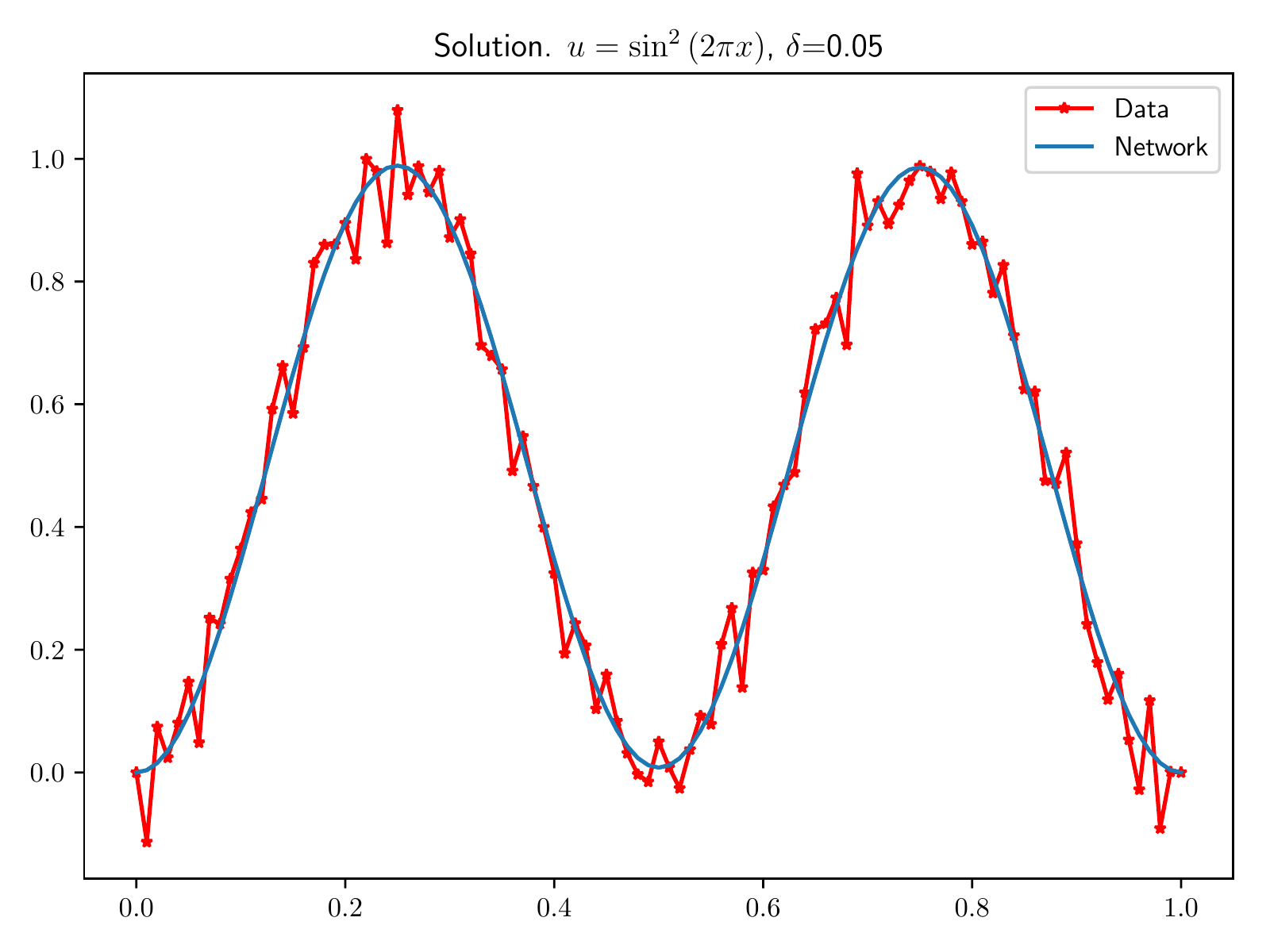}
\caption{Network solution with the optimized coefficient and 5\% noise level.}
\end{subfigure}
\begin{subfigure}[t]{0.45\textwidth}
\centering
\includegraphics[width=\textwidth]{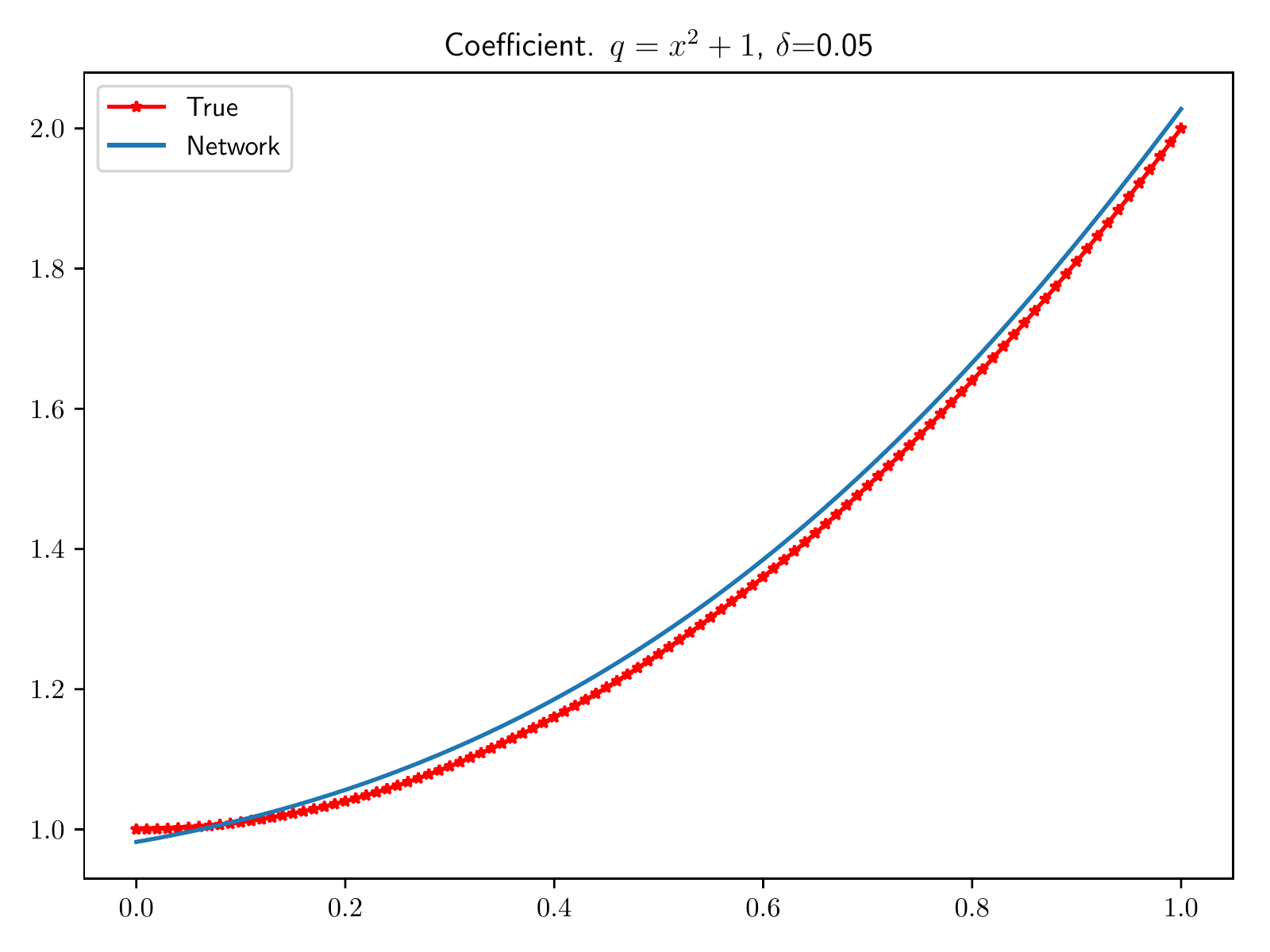}
\caption{The optimized network coefficient with some noise.}
\end{subfigure}
\begin{subfigure}[t]{0.45\textwidth}
\centering
\includegraphics[width=\textwidth]{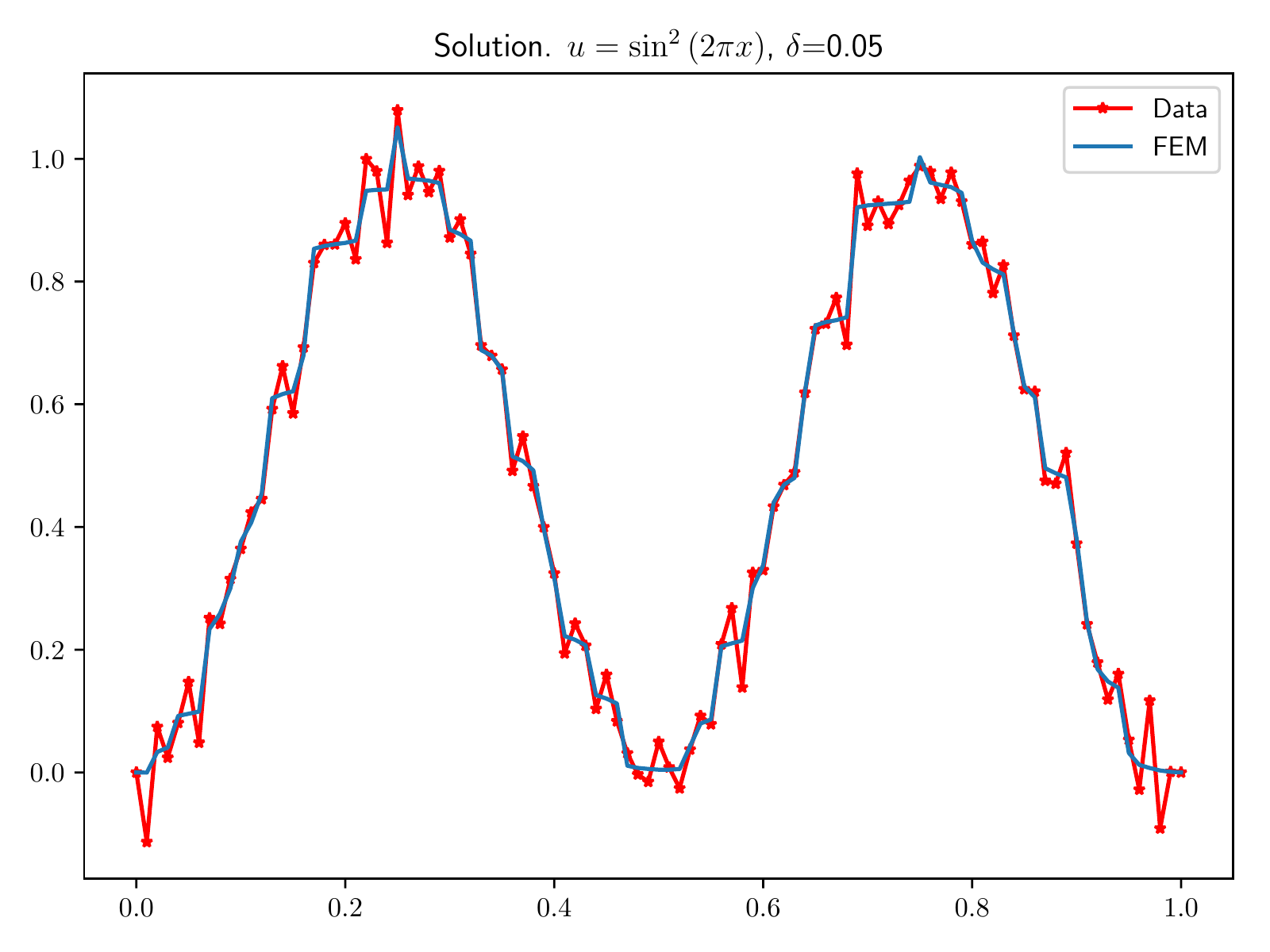}
\caption{FEM solution with the optimized coefficient and 5\% noise level.}
\end{subfigure}
\begin{subfigure}[t]{0.45\textwidth}
\centering
\includegraphics[width=\textwidth]{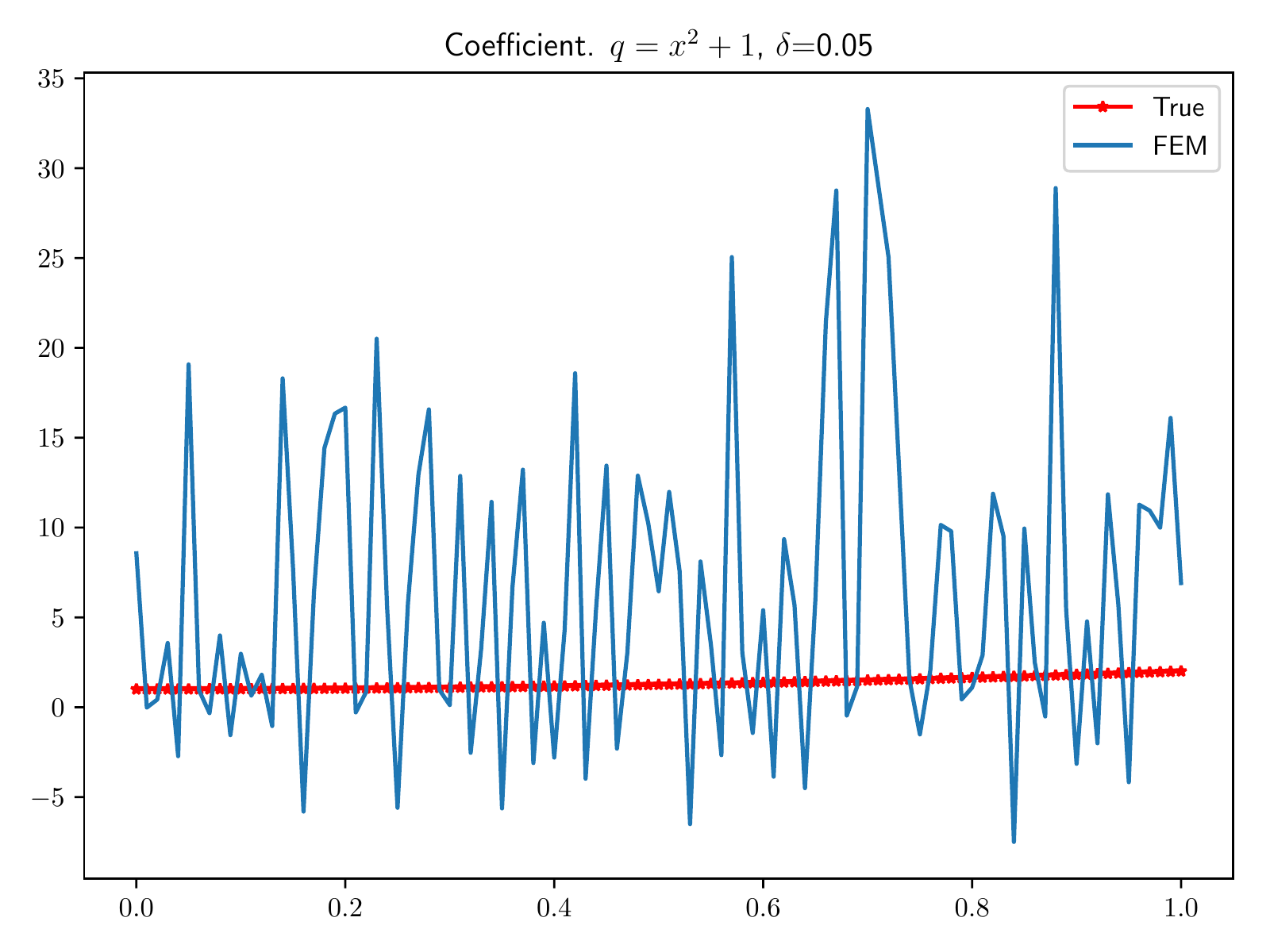}
\caption{The optimized FEM coefficient with some noise.}
\end{subfigure}
\caption{Comparison between FEM and a neural network with quadratic coefficient $\hat{q} = 1+x^2$.}
\label{heat1dqquadfigs}
\end{figure}

\begin{figure}[htp]
\centering
\begin{subfigure}[t]{0.45\textwidth}
\centering
\includegraphics[width=\textwidth]{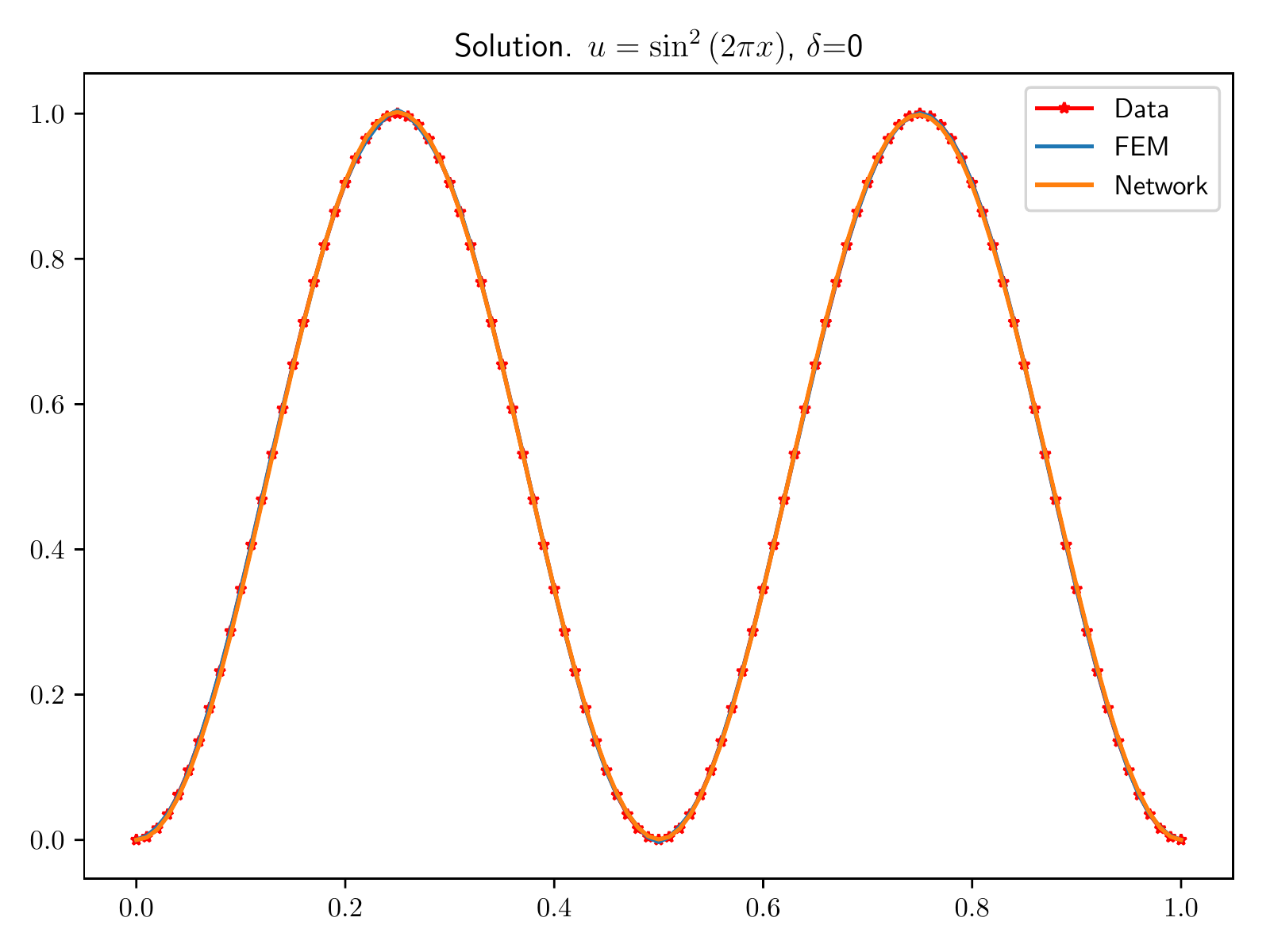}
\caption{Solutions with the optimized coefficients and no noise.}
\end{subfigure}
\begin{subfigure}[t]{0.45\textwidth}
\centering
\includegraphics[width=\textwidth]{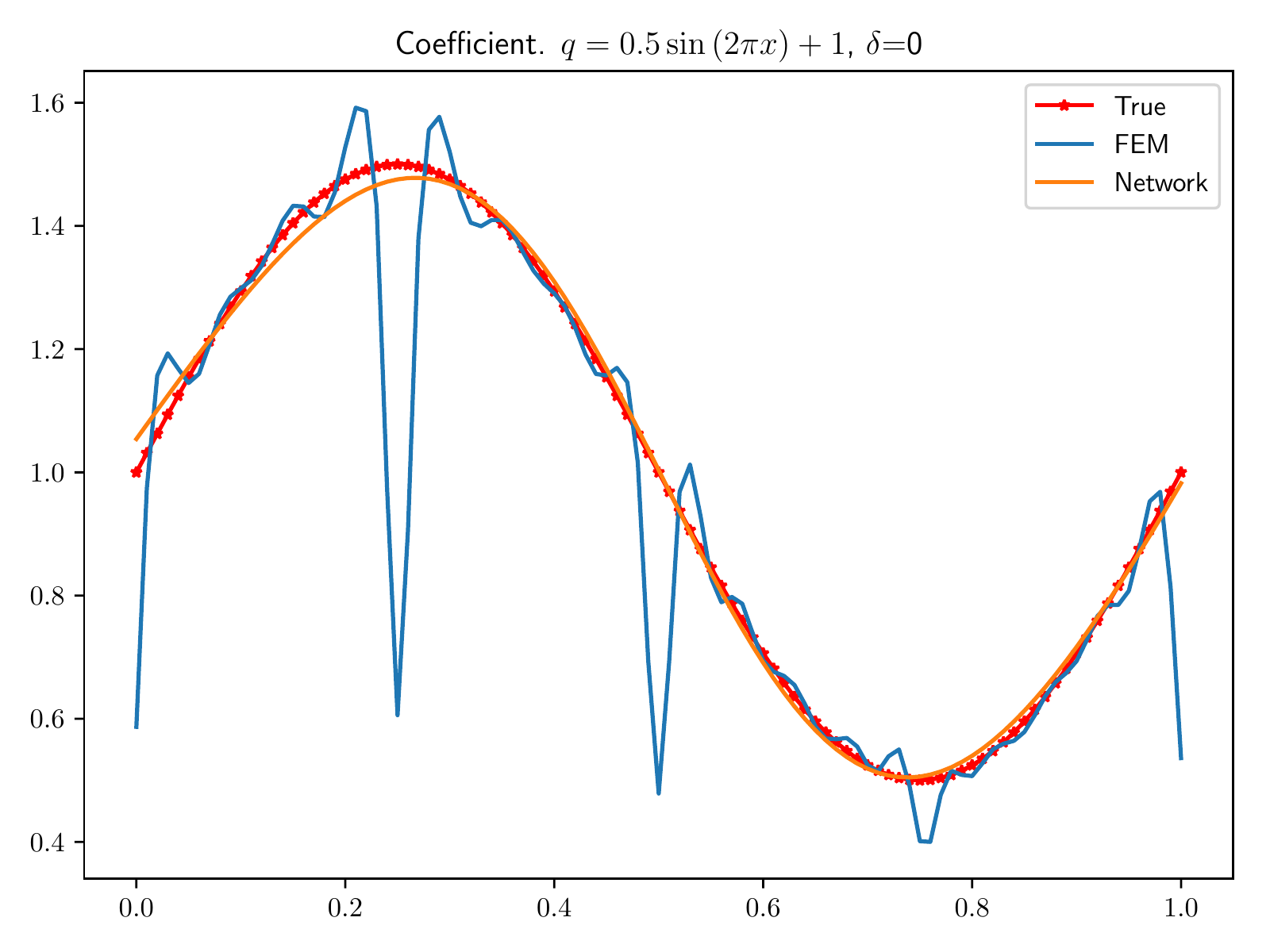}
\caption{The optimized coefficients without noise.}
\end{subfigure}
\begin{subfigure}[t]{0.45\textwidth}
\centering
\includegraphics[width=\textwidth]{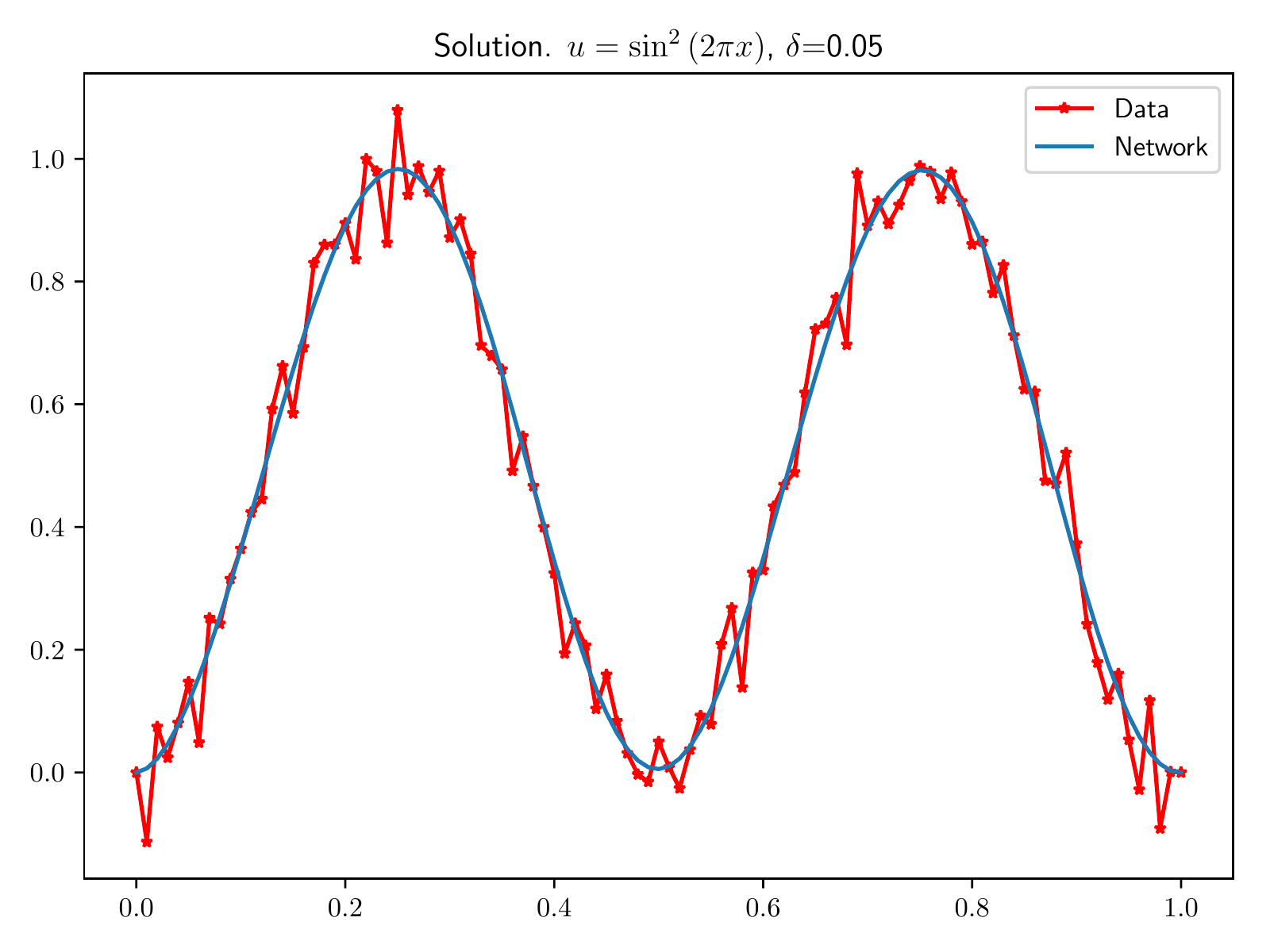}
\caption{Network solution with the optimized coefficient and 5\% noise level.}
\end{subfigure}
\begin{subfigure}[t]{0.45\textwidth}
\centering
\includegraphics[width=\textwidth]{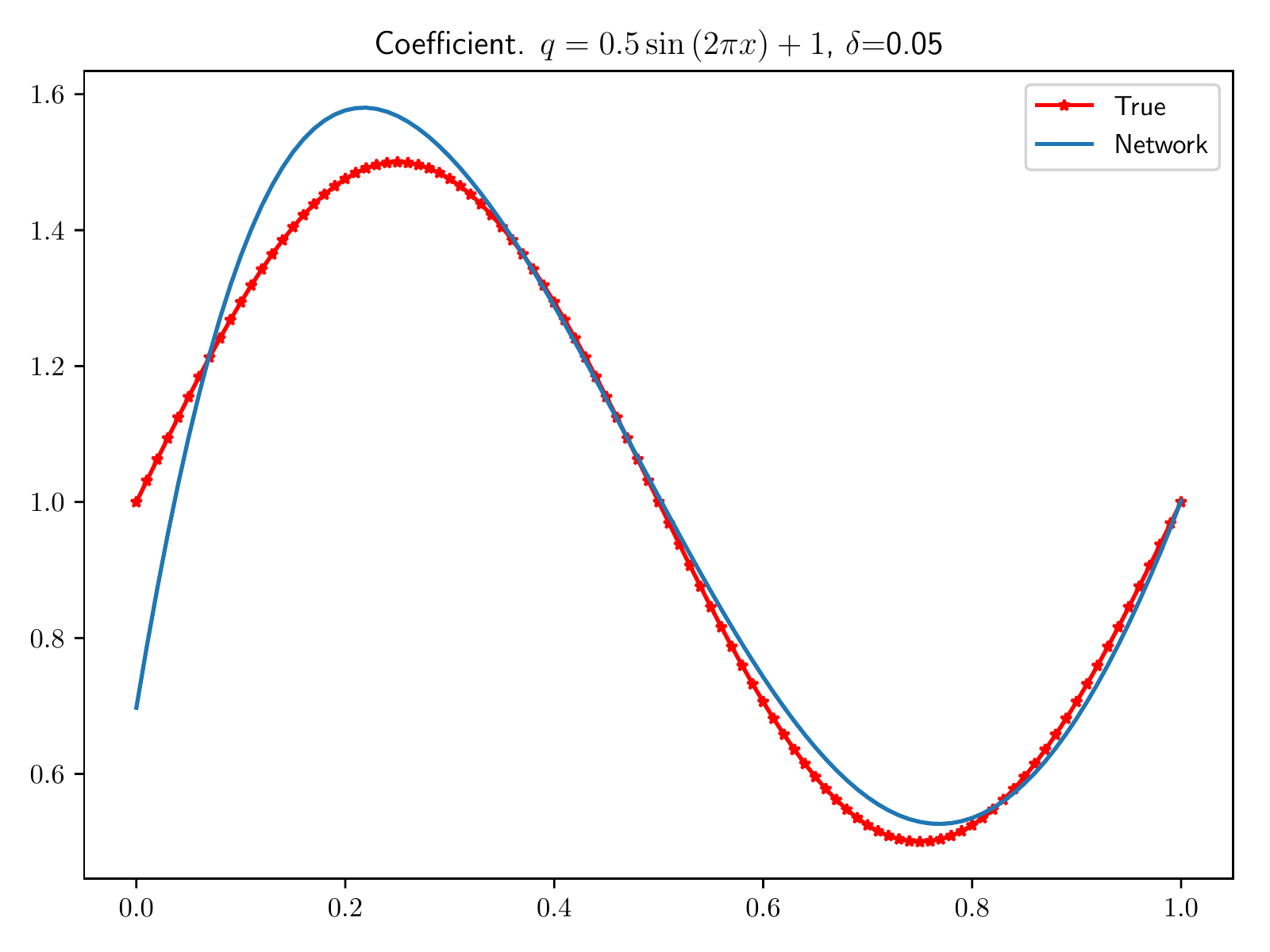}
\caption{The optimized network coefficient with some noise.}
\end{subfigure}
\begin{subfigure}[t]{0.45\textwidth}
\centering
\includegraphics[width=\textwidth]{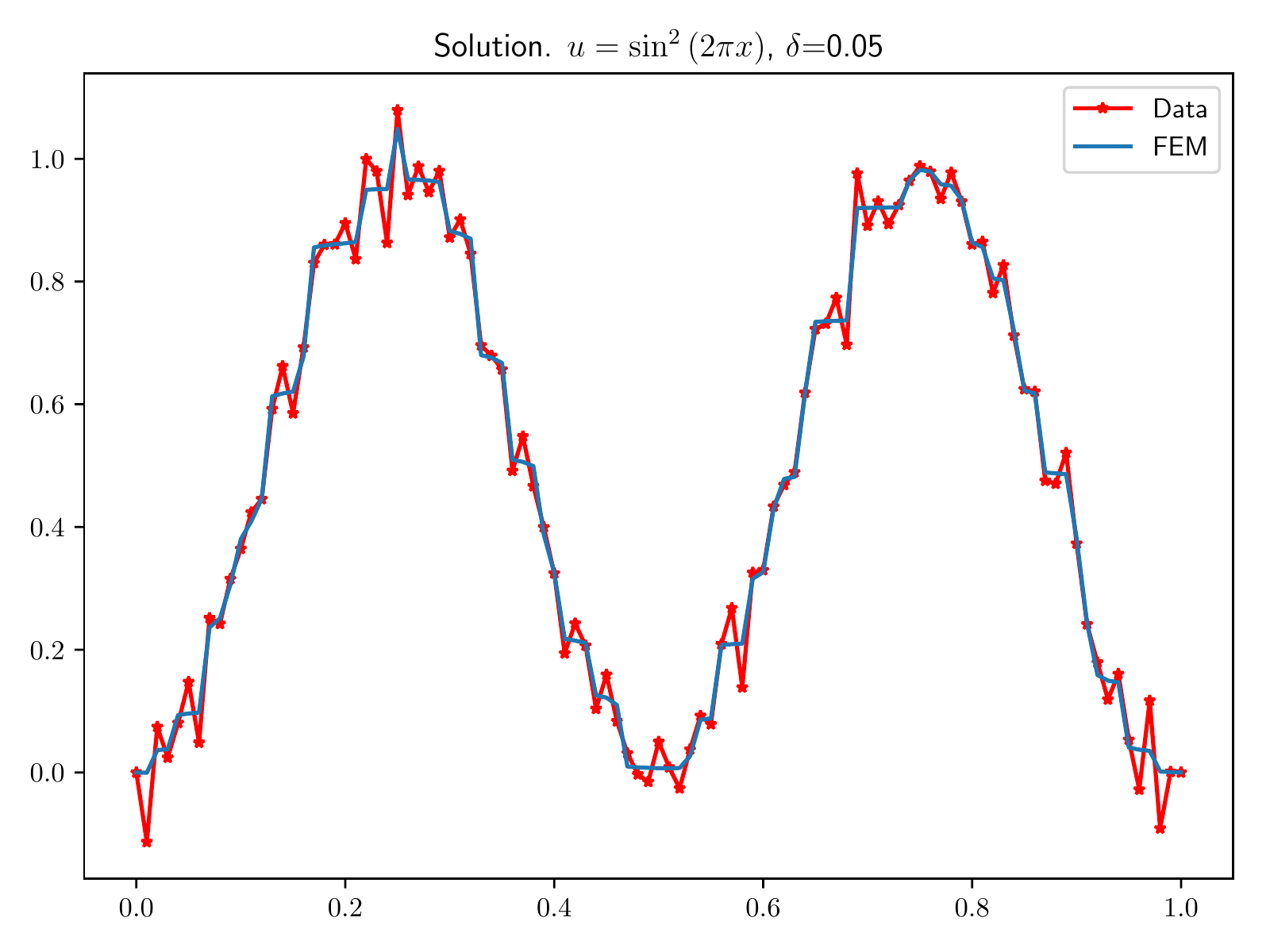}
\caption{FEM solution with the optimized coefficient and 5\% noise level.}
\end{subfigure}
\begin{subfigure}[t]{0.45\textwidth}
\centering
\includegraphics[width=\textwidth]{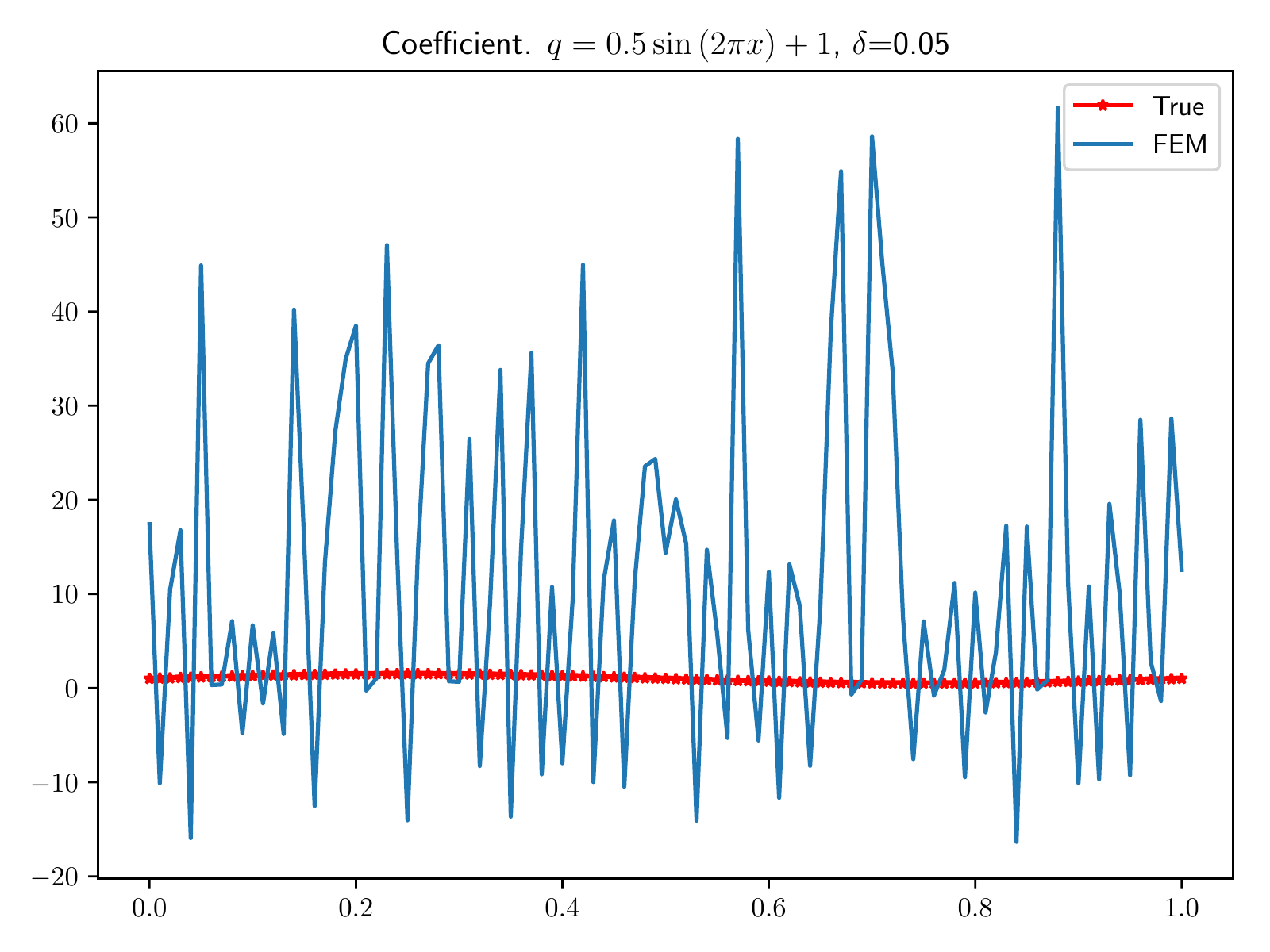}
\caption{The optimized FEM coefficient with some noise.}
\end{subfigure}
\caption{Comparison between FEM and a neural network with coefficient $\hat{q} = 1 + 0.5 \sin(2 \pi x)$.}
\label{heat1dqsinefigs}
\end{figure}

The neural network is a tiny feedforward network with one input neuron, one linear output neuron, and one hidden sigmoid layer with 3 neurons. We choose the solution $u$ to be
\begin{equation}
u(x) = \sin^2(2\pi x)
\end{equation}
and we will compute the inverse problem for a few different exact coefficients $\hat{q}$ and noise levels. The neural network has 10 parameters which will be optimized, compared to the FEM problem which has 101 (same as the number of elements). We consider the exact coefficients $\hat{q}(x) = 1$, $\hat{q}(x) = 1+x$, $\hat{q}(x) = 1+x^2$, and $\hat{q}(x) = 1 + 0.5 \sin(2 \pi x)$ with noise level $\delta=0$ and $\delta=5*10^{-2}$. The results can be seen in Figures~\ref{heat1dqconstfigs}--\ref{heat1dqsinefigs} and Table~\ref{heat1dtable}. It is clear that the network representation outperforms FEM in this case. The network representation is insensitive to noise and always produces smooth solutions and coefficients. FEM does not converge to smooth solutions and coefficients due to the lack of regularization in the error functional. We can also see that the number of iterations increase in the network case as the coefficient becomes more oscillatory. A deeper network with higher capacity might help to mitigate the increase as was seen in \cite{unified}.
\begin{remark}
For one-dimensional problems the number of mesh elements is usually rather limited. This means that the network is at a high risk of overfitting. In particular with noisy data, a network with too high capacity will fit the noise which causes a severe increase in the number of optimization iterations, and of course also lack of generalization. This is the reason we have used a tiny network with only three neurons in the hidden layer. A larger network would require weight regularization, dropout and/or  any other method which reduces overfitting. See for example \cite{efficientbackprop, dropout1, dropout2}.
\end{remark}

\def\arraystretch{1.2}
\begin{table}[htp]
\centering
\begin{tabular}{|c|c|c|c|c|c|c|c|c|}
\hline
& \multicolumn{8}{|c|}{$\hat{q}=1$} \\
\hline
& \multicolumn{4}{c}{$\delta = 0$} &  \multicolumn{4}{|c|}{$\delta = 0.05$} \\
\hline
& \#it & Time (s) & $||u - \hat{u}||$ & $||q-\hat{q}||$ & \#it & Time (s) & $||u-\hat{u}||$ & $||q-\hat{q}||$ \\
\hline
Network & 19 & 1 & 3.65e-5 & 1.38e-3 & 19 & 1 & 6.67e-3 & 9.72e-3 \\
FEM & 184 & 9 & 1.11e-3 & 1.30e-1 & 1748 & 103 & 3.25e-2 & 5.38e+00 \\
\hline
\hline
& \multicolumn{8}{|c|}{$\hat{q}=x + 1$} \\
\hline
& \multicolumn{4}{c}{$\delta = 0$} &  \multicolumn{4}{|c|}{$\delta = 0.05$} \\
\hline
& \#it & Time (s) & $||u - \hat{u}||$ & $||q-\hat{q}||$ & \#it & Time (s) & $||u-\hat{u}||$ & $||q-\hat{q}||$ \\
\hline
Network & 29 & 2 & 2.26e-4 & 3.43e-3 & 37 & 2 & 7.47e-3 & 2.55e-2 \\
FEM & 195 & 10 & 1.44e-3 & 2.50e-1 & 1228 & 75 & 3.07e-2 & 5.70e+00 \\
\hline
\hline
& \multicolumn{8}{|c|}{$\hat{q}=x^{2} + 1$} \\
\hline
& \multicolumn{4}{c}{$\delta = 0$} &  \multicolumn{4}{|c|}{$\delta = 0.05$} \\
\hline
& \#it & Time (s) & $||u - \hat{u}||$ & $||q-\hat{q}||$ & \#it & Time (s) & $||u-\hat{u}||$ & $||q-\hat{q}||$ \\
\hline
Network & 61 & 3 & 1.20e-3 & 1.10e-2 & 56 & 3 & 7.45e-3 & 2.28e-2 \\
FEM & 218 & 13 & 1.05e-3 & 2.05e-1 & 1856 & 105 & 3.14e-2 & 8.84e+00 \\
\hline
\hline
& \multicolumn{8}{|c|}{$\hat{q}=0.5 \sin{\left (2 \pi x \right )} + 1$} \\
\hline
& \multicolumn{4}{c}{$\delta = 0$} &  \multicolumn{4}{|c|}{$\delta = 0.05$} \\
\hline
& \#it & Time (s) & $||u - \hat{u}||$ & $||q-\hat{q}||$ & \#it & Time (s) & $||u-\hat{u}||$ & $||q-\hat{q}||$ \\
\hline
Network & 324 & 14 & 2.56e-3 & 1.78e-2 & 240 & 11 & 1.07e-2 & 6.11e-2 \\
FEM & 218 & 11 & 9.55e-4 & 1.42e-1 & 2116 & 117 & 3.14e-2 & 1.79e+01 \\
\hline
\end{tabular}
\caption{Performance comparison between FEM and neural network representation of the coefficient for the 1D Poisson equation. \#it is the number of iterations until the norm of the gradient of the error functional is less than $10^{-6}$. The norm is measured as the integral of a high-order interpolation onto the finite element space as provided by the \texttt{errornorm} function in \texttt{FEniCS}, and $\hat{u}$ is the unperturbed exact solution in the case of added noise.}
\label{heat1dtable}
\end{table}

\subsection{Two-dimensional Poisson}
The higher-dimensional Poisson equation with Dirichlet boundary conditions is given by
\begin{equation}
\begin{aligned}
-\nabla \cdot (q \nabla u) &= f, && x \in \Omega, \\
u &= g, && x \in \partial \Omega,
\end{aligned}
\label{heat2d}
\end{equation}
where $\Omega \subset \mathbb{R}^N$ is the computational domain, $\partial \Omega$ its boundary. The functions $f$, $g$ are known, and $q = q(x, y): \mathbb{R}^N \to \mathbb{R}_+$ is the a priori unknown coefficient to be estimated. Here, we choose the solution to be
\begin{equation}
u = \sin(\pi x) \sin(\pi y),
\label{heat2dsol}
\end{equation}
for the exact coefficients $\hat{q}(x, y) = 1$, $\hat{q}(x, y) = 1 + x + y$, $\hat{q}(x, y) = 1 + x^2 + y^2$, and $\hat{q}(x, y) = 1 + 0.5 \sin(2 \pi x) \sin(2 \pi y)$. The network is a single hidden layer feedforward network with 10 neurons in the hidden layer. The optimization is performed using BFGS and we iterate until the norm of the gradient of the error functional is less than $10^{-7}$.

\subsubsection{Unit square with uniform mesh}
Here, we let $\Omega = [0 ,1] \times [0, 1]$ be the unit square discretized by $101 \times 101$ piecewise linear elements elements.  The coefficient represented by the network has 41 parameters, while it has 10201 when represented in the finite element space. Due to the very different nature of the optimization problems, a direct comparison is no longer meaningful. The BFGS method scales quadratically in complexity with the number of optimization parameters, and is only useful for small-scale optimization problems, where it is very efficient. The FEM minimization problems can instead be solved by, for example, the memory limited BFGS method \cite{lbfgs, lbfgsb}. By doing so, we see the same pattern as in the 1D case. FEM does not converge to smooth solutions or coefficients with added noise due to the lack of regularization.

\begin{figure}[htp]
\centering
\begin{subfigure}[t]{0.49\textwidth}
\centering
\includegraphics[width=\textwidth]{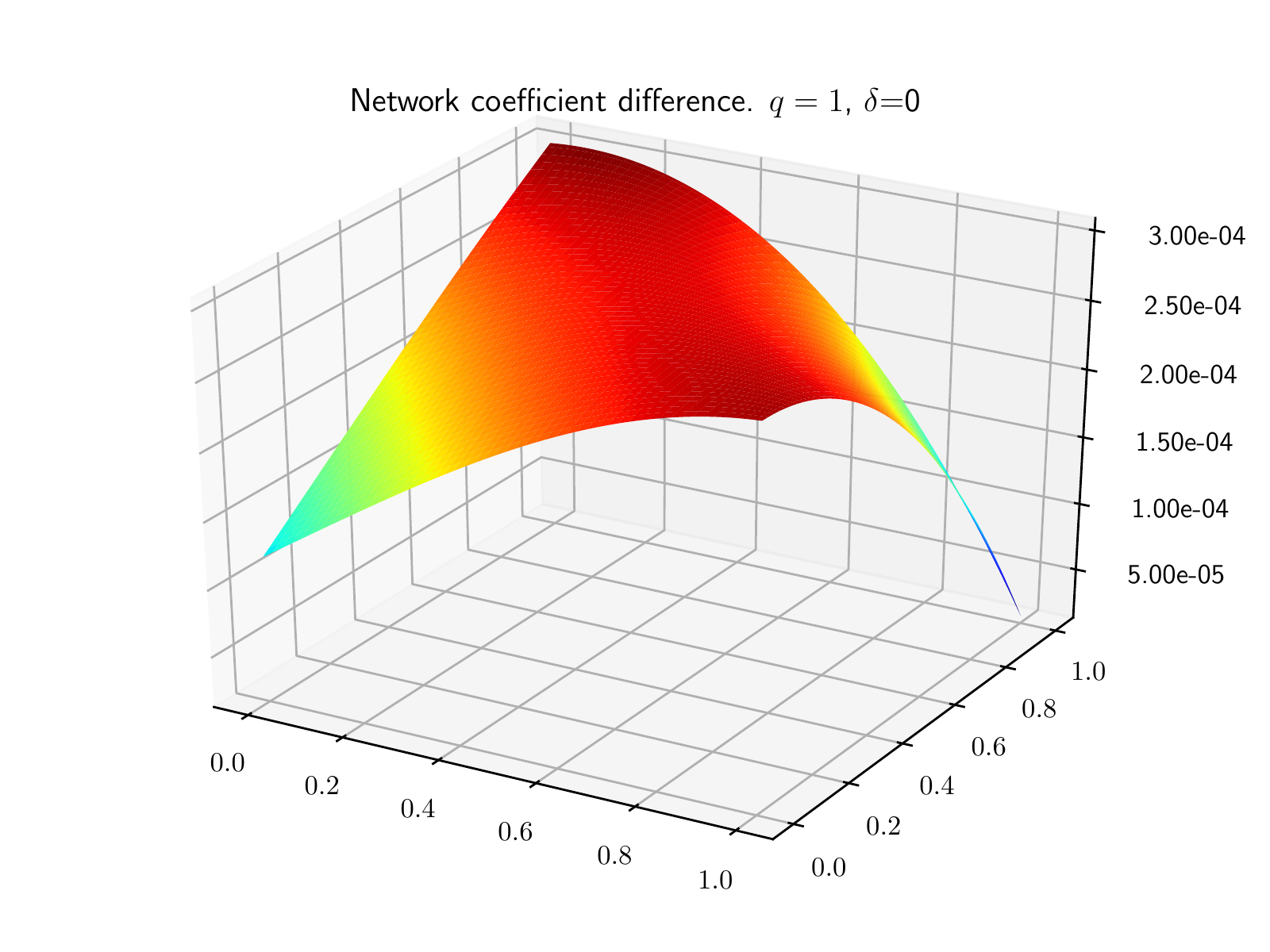}
\caption{The optimized network coefficient without noise.}
\end{subfigure}
\begin{subfigure}[t]{0.49\textwidth}
\centering
\includegraphics[width=\textwidth]{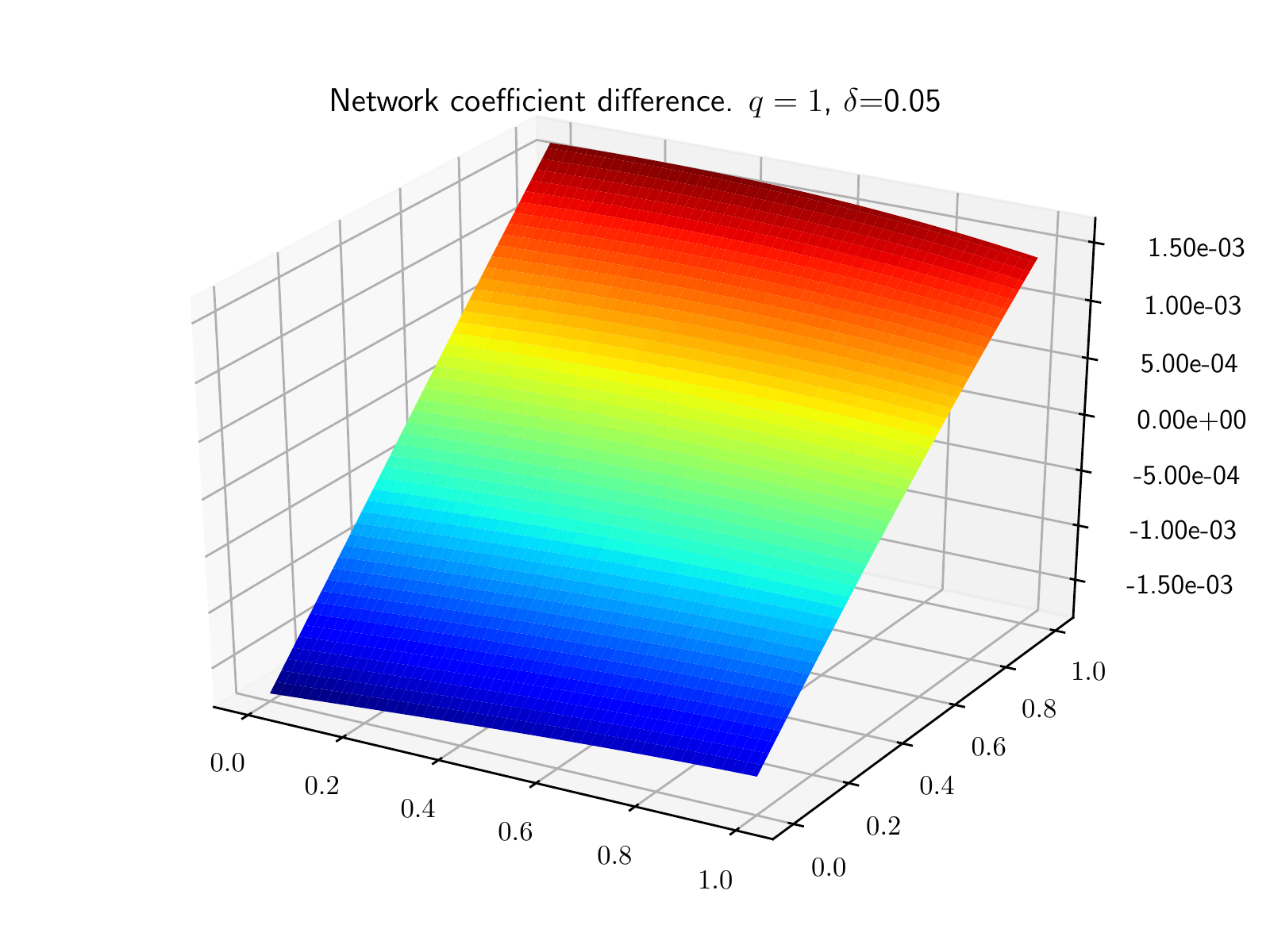}
\caption{The optimized network coefficient with some noise.}
\end{subfigure}
\caption{Difference between the optimized network and exact coefficient when $\hat{q}=1$.}
\label{heat2dsquareqconstfigs}
\end{figure}

\begin{figure}[htp]
\centering
\begin{subfigure}[t]{0.49\textwidth}
\centering
\includegraphics[width=\textwidth]{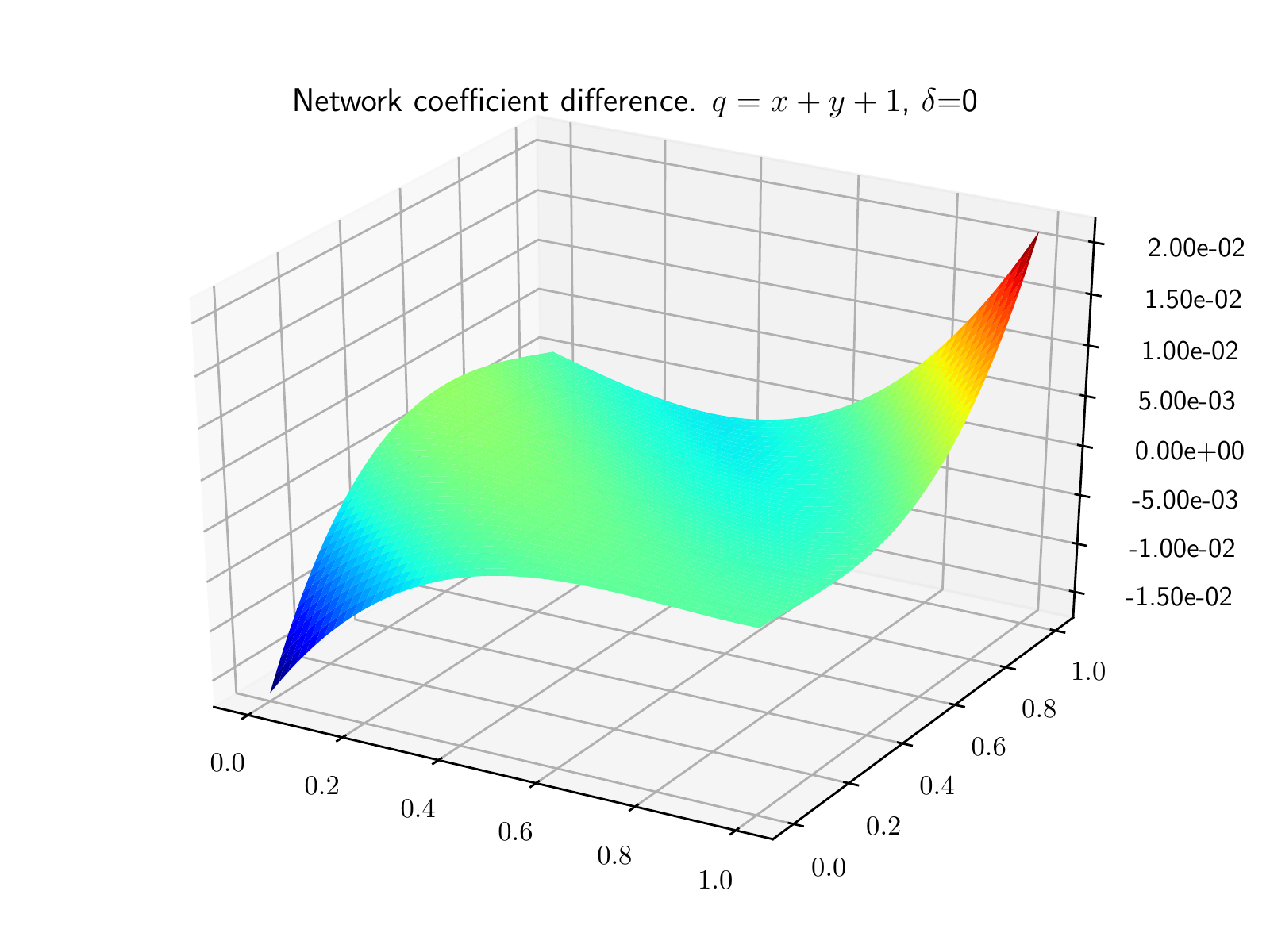}
\caption{The optimized network coefficient without noise.}
\end{subfigure}
\begin{subfigure}[t]{0.49\textwidth}
\centering
\includegraphics[width=\textwidth]{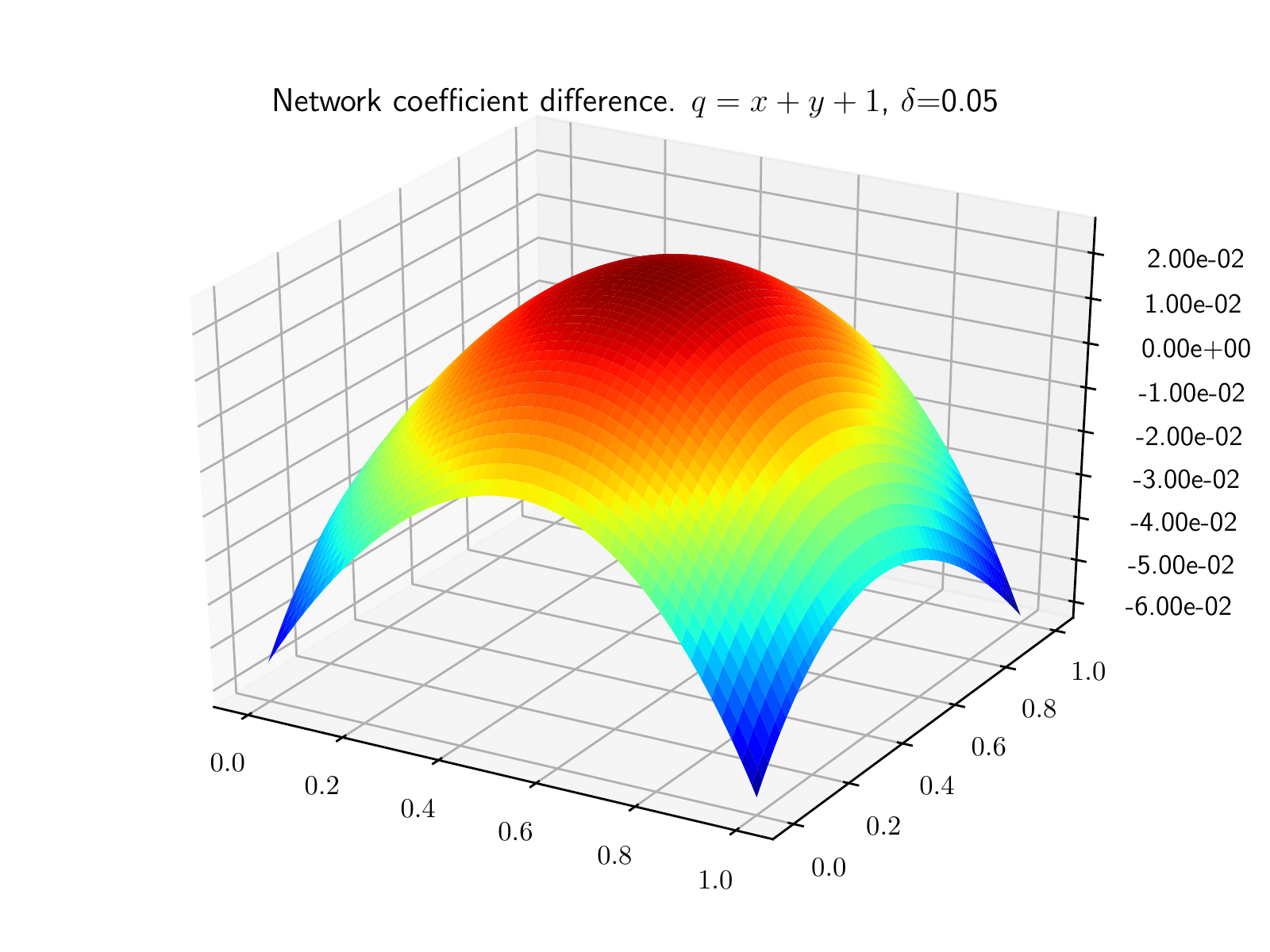}
\caption{The optimized network coefficient with some noise.}
\end{subfigure}
\caption{Difference between the optimized network and exact coefficient when $\hat{q}=1 + x + y$.}
\label{heat2dsquareqlinefigs}
\end{figure}

\begin{figure}[htp]
\centering
\begin{subfigure}[t]{0.49\textwidth}
\centering
\includegraphics[width=\textwidth]{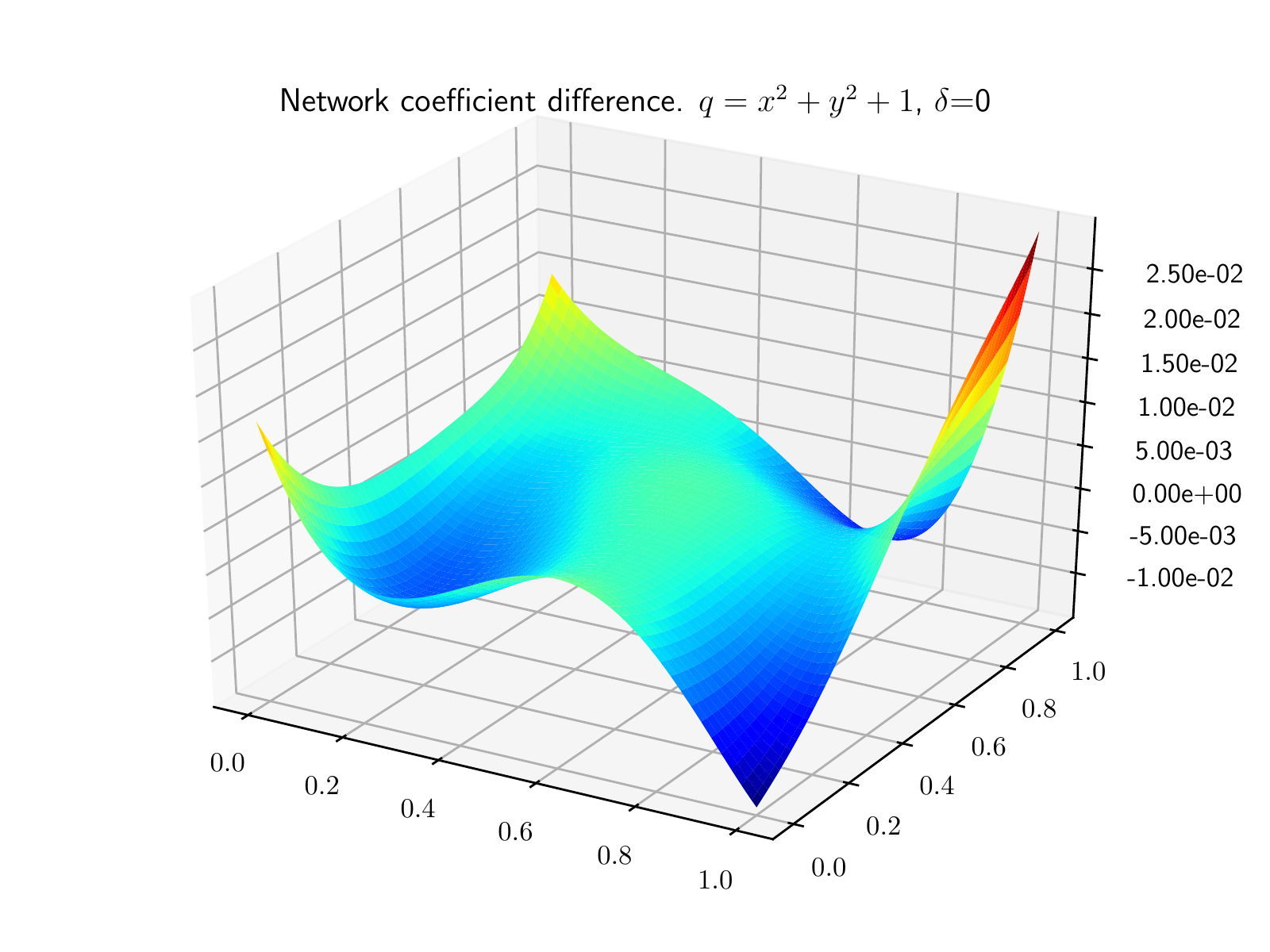}
\caption{The optimized network coefficient without noise.}
\end{subfigure}
\begin{subfigure}[t]{0.49\textwidth}
\centering
\includegraphics[width=\textwidth]{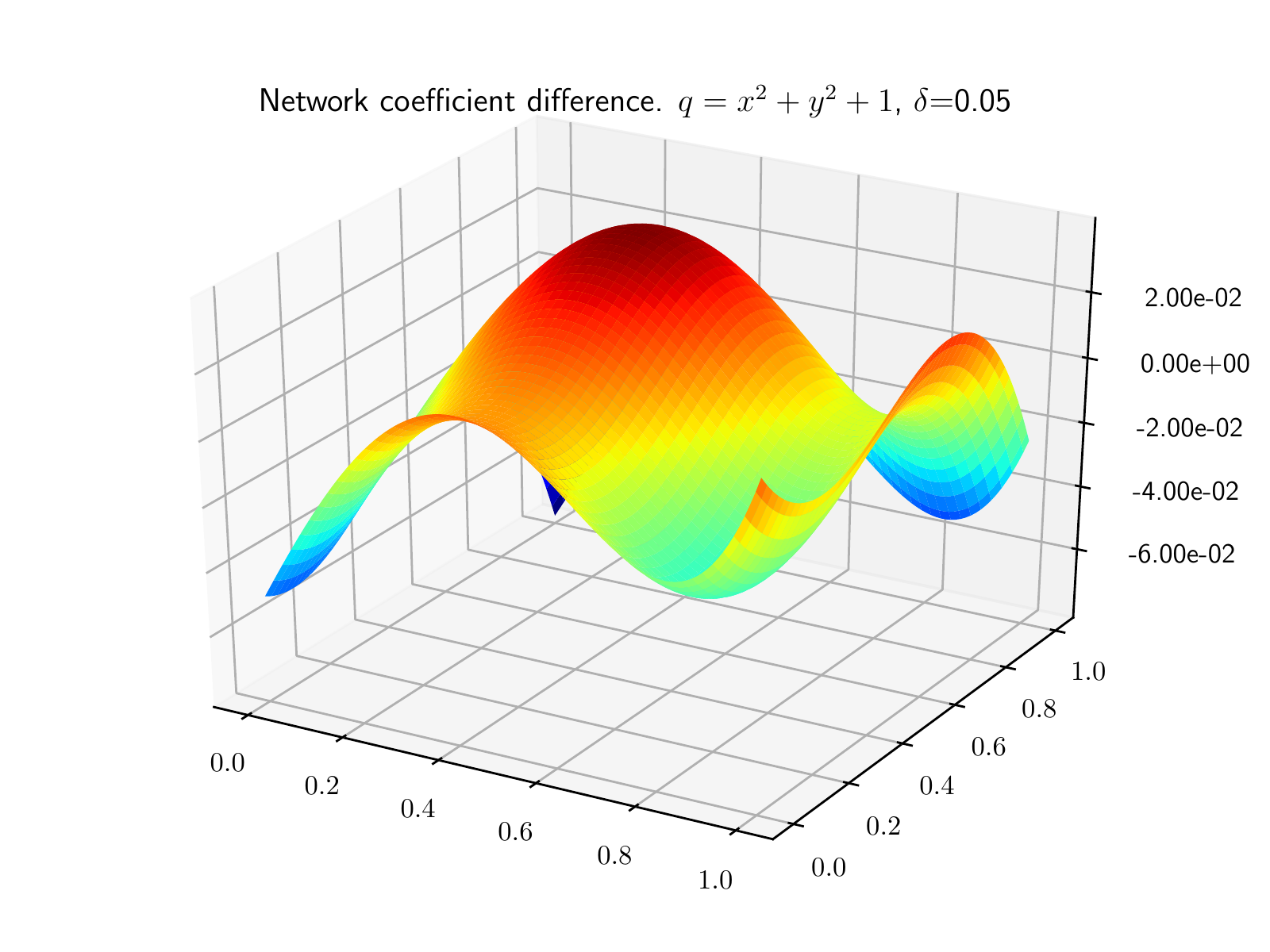}
\caption{The optimized network coefficient with some noise.}
\end{subfigure}
\caption{Difference between the optimized network and exact coefficient when $\hat{q}=1 + x^2 + y^2$.}
\label{heat2dsquareqquadfigs}
\end{figure}

\begin{figure}[htp]
\centering
\begin{subfigure}[t]{0.49\textwidth}
\centering
\includegraphics[width=\textwidth]{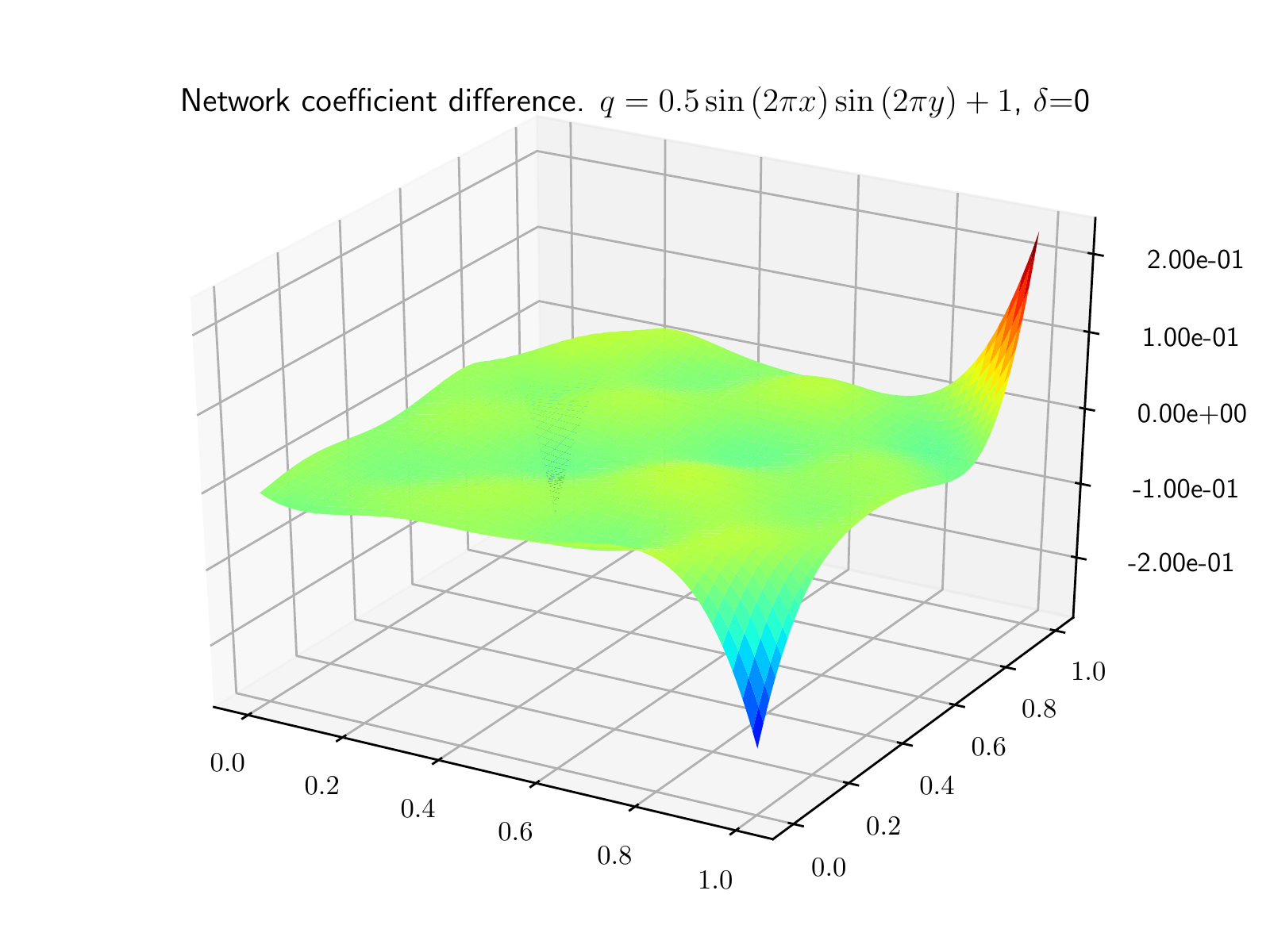}
\caption{The optimized network coefficient without noise.}
\end{subfigure}
\begin{subfigure}[t]{0.49\textwidth}
\centering
\includegraphics[width=\textwidth]{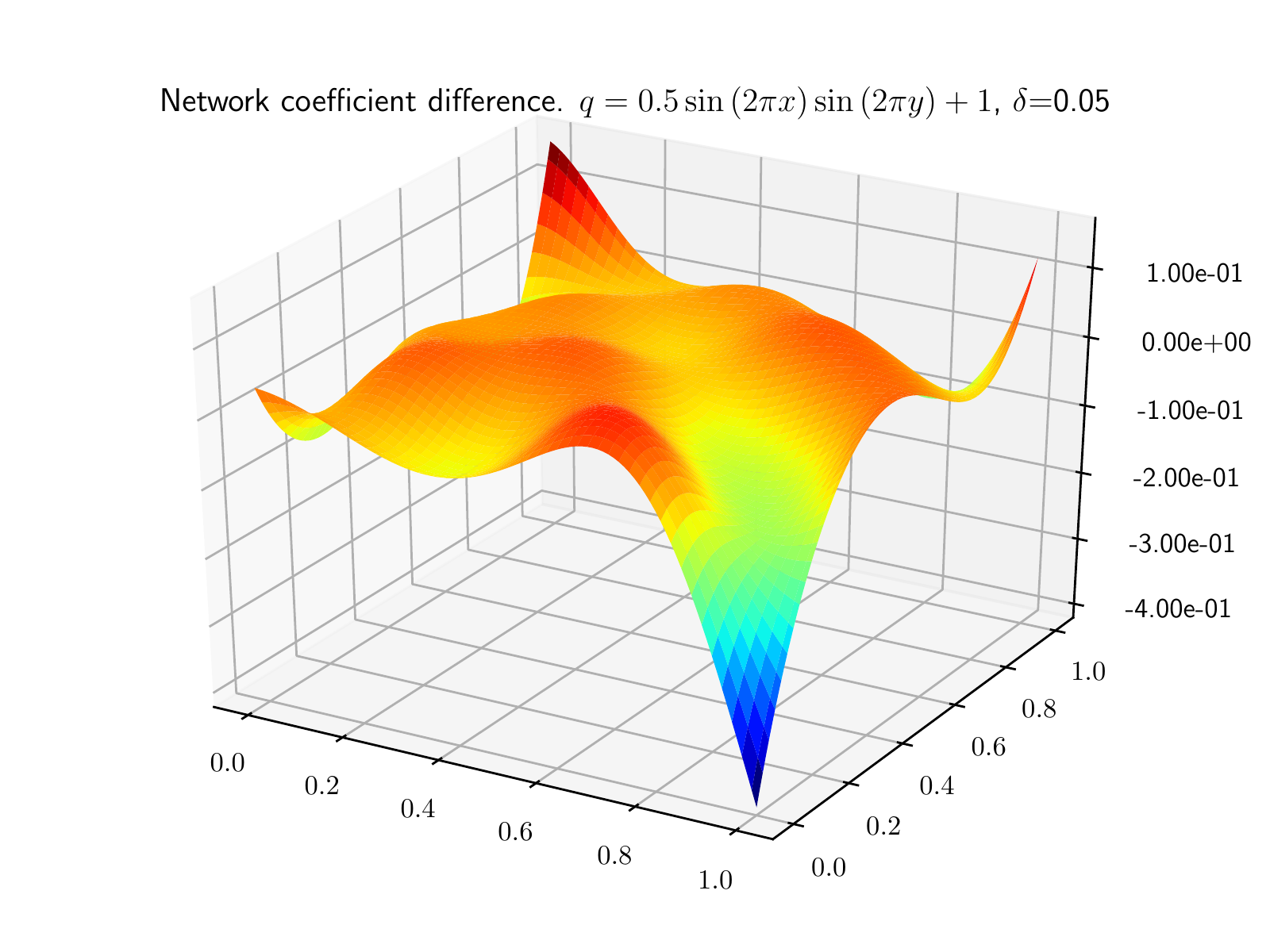}
\caption{The optimized network coefficient with some noise.}
\end{subfigure}
\caption{Difference between the optimized network and exact coefficient when $\hat{q}=1 + 0.5 \sin(2 \pi x) \sin(2 \pi y)$.}
\label{heat2dsquareqsinefigs}
\end{figure}

We plot the difference between the exact and computed coefficient in Figures~\ref{heat2dsquareqconstfigs}--\ref{heat2dsquareqsinefigs} and summarize the results in Table~\ref{heat2dsquaretable}. We can see the same pattern as in the 1D case. The neural network representation is insensitive to noise and we always recover a smooth coefficient even without regularization. Note that the errors are larger at the corners of the domain. This is probably due to the fact that the optimization is performed in the $L^2$ norm which allows larger errors in isolated points. The errors can probably be reduced by performing the optimization under the $H^1$ norm instead. See for example \cite{meshindependencebook} and references therein.

\def\arraystretch{1.2}
\begin{table}[htp]
\centering
\begin{tabular}{|c|c|c|c|c|c|c|c|c|}
\hline
& \multicolumn{8}{|c|}{$\hat{q}=1$} \\
\hline
& \multicolumn{4}{c}{$\delta = 0$} &  \multicolumn{4}{|c|}{$\delta = 0.05$} \\
\hline
& \#it & Time (s) & $||u - \hat{u}||$ & $||q-\hat{q}||$ & \#it & Time (s) & $||u-\hat{u}||$ & $||q-\hat{q}||$ \\
\hline
Network & 41 & 13 & 1.11e-5 & 2.51e-4 & 40 & 12 & 2.35e-4 & 9.28e-4 \\
\hline
\hline
& \multicolumn{8}{|c|}{$\hat{q}=x + y + 1$} \\
\hline
& \multicolumn{4}{c}{$\delta = 0$} &  \multicolumn{4}{|c|}{$\delta = 0.05$} \\
\hline
& \#it & Time (s) & $||u - \hat{u}||$ & $||q-\hat{q}||$ & \#it & Time (s) & $||u-\hat{u}||$ & $||q-\hat{q}||$ \\
\hline
Network & 90 & 24 & 1.46e-4 & 2.99e-3 & 332 & 93 & 1.40e-3 & 1.74e-2 \\
\hline
\hline
& \multicolumn{8}{|c|}{$\hat{q}=x^{2} + y^{2} + 1$} \\
\hline
& \multicolumn{4}{c}{$\delta = 0$} &  \multicolumn{4}{|c|}{$\delta = 0.05$} \\
\hline
& \#it & Time (s) & $||u - \hat{u}||$ & $||q-\hat{q}||$ & \#it & Time (s) & $||u-\hat{u}||$ & $||q-\hat{q}||$ \\
\hline
Network & 402 & 111 & 2.07e-4 & 4.22e-3 & 517 & 138 & 1.76e-3 & 2.04e-2 \\
\hline
\hline
& \multicolumn{8}{|c|}{$\hat{q}=0.5 \sin{\left (2 \pi x \right )} \sin{\left (2 \pi y \right )} + 1$} \\
\hline
& \multicolumn{4}{c}{$\delta = 0$} &  \multicolumn{4}{|c|}{$\delta = 0.05$} \\
\hline
& \#it & Time (s) & $||u - \hat{u}||$ & $||q-\hat{q}||$ & \#it & Time (s) & $||u-\hat{u}||$ & $||q-\hat{q}||$ \\
\hline
Network & 2112 & 560 & 6.32e-4 & 1.86e-2 & 1553 & 406 & 3.09e-3 & 4.48e-2 \\
\hline
\end{tabular}
\caption{Summary for the neural network representation of the coefficient for the 2D Poisson equation. \#it is the number of iterations. The norm is measured as the integral of a high-order interpolation onto the finite element space as provided by the \texttt{errornorm} function in \texttt{FEniCS}, and $\hat{u}$ is the unperturbed exact solution in the case of added noise.}
\label{heat2dsquaretable}
\end{table}

\subsubsection{Mesh independent convergence}
It was shown in \cite{inversemeshdep} that the number of optimization iterations grows polynomially with the ratio between the volumes of the smallest and largest mesh elements, $h_{\max}/h_{\min}$. The reason is that most gradient based optimization libraries do not take the underlying function space inner product into account when computing the step size and direction. As the neural network augmentation is mesh independent, we expect that the number of iterations remains constant when locally refining the mesh, as long as the mesh is fine enough to accurately compute the solution of the forward problem.

\begin{figure}[htp]
\centering
\begin{subfigure}[t]{0.49\textwidth}
\centering
\includegraphics[width=\textwidth]{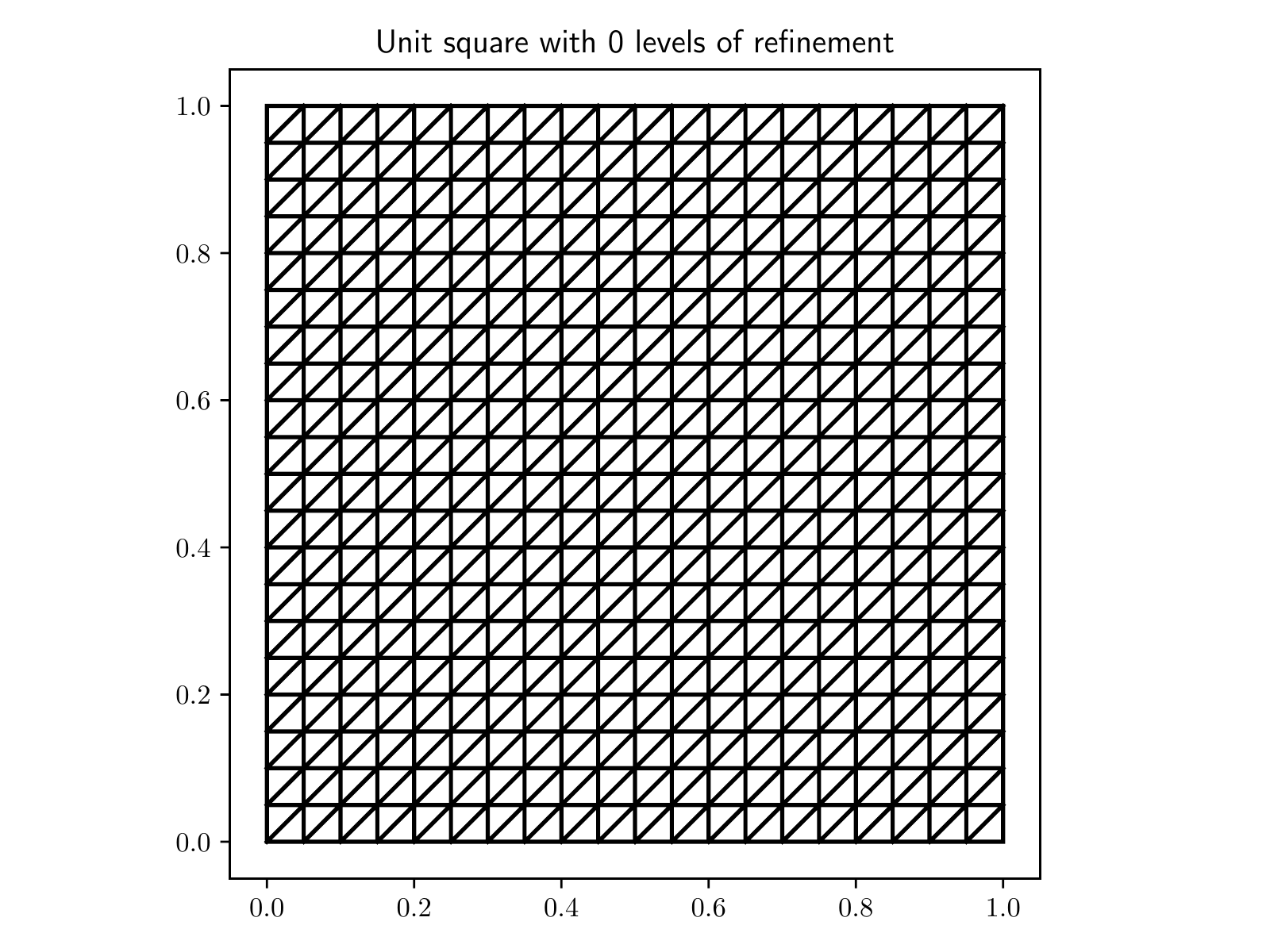}
\caption{Initial uniform mesh}
\end{subfigure}
\begin{subfigure}[t]{0.49\textwidth}
\centering
\includegraphics[width=\textwidth]{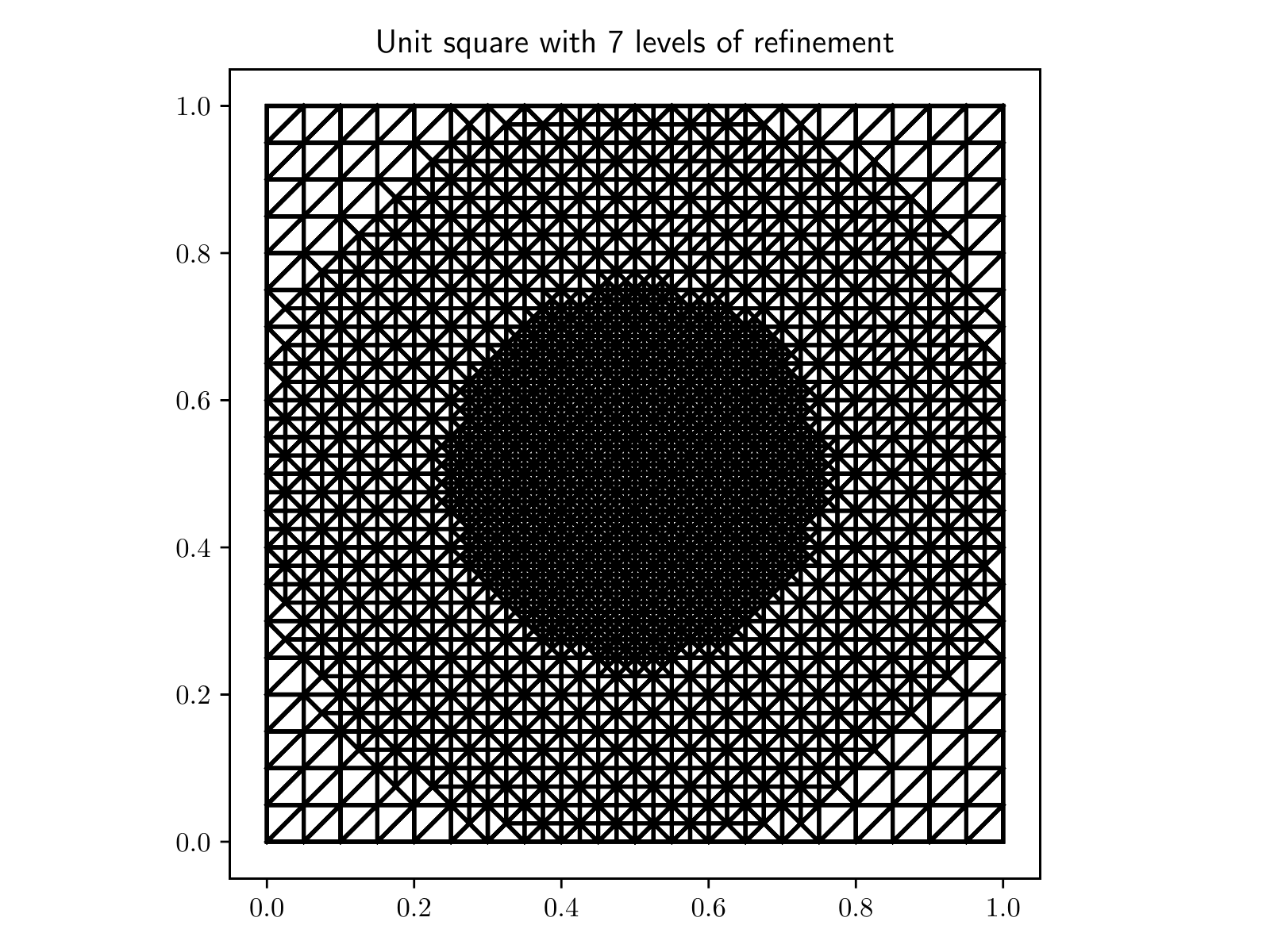}
\caption{Final refined mesh.}
\end{subfigure}
\caption{Concentric mesh refinement.}
\label{heat2dmeshrefinefigure}
\end{figure}

To test the mesh independent convergence, we consider a simple test case. We solve the inverse problem for the Poisson equation \eqref{heat2d}, on the unit square, with coefficient $\hat{q} = 1$ and the exact solution given by \ref{heat2dsol} without any added noise. We locally refine the mesh in concentric circles around the center of the domain with radius $r(n) = 0.5/n$, for $n = 1, \ldots, 7$, as can be seen in Figure~\ref{heat2dmeshrefinefigure}. The results are summarized in Table~\ref{heat2dmeshrefinetable}. We can clearly see that the number of iterations required until convergence as the mesh is refined becomes constant, and a stable minimum is found.

\def\arraystretch{1.2}
\begin{table}[htp]
\centering
{\small
\begin{tabular}{|c|c|c|c|c|c|c|c|c|}
\hline
$n$ & 0 & 1 & 2 & 3 & 4 & 5 & 6 & 7 \\
\hline
\#dofs & 441 & 725 & 1017 & 1479 & 2482 & 5040 & 11922 & 32000 \\
\hline
$h_{\max}/h_{\min}$ & 1 & 2 & 4 & 8 & 16 & 32 & 64 & 128 \\
\hline
\#it & 32 & 35 & 40 & 40 & 40 & 40 & 40 & 40 \\
\hline
$||q-\hat{q}||$ & 6.14e-3 & 4.91e-3 & 4.85e-3 & 4.84e-3 & 4.84e-3 & 4.84e-3 & 4.84e-3 & 4.84e-3 \\
\hline
\end{tabular}
}
\caption{Number of iterations for different levels of refinement.}
\label{heat2dmeshrefinetable}
\end{table}

\subsection{Three-dimensional Poisson}
\begin{figure}[htp]
\centering
\begin{subfigure}[t]{0.45\textwidth}
\centering
\includegraphics[width=\textwidth]{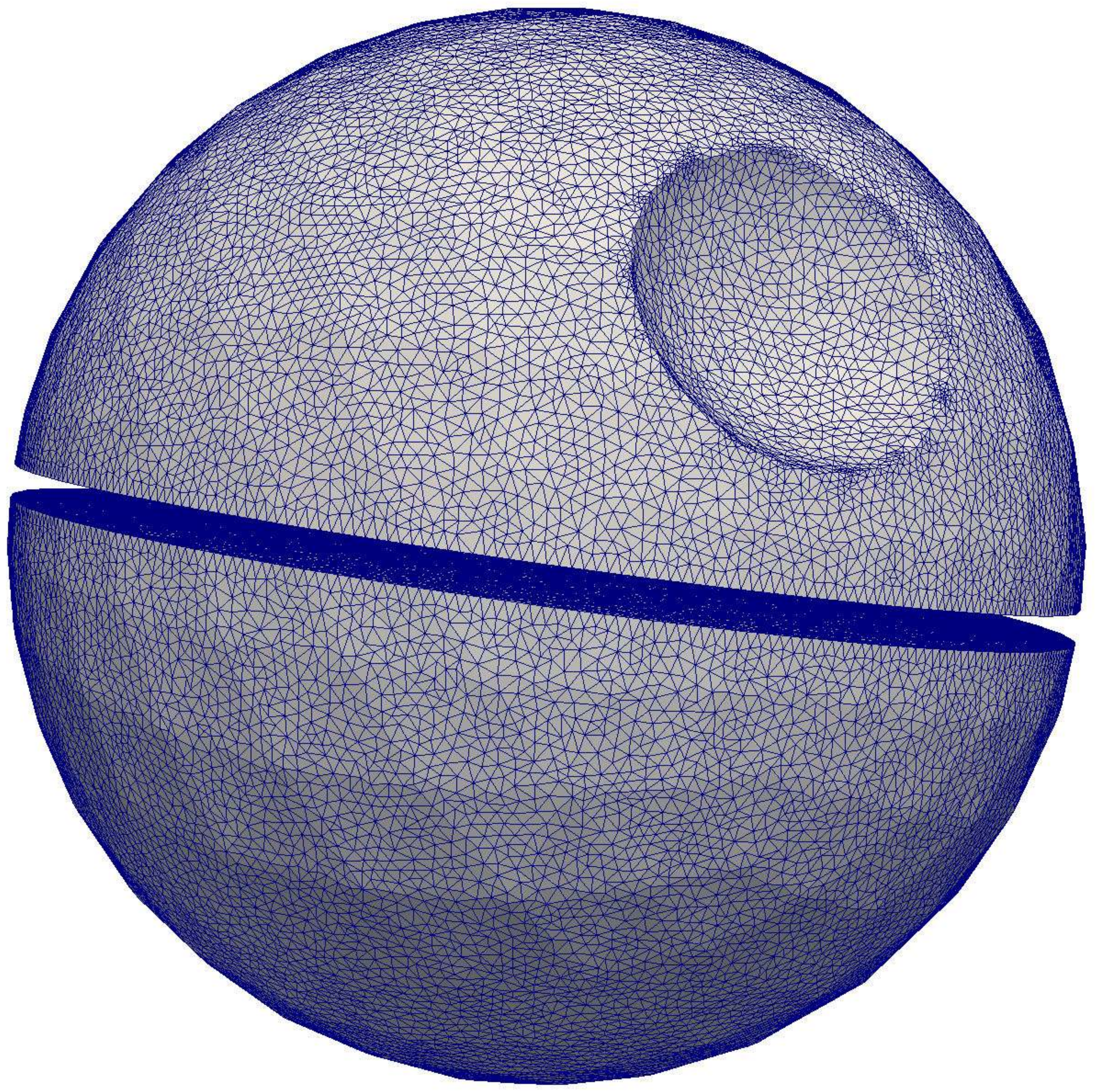}
\caption{Domain and surface mesh.}
\end{subfigure}
\begin{subfigure}[t]{0.535\textwidth}
\centering
\includegraphics[width=\textwidth]{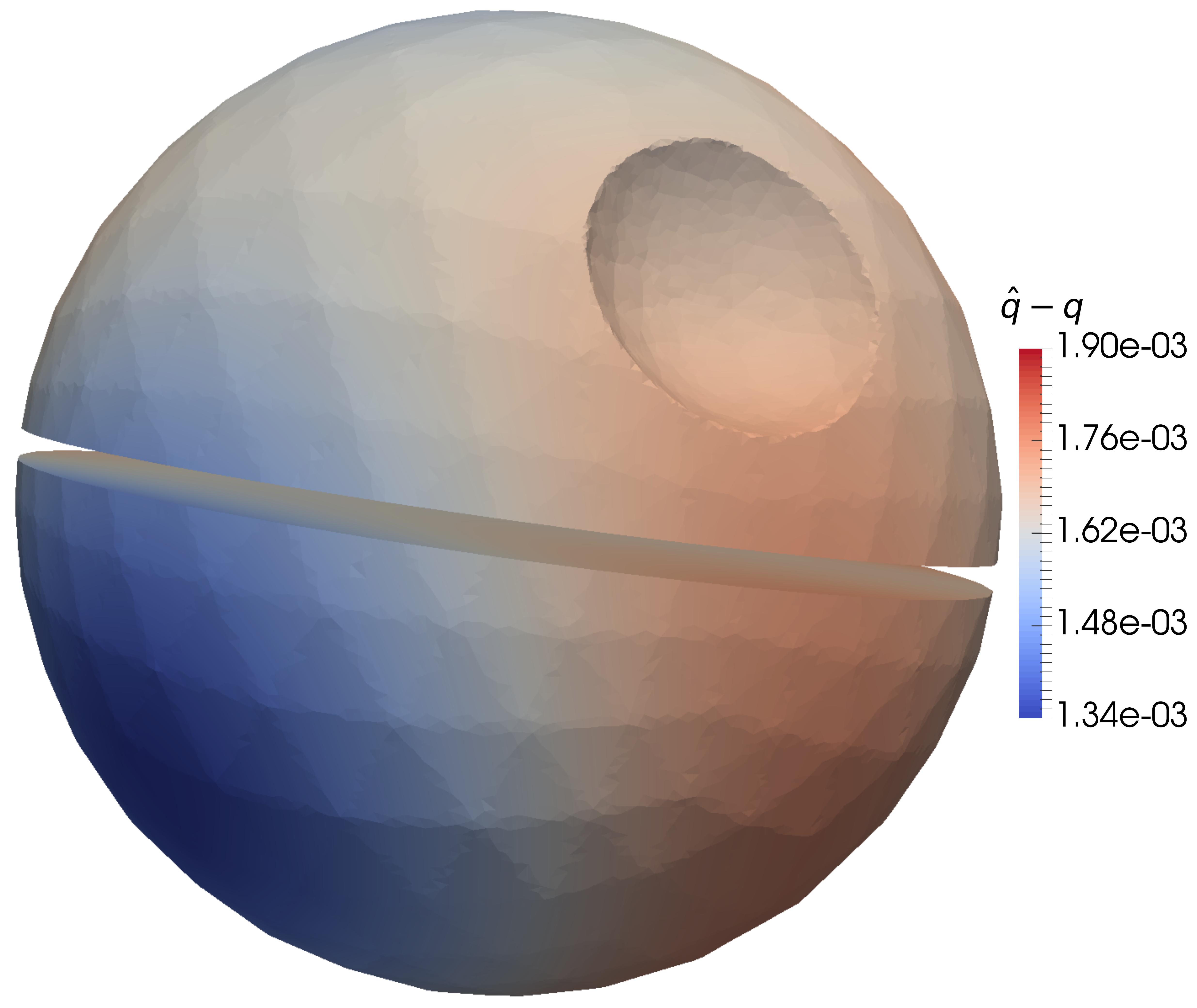}
\caption{Coefficient difference.}
\end{subfigure}
\caption{The 3D Poisson equation on a complex geometry with 362172 degrees of freedom.}
\label{deathstarfigure}
\end{figure}

As a final example, we consider the three-dimensional Poisson equation \eqref{heat2d} on a complex geometry from \cite{deathstar}. We take the exact coefficient $\hat{q} = 1$ and the analytical solution
\begin{equation}
u = \sin(\pi x) \sin(\pi y) \sin(\pi z)
\end{equation}
with no noise and compute the inverse problem. We use a network with 10 neurons in the single hidden layer which gives a total of 51 optimization parameters. Convergence was achieved after 41 BFGS iterations with gradient norm tolerance $10^{-7}$, which is consistent with the result in Table~\ref{heat2dmeshrefinetable}. This indicates that the neural network augmentation method is also independent of the geometry. The domain and difference between the exact and computed network coefficient can be seen in Figure~\ref{deathstarfigure}.

\begin{remark}
The extra BFGS iteration required in the 3D case is because the forward problem could not be solved by a direct method due to memory requirements. We used the GMRES iterative method with an incomplete LU preconditioner to solve the forward problem. The solution to the forward problem is thus slightly less accurate due to the tolerance settings of the iterative method, which in turn means that the computed error functional gradients are also less accurate and more BFGS iterations are required.
\end{remark}

\section{Severely ill-posed examples} \label{severely}
In the previous section we discussed moderately ill-posed problems where we have measurements of the smooth coefficients in the whole domain. Here we will show some examples of severely ill-posed problems where we have discontinuous coefficients or incomplete measurements.

\subsection{1D discontinuous coefficient} \label{discont}
\begin{figure}[htp]
\centering
\begin{subfigure}[t]{0.45\textwidth}
\centering
\includegraphics[width=\textwidth]{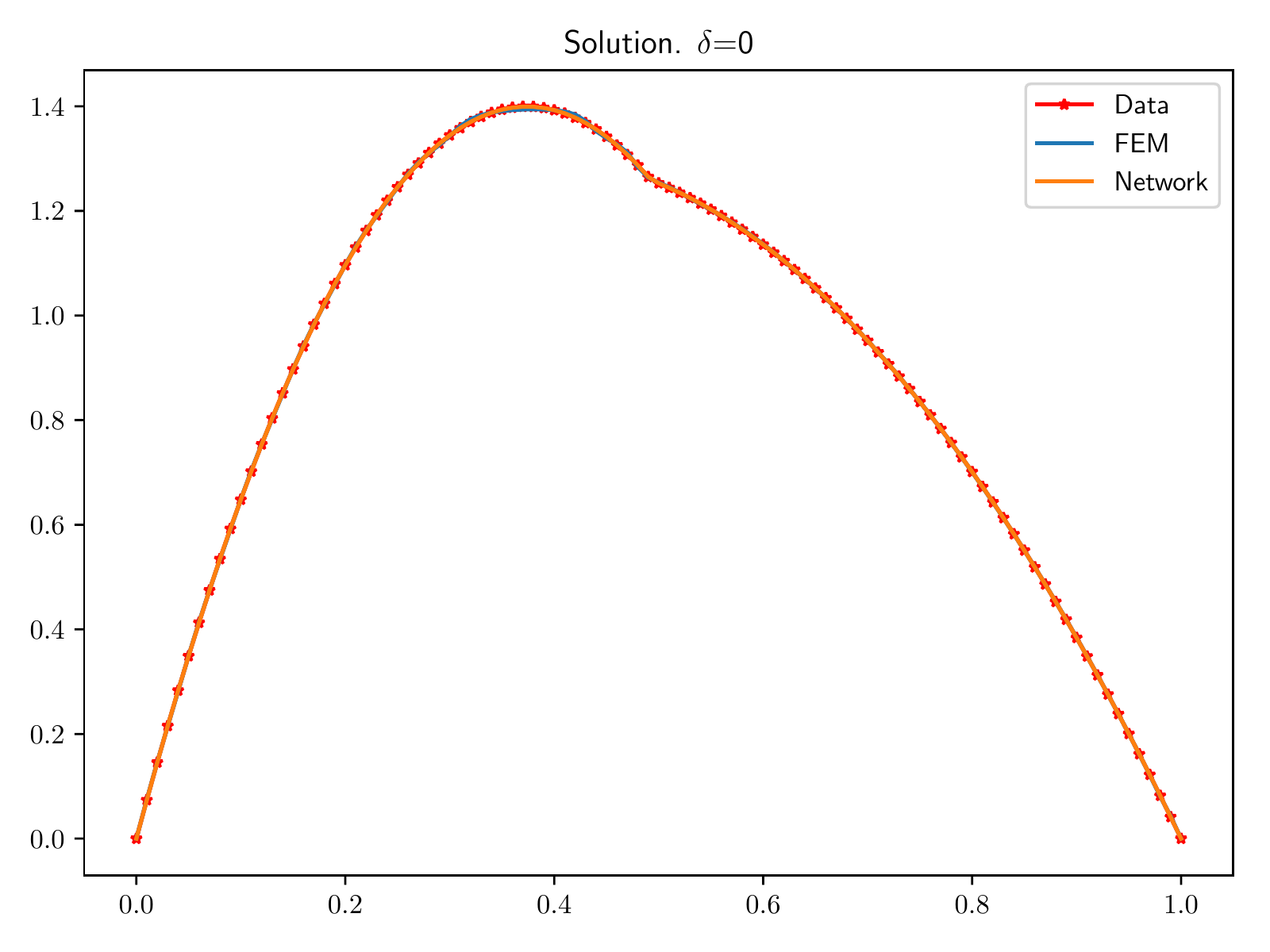}
\caption{Solutions with the optimized coefficients and no noise.}
\end{subfigure}
\begin{subfigure}[t]{0.45\textwidth}
\centering
\includegraphics[width=\textwidth]{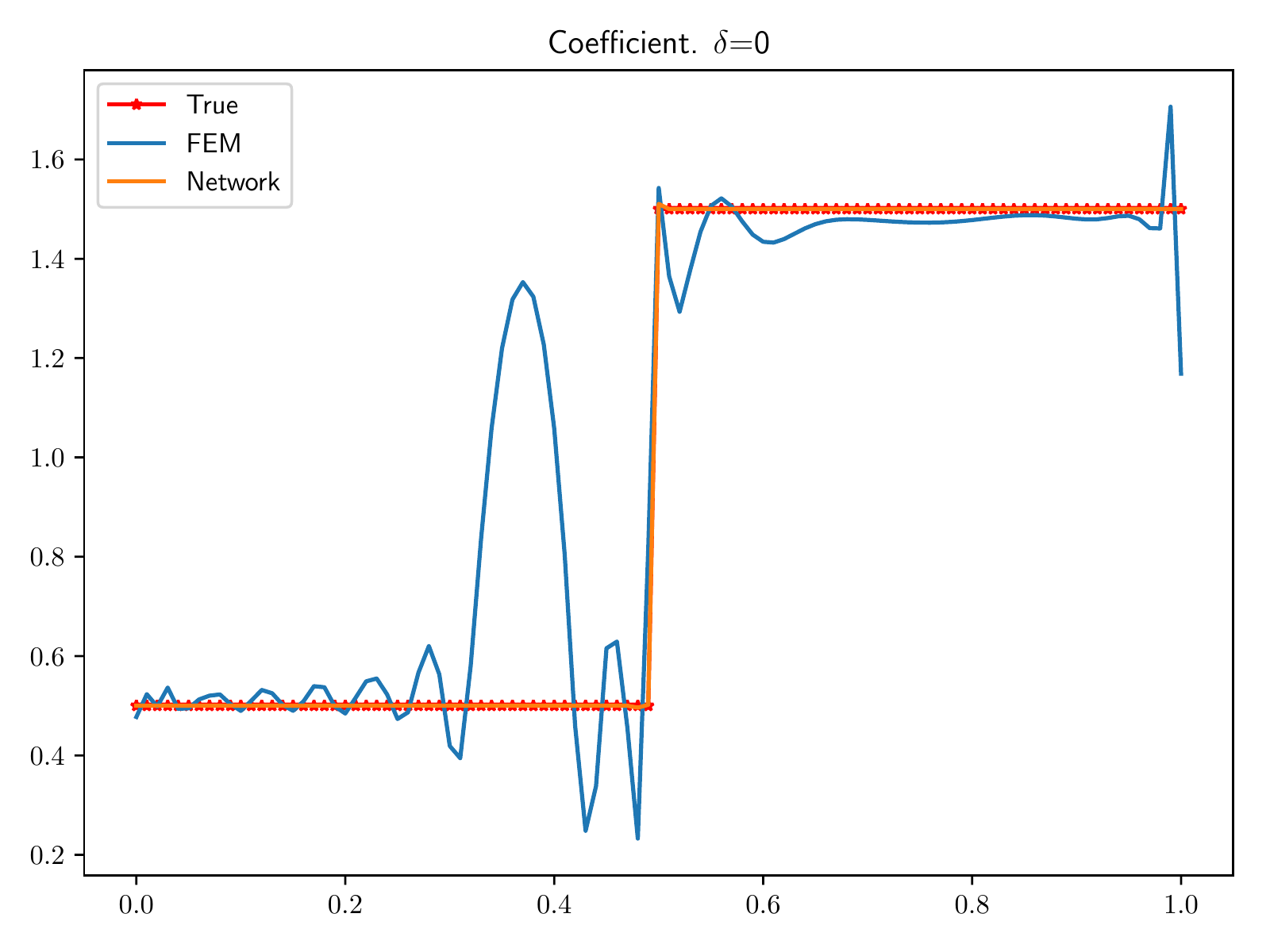}
\caption{The optimized coefficients without noise.}
\end{subfigure}
\begin{subfigure}[t]{0.45\textwidth}
\centering
\includegraphics[width=\textwidth]{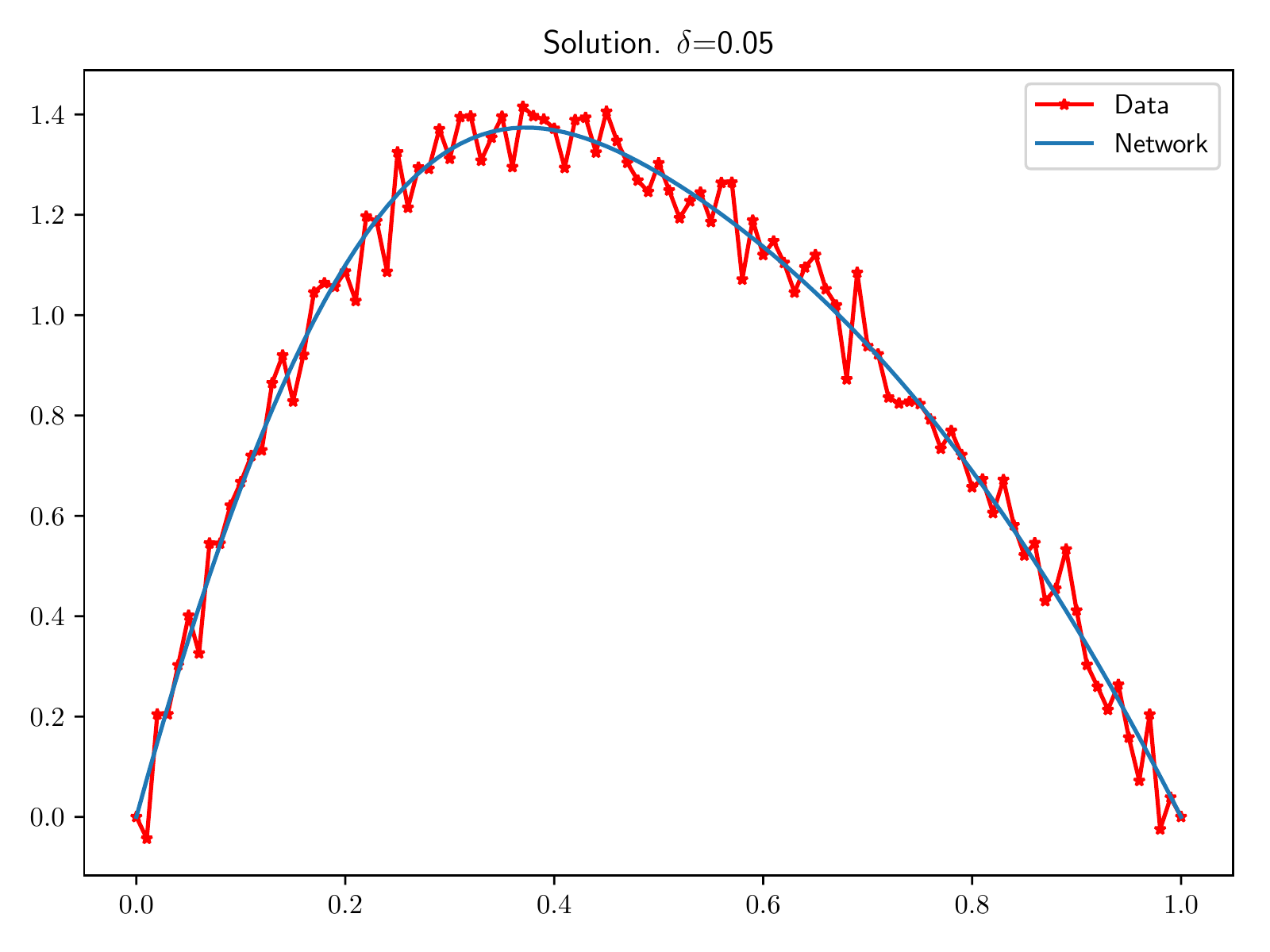}
\caption{Network solution with the optimized coefficient and 5\% noise level.}
\end{subfigure}
\begin{subfigure}[t]{0.45\textwidth}
\centering
\includegraphics[width=\textwidth]{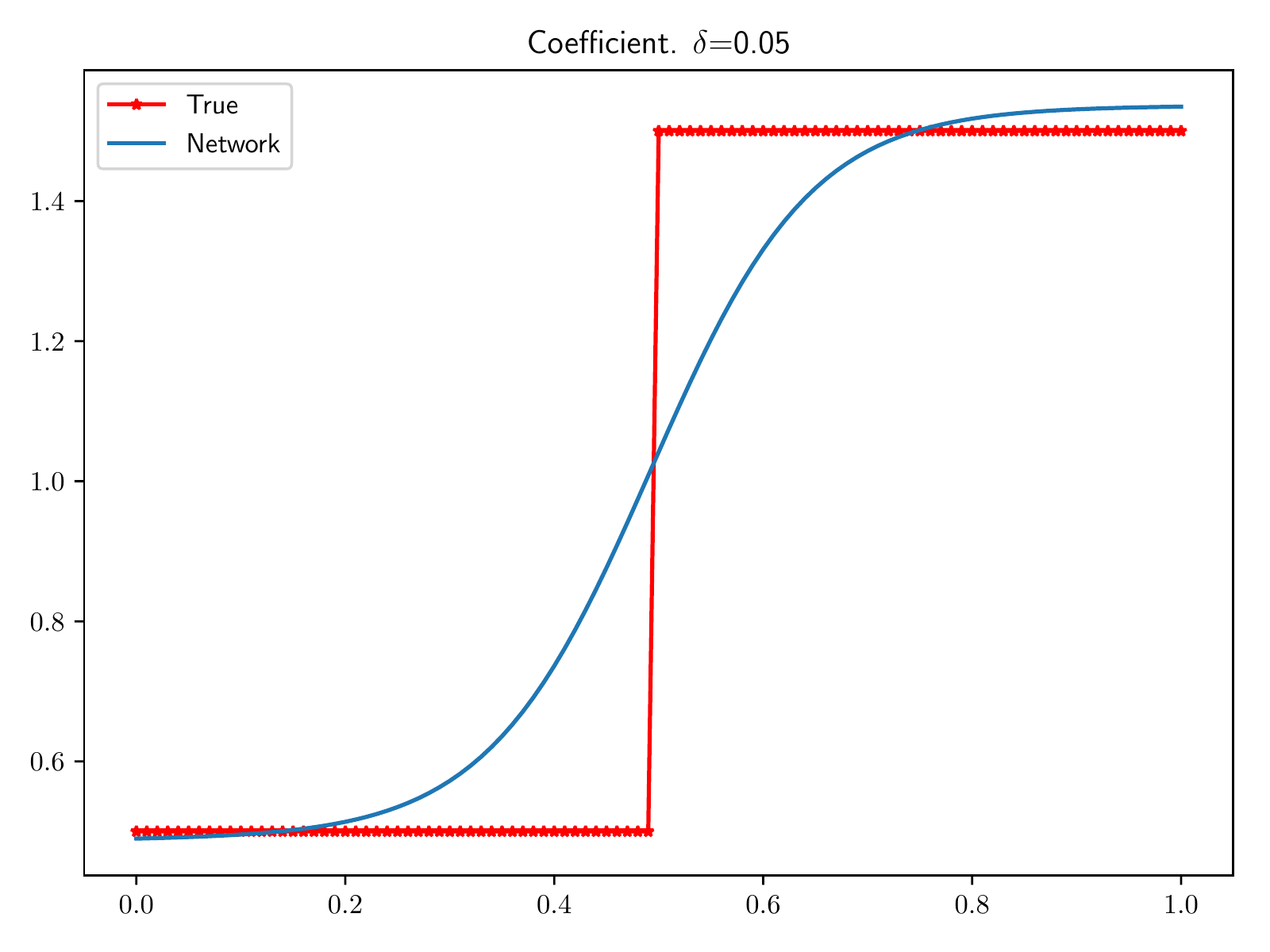}
\caption{The optimized network coefficient and 5\% noise level.}
\end{subfigure}
\begin{subfigure}[t]{0.45\textwidth}
\centering
\includegraphics[width=\textwidth]{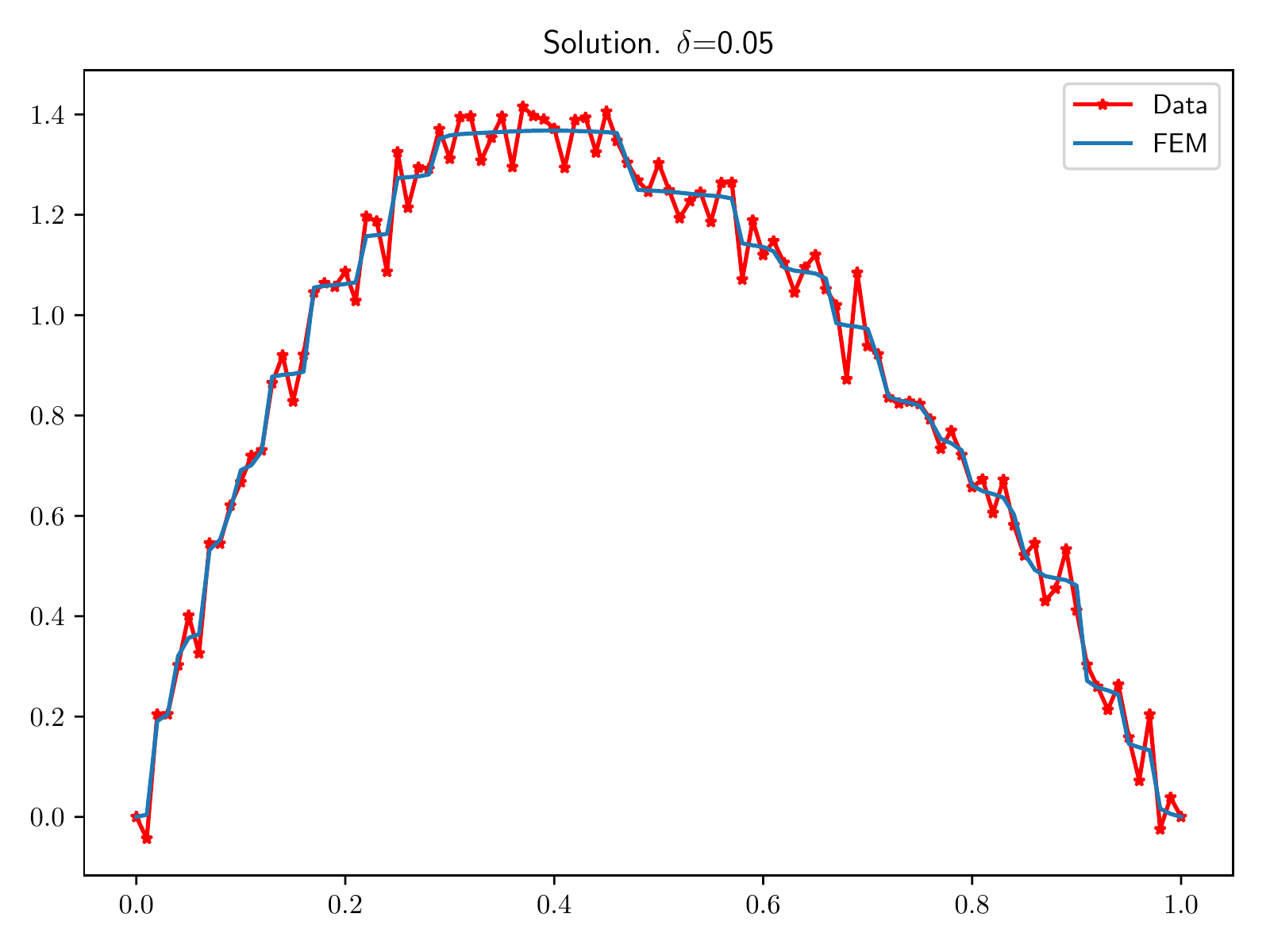}
\caption{FEM solution with the optimized coefficient and 5\% noise level.}
\end{subfigure}
\begin{subfigure}[t]{0.45\textwidth}
\centering
\includegraphics[width=\textwidth]{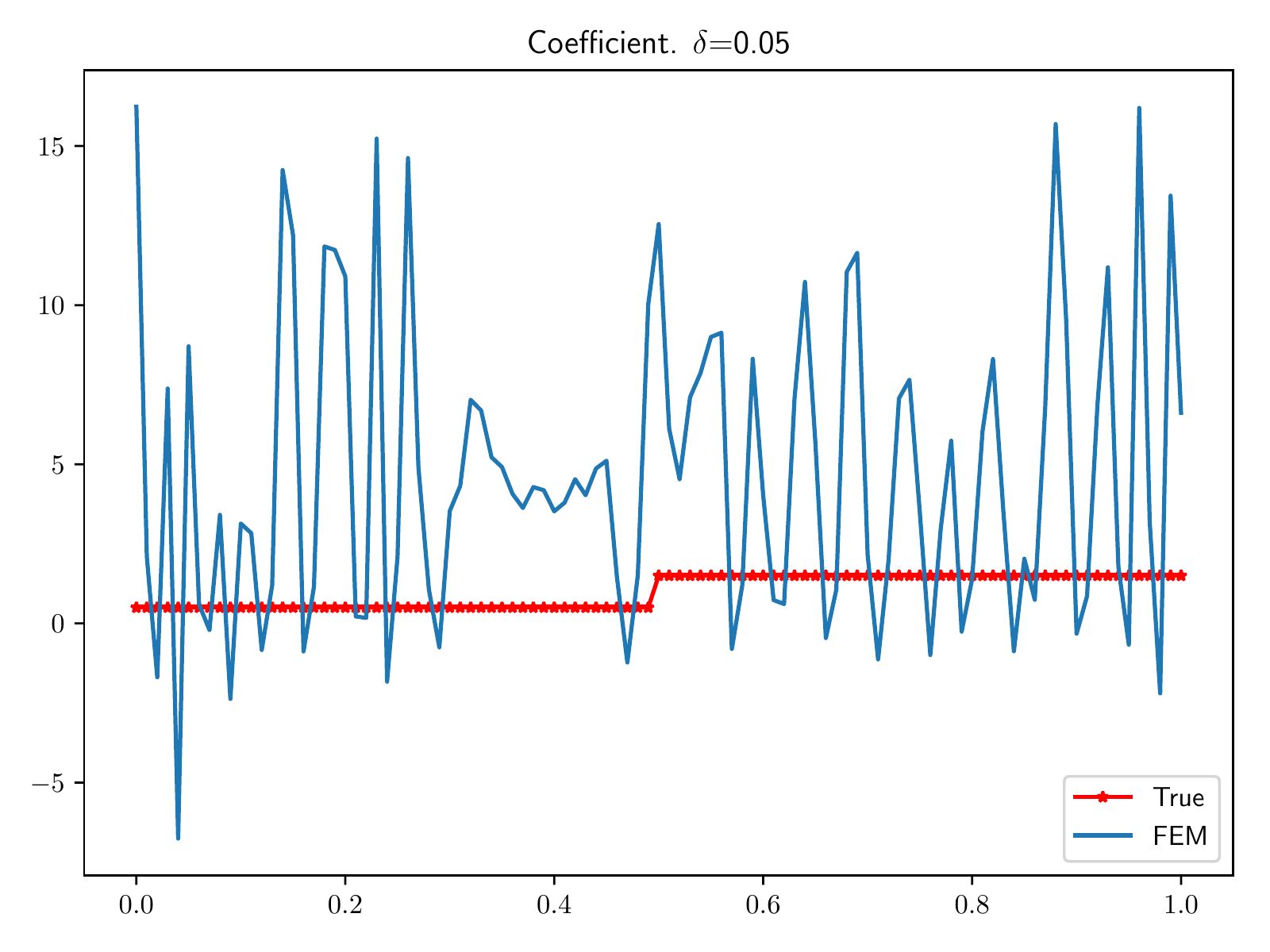}
\caption{The optimized FEM coefficient and 5\% noise level.}
\end{subfigure}
\caption{Comparison between FEM and a neural network with a discontinuous coefficient.}
\label{heat1dqdiscontfigs}
\end{figure}

Here we repeat the examples in section~\ref{secheat1d} for the Poisson problem \eqref{heat1d} with
\begin{equation}
\begin{aligned}
q = \begin{cases}
0.5, & 0 \leq x < 0.5 \\
1.5, & 0.5 \leq x \leq 1
\end{cases}
, &&
f = 10, && g_{1} = g_2 = 0.
\end{aligned}
\end{equation}
In this case, the analytical solution is not known and we compute the reference solution and data using FEM. We used the same small network with a single hidden layer with 3 neurons and the sigmoid activation function. The results can be seen in Figure~\ref{heat1dqdiscontfigs} and Table~\ref{heat1dqdisconttable}. The network captures the discontinuity perfectly for noiseless data, and the smoothing effect of the implicit regularization is clearly seen when noise is added.

\def\arraystretch{1.2}
\begin{table}[htp]
\begin{tabular}{|c|c|c|c|c|c|c|c|c|}
\hline
& \multicolumn{8}{|c|}{Discontinuous coefficient} \\
\hline
& \multicolumn{4}{c}{$\delta = 0$} &  \multicolumn{4}{|c|}{$\delta = 0.05$} \\
\hline
& \#it & Time (s) & $||u - \hat{u}||$ & $||q-\hat{q}||$ & \#it & Time (s) & $||u-\hat{u}||$ & $||q-\hat{q}||$ \\
\hline
Network & 310 & 37 & 1.19e-05 & 9.33e-04 & 346 & 118 & 1.25e-02 & 1.64e-01 \\
FEM & 332 & 76 & 1.58e-03 & 2.07e-01 & 788 & 408 & 2.90e-02 & 5.34e+00 \\
\hline
\end{tabular}
\caption{Comparison between FEM and a neural network with a discontinuous coefficient.}
\label{heat1dqdisconttable}
\end{table}

\subsection{1D incomplete data} \label{incomplete}
\begin{figure}[htp]
\centering
\begin{subfigure}[t]{0.45\textwidth}
\centering
\includegraphics[width=\textwidth]{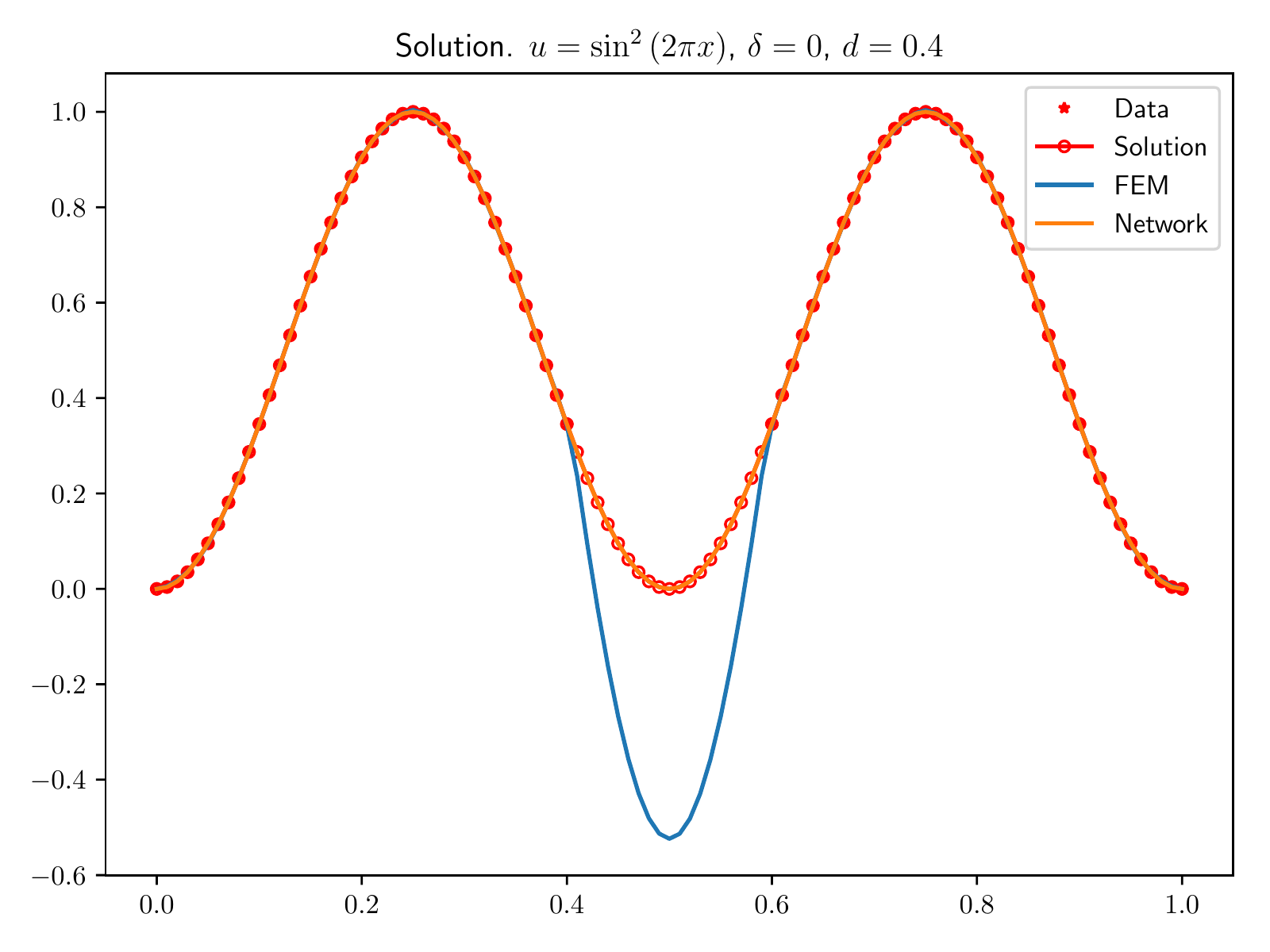}
\caption{Solutions with the optimized coefficients and no noise.}
\end{subfigure}
\begin{subfigure}[t]{0.45\textwidth}
\centering
\includegraphics[width=\textwidth]{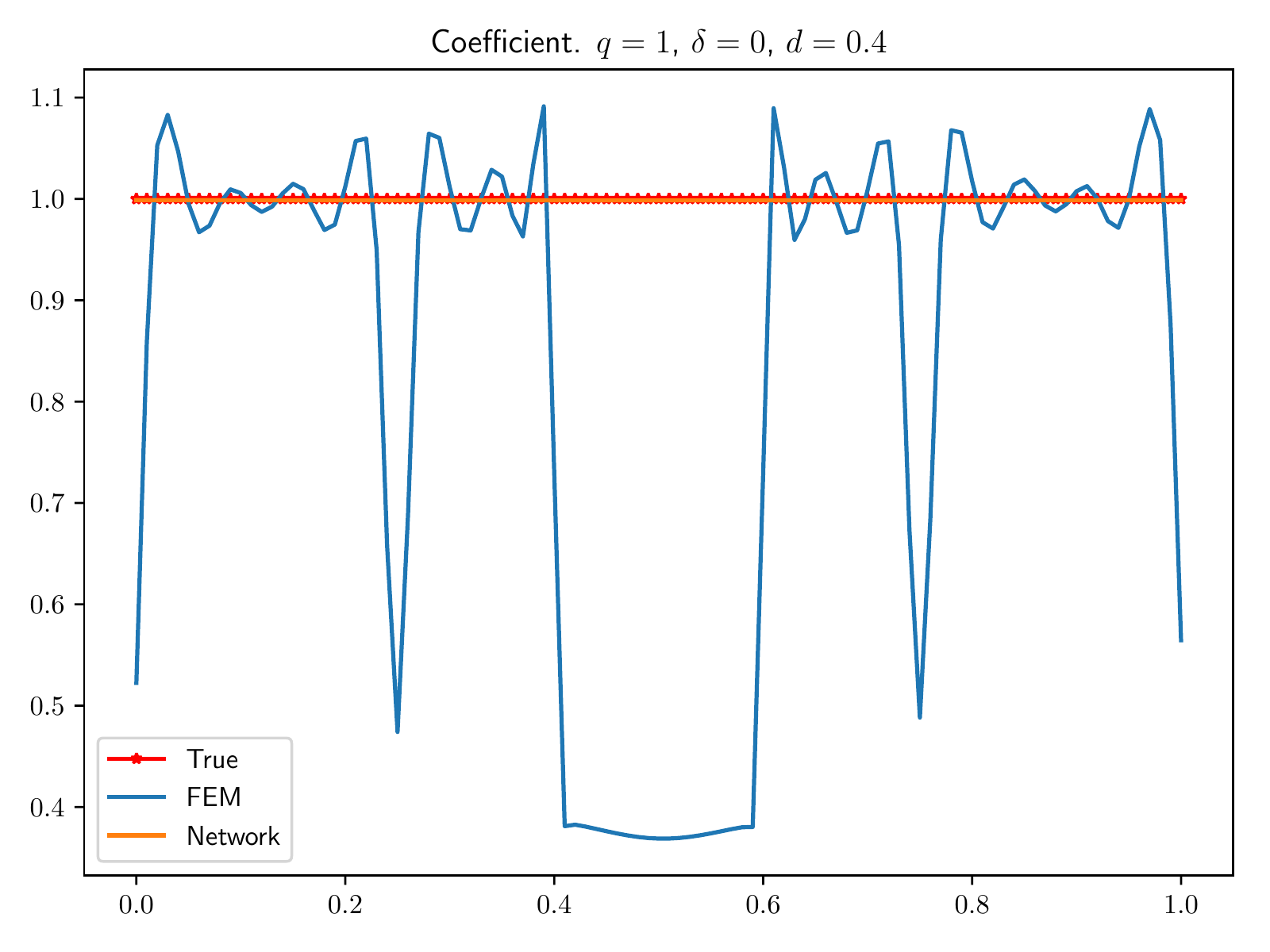}
\caption{The optimized coefficients without noise.}
\end{subfigure}
\begin{subfigure}[t]{0.45\textwidth}
\centering
\includegraphics[width=\textwidth]{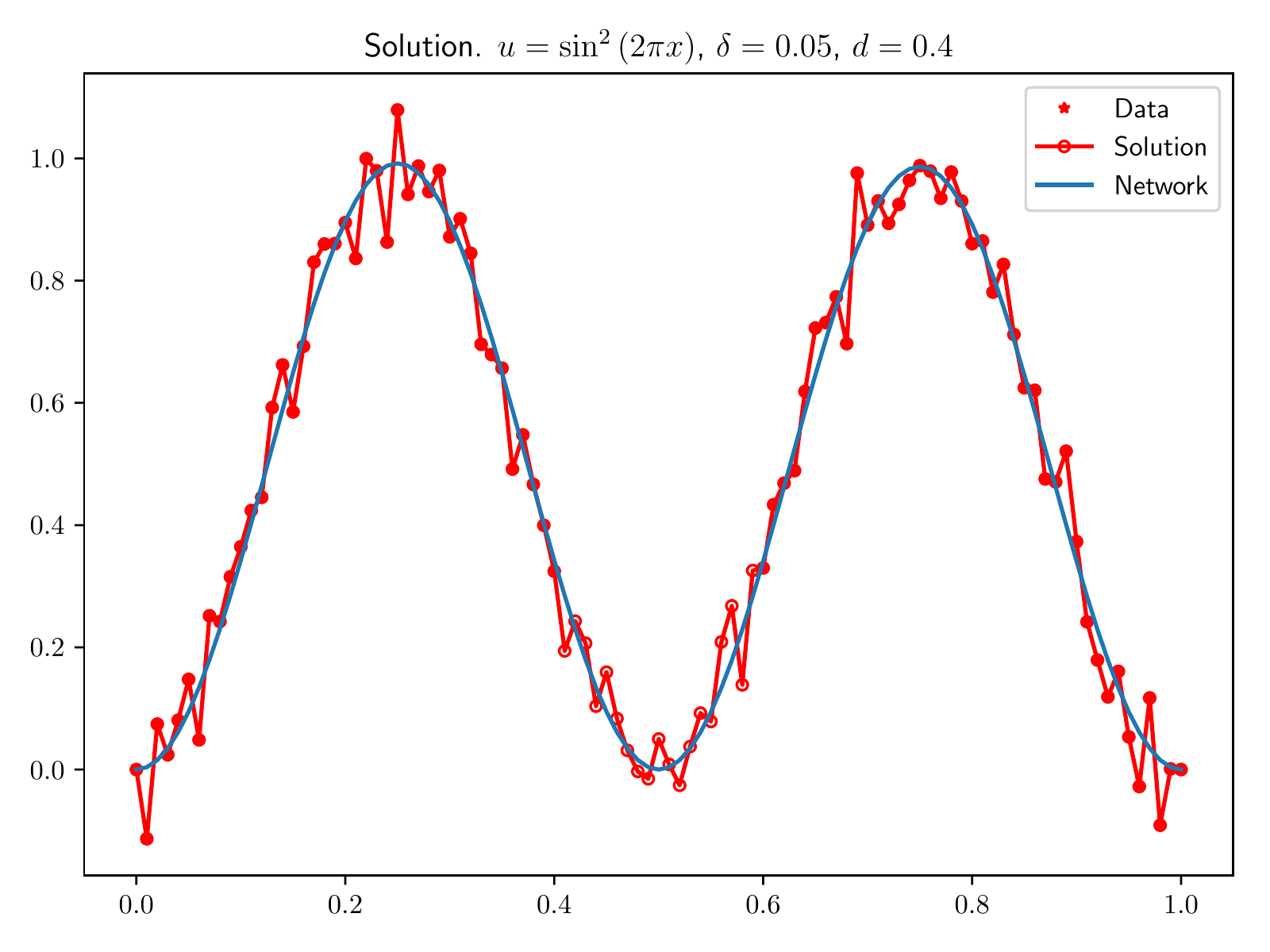}
\caption{Network solution with the optimized coefficient and 5\% noise level.}
\end{subfigure}
\begin{subfigure}[t]{0.45\textwidth}
\centering
\includegraphics[width=\textwidth]{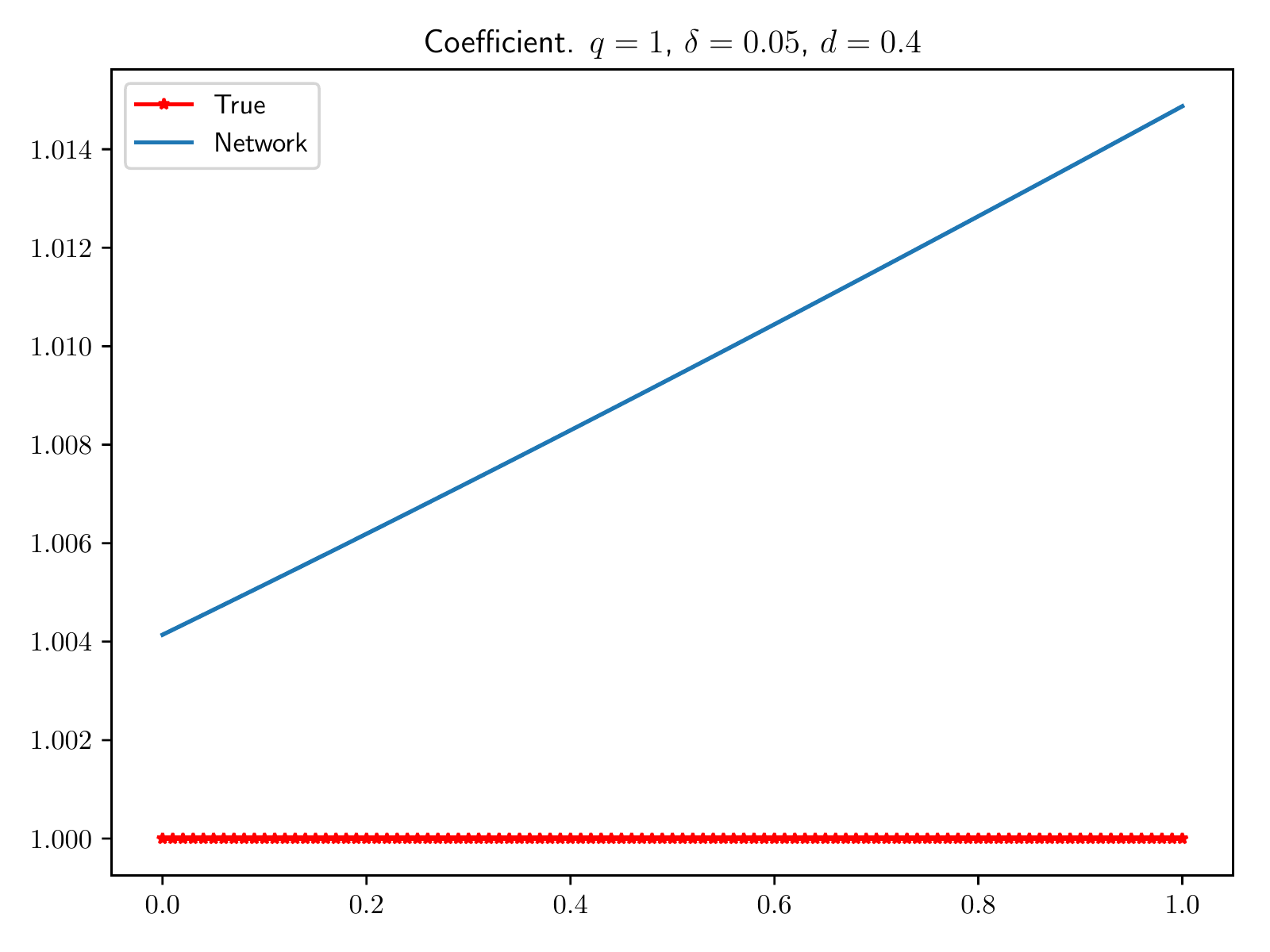}
\caption{The optimized network coefficient and 5\% noise level.}
\end{subfigure}
\begin{subfigure}[t]{0.45\textwidth}
\centering
\includegraphics[width=\textwidth]{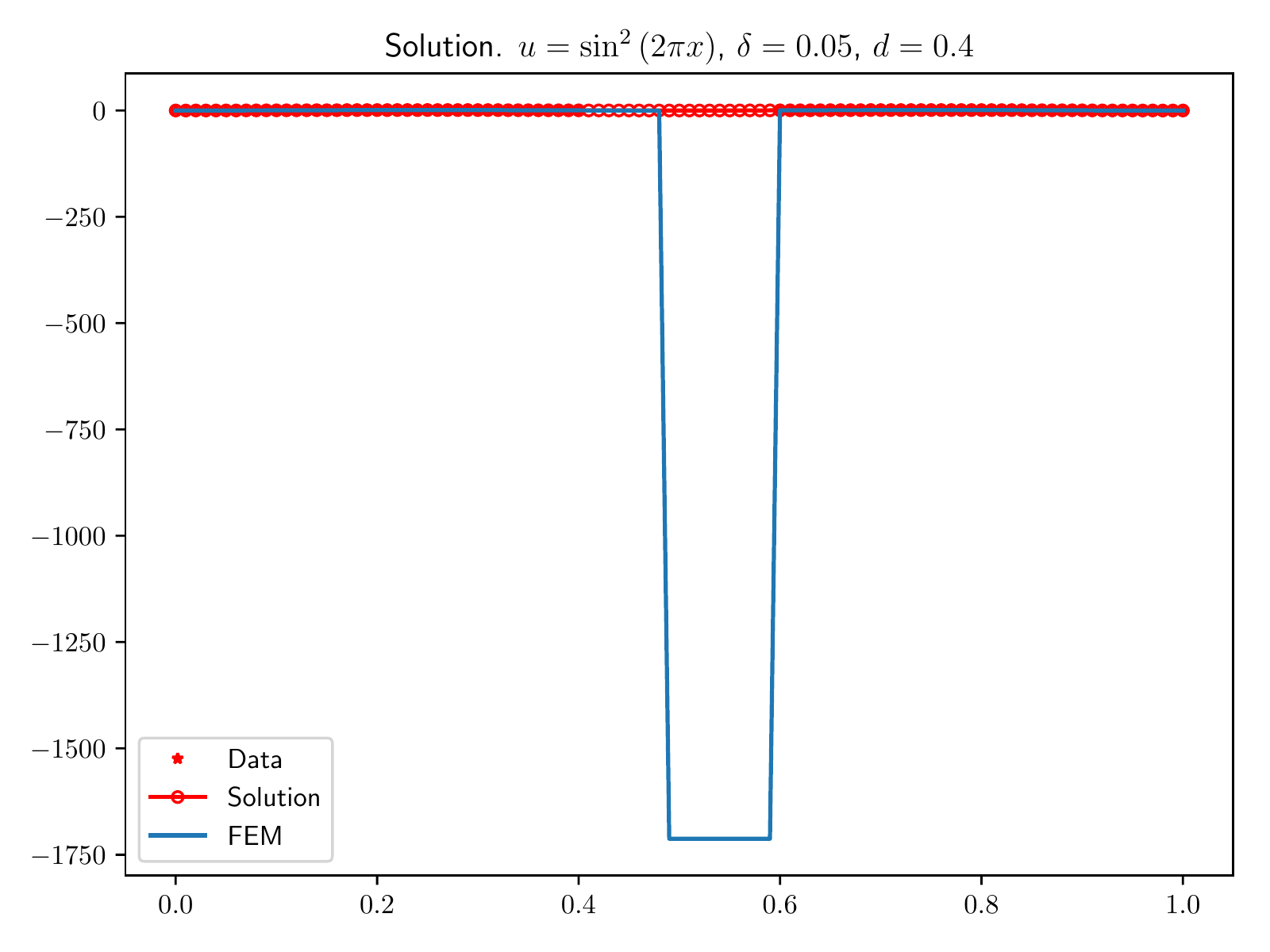}
\caption{FEM solution with the optimized coefficient and 5\% noise level.}
\end{subfigure}
\begin{subfigure}[t]{0.45\textwidth}
\centering
\includegraphics[width=\textwidth]{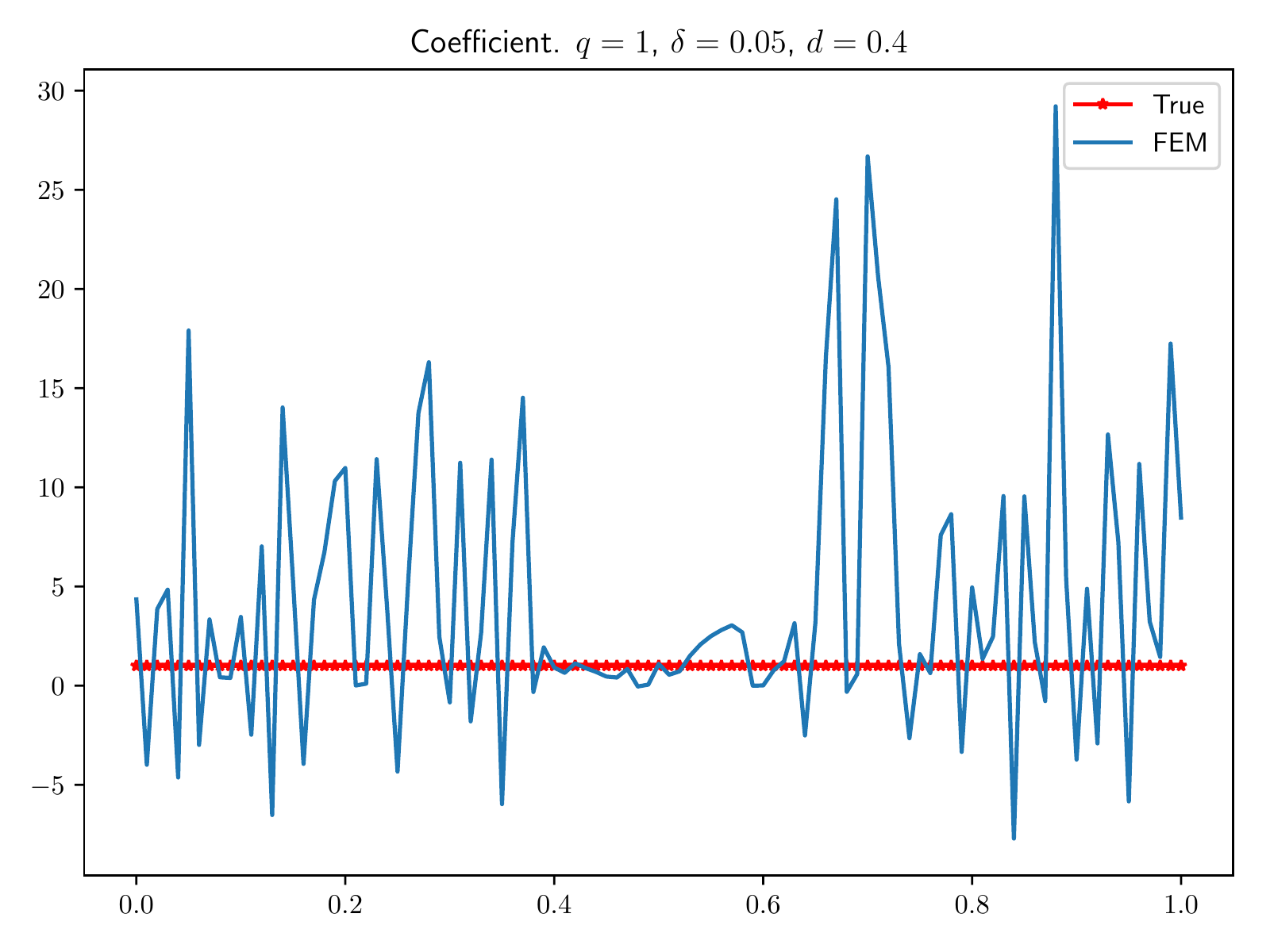}
\caption{The optimized FEM coefficient and 5\% noise level.}
\end{subfigure}
\caption{Comparison between FEM and a neural network with a constant coefficient and $d=0.4$.}
\label{heat1dqconstinc1figs}
\end{figure}

\begin{figure}[htp]
\centering
\begin{subfigure}[t]{0.45\textwidth}
\centering
\includegraphics[width=\textwidth]{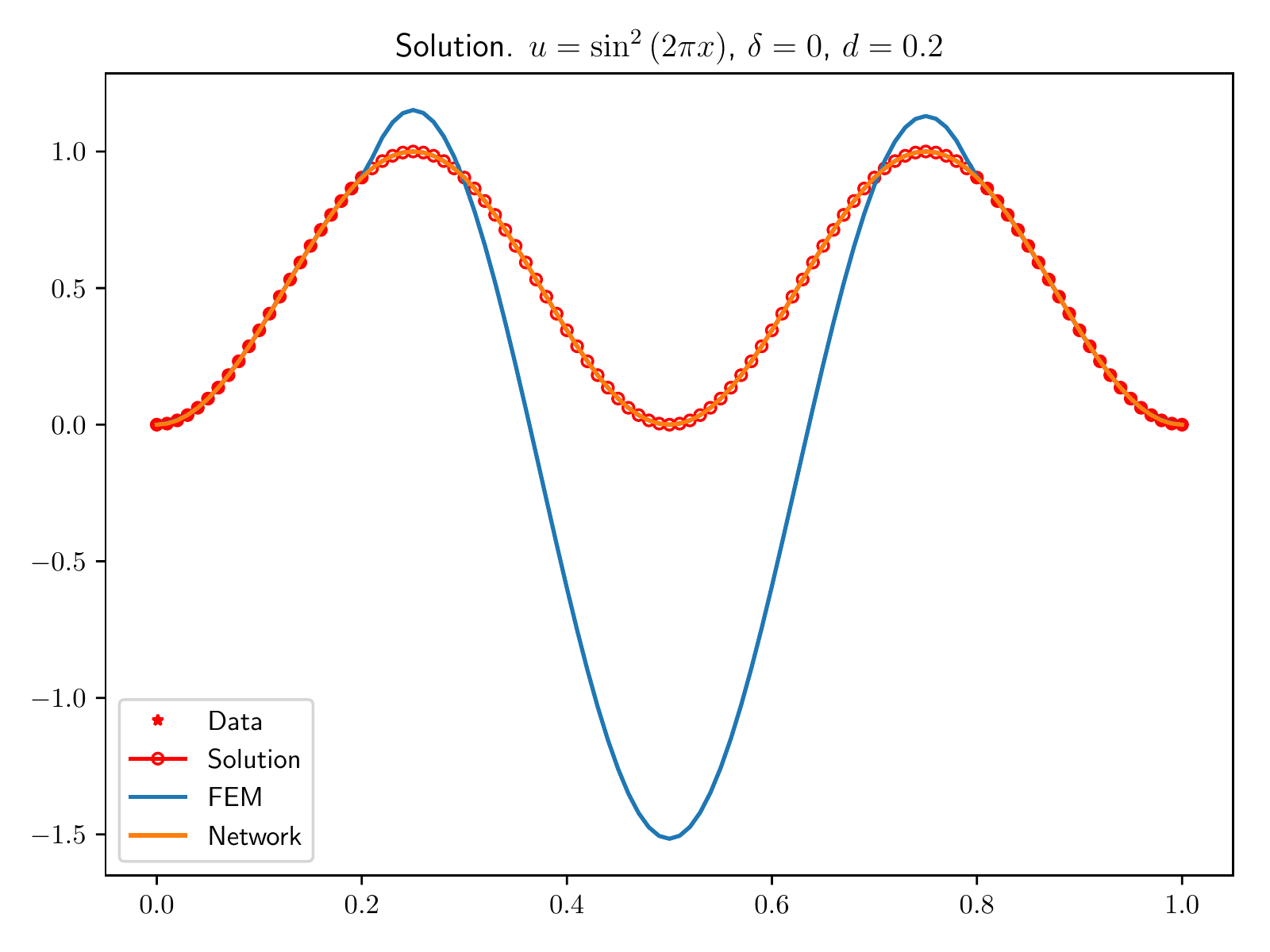}
\caption{Solutions with the optimized coefficients and no noise.}
\end{subfigure}
\begin{subfigure}[t]{0.45\textwidth}
\centering
\includegraphics[width=\textwidth]{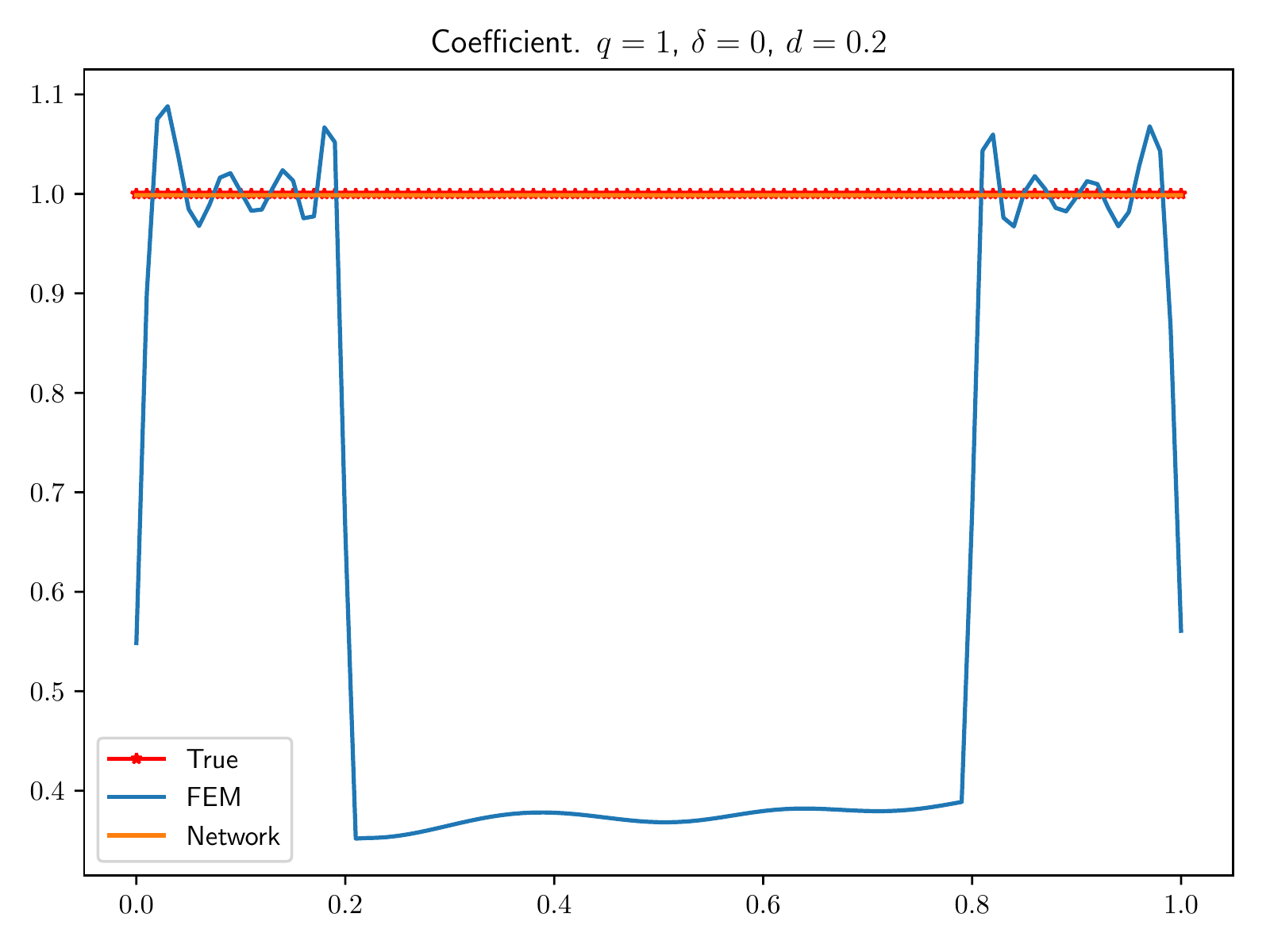}
\caption{The optimized coefficients without noise.}
\end{subfigure}
\begin{subfigure}[t]{0.45\textwidth}
\centering
\includegraphics[width=\textwidth]{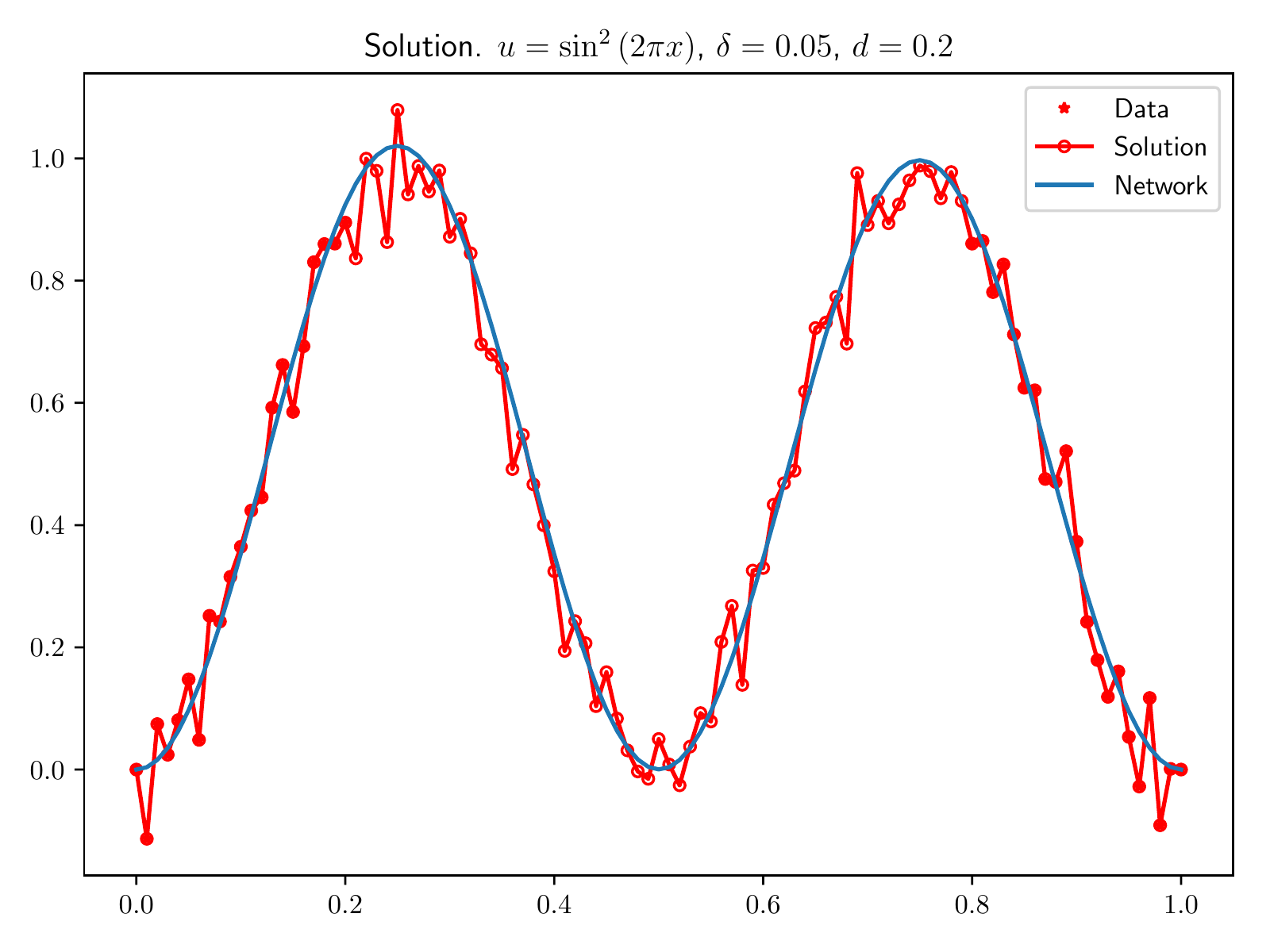}
\caption{Network solution with the optimized coefficient and 5\% noise level.}
\end{subfigure}
\begin{subfigure}[t]{0.45\textwidth}
\centering
\includegraphics[width=\textwidth]{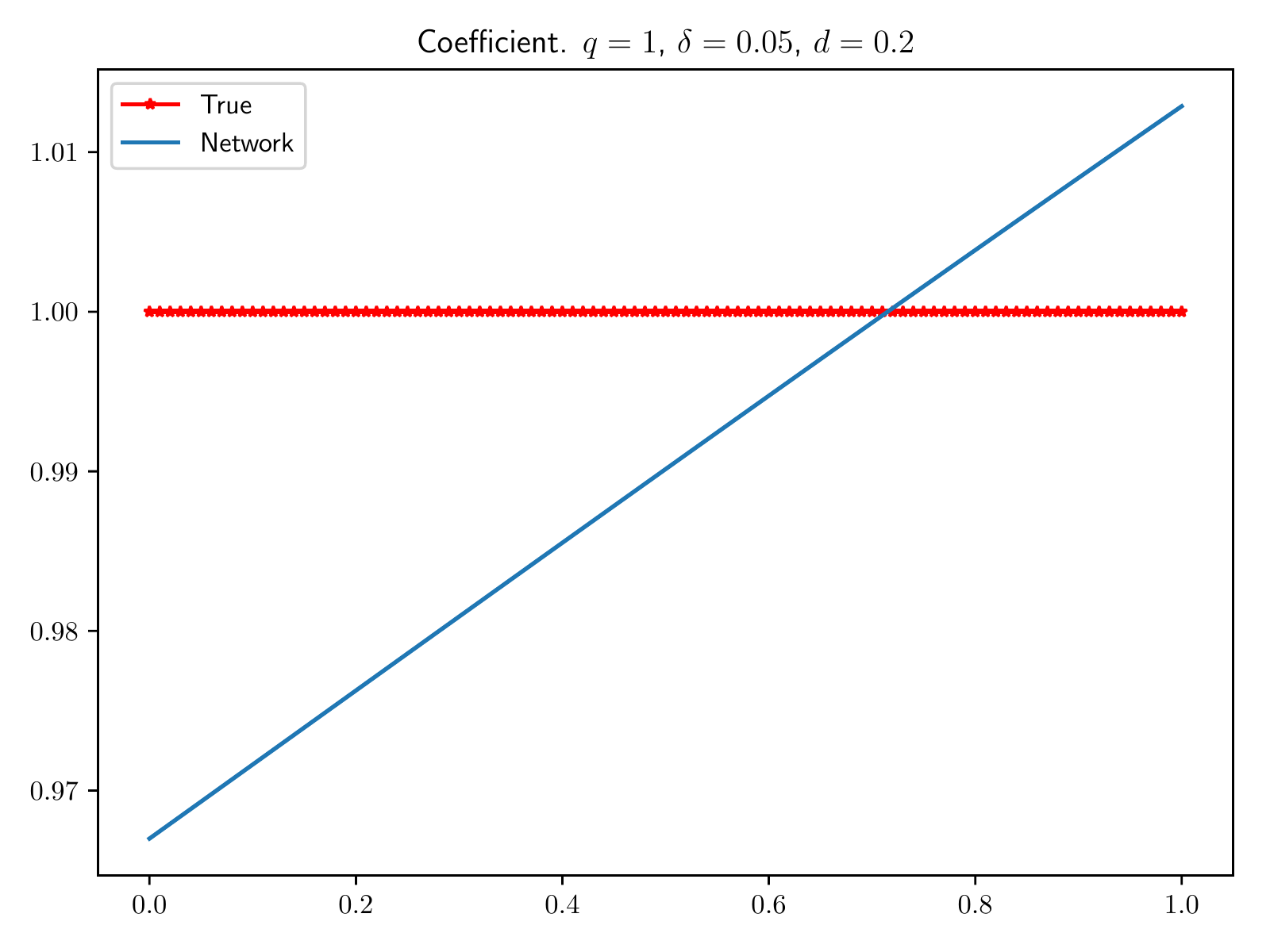}
\caption{The optimized network coefficient and 5\% noise level.}
\end{subfigure}
\begin{subfigure}[t]{0.45\textwidth}
\centering
\includegraphics[width=\textwidth]{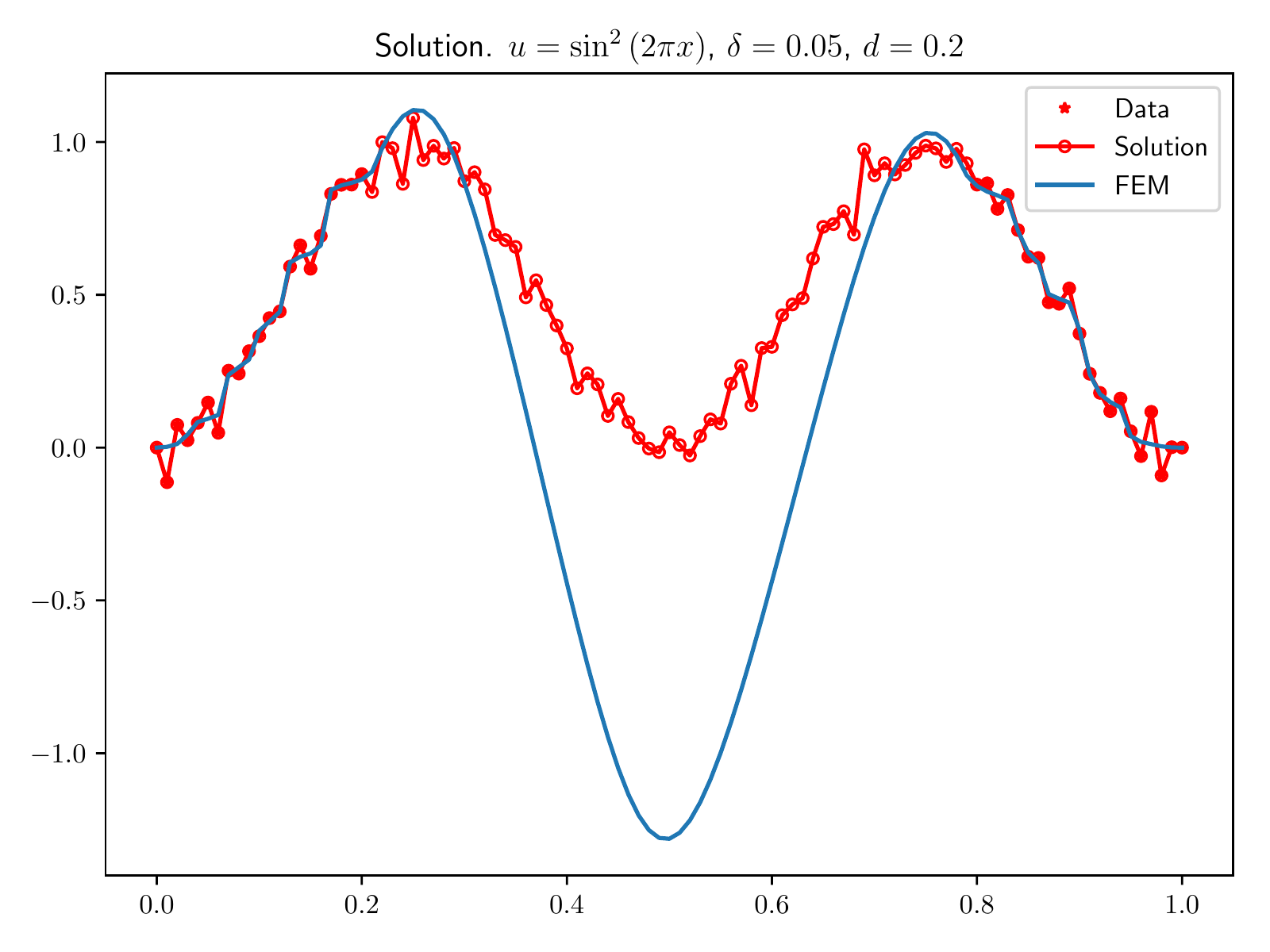}
\caption{FEM solution with the optimized coefficient and 5\% noise level.}
\end{subfigure}
\begin{subfigure}[t]{0.45\textwidth}
\centering
\includegraphics[width=\textwidth]{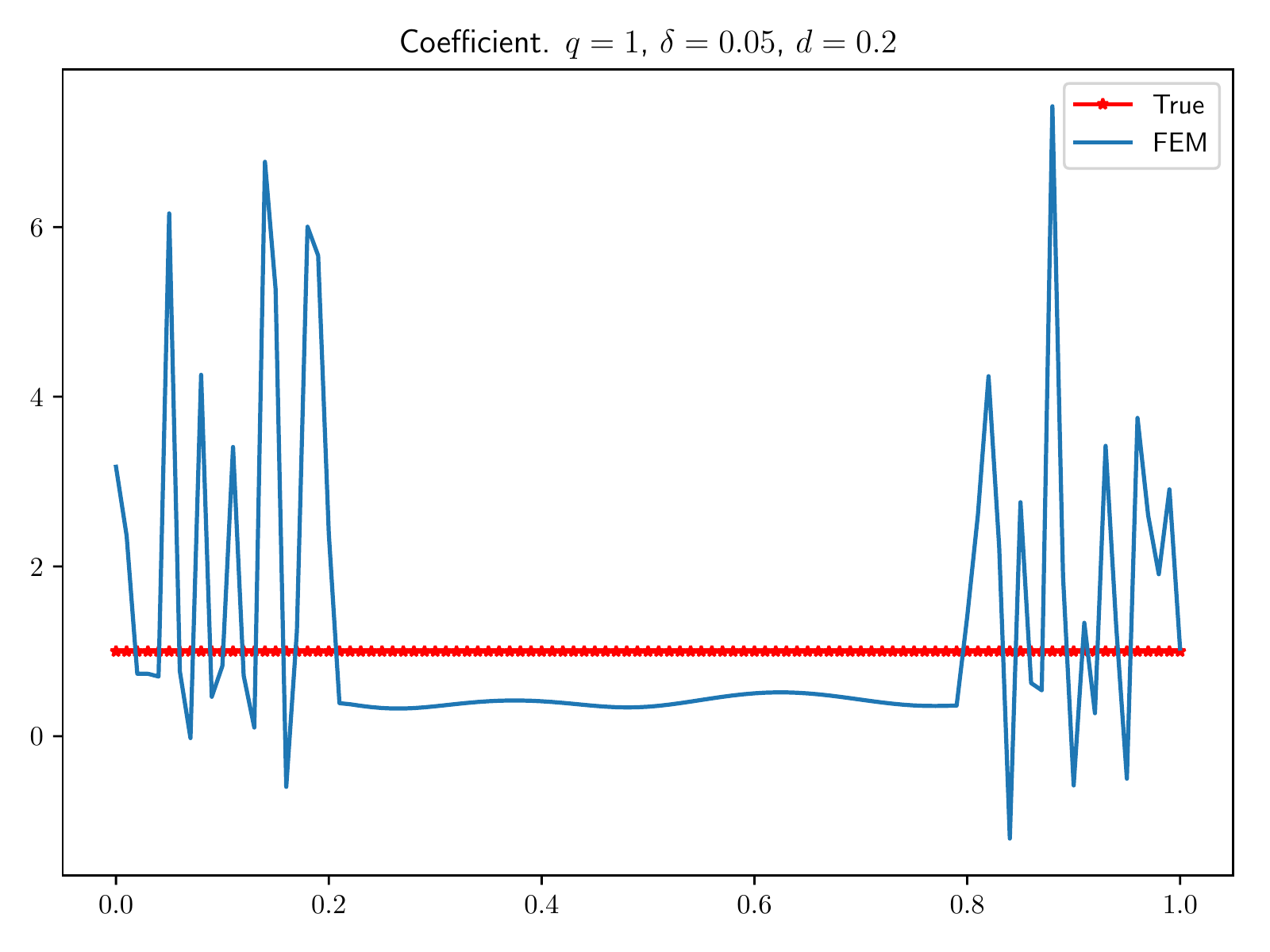}
\caption{The optimized FEM coefficient and 5\% noise level.}
\end{subfigure}
\caption{Comparison between FEM and a neural network with a constant coefficient and $d=0.2$.}
\label{heat1dqconstinc2figs}
\end{figure}

\begin{figure}[htp]
\centering
\begin{subfigure}[t]{0.45\textwidth}
\centering
\includegraphics[width=\textwidth]{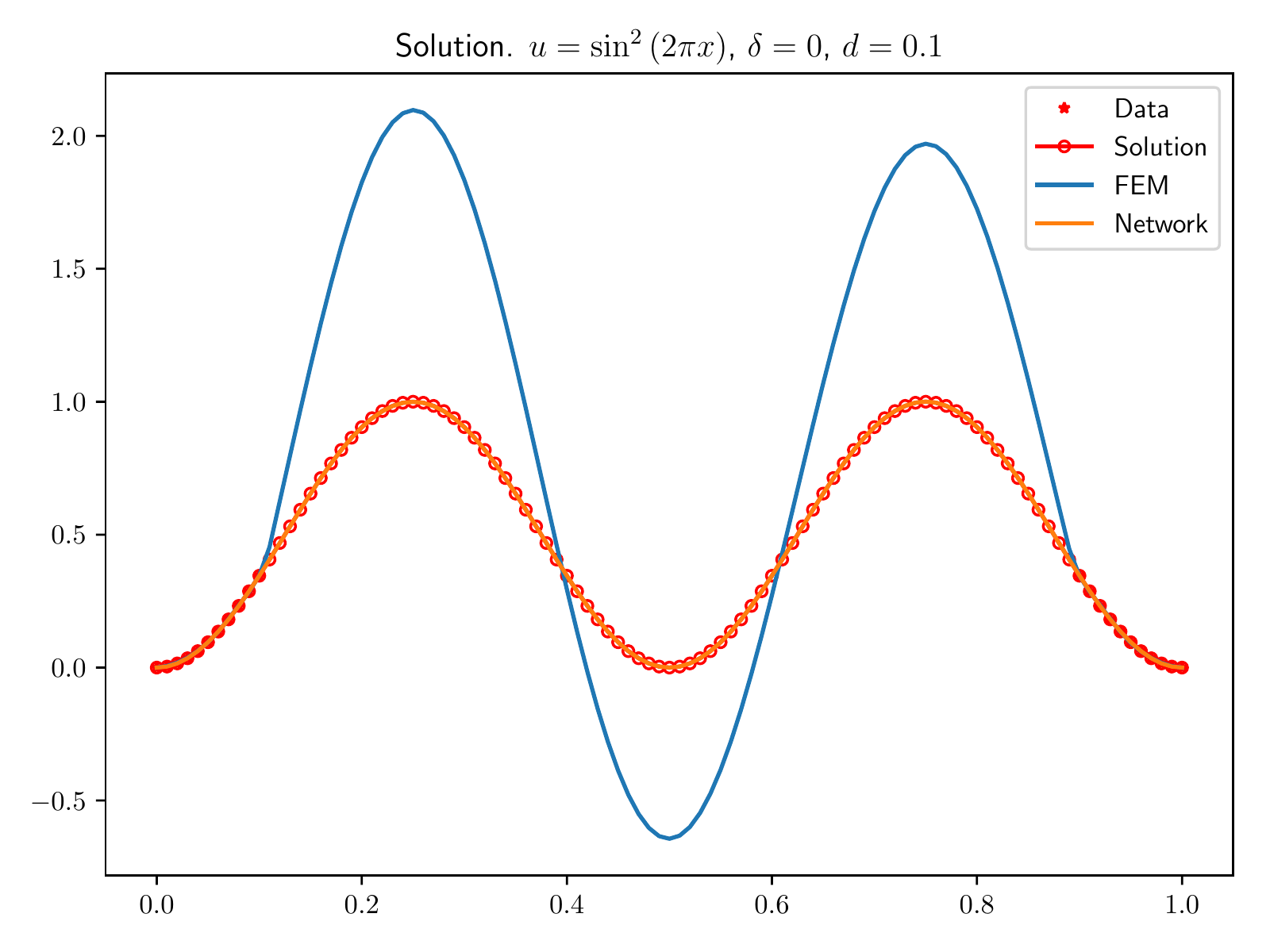}
\caption{Solutions with the optimized coefficients and no noise.}
\end{subfigure}
\begin{subfigure}[t]{0.45\textwidth}
\centering
\includegraphics[width=\textwidth]{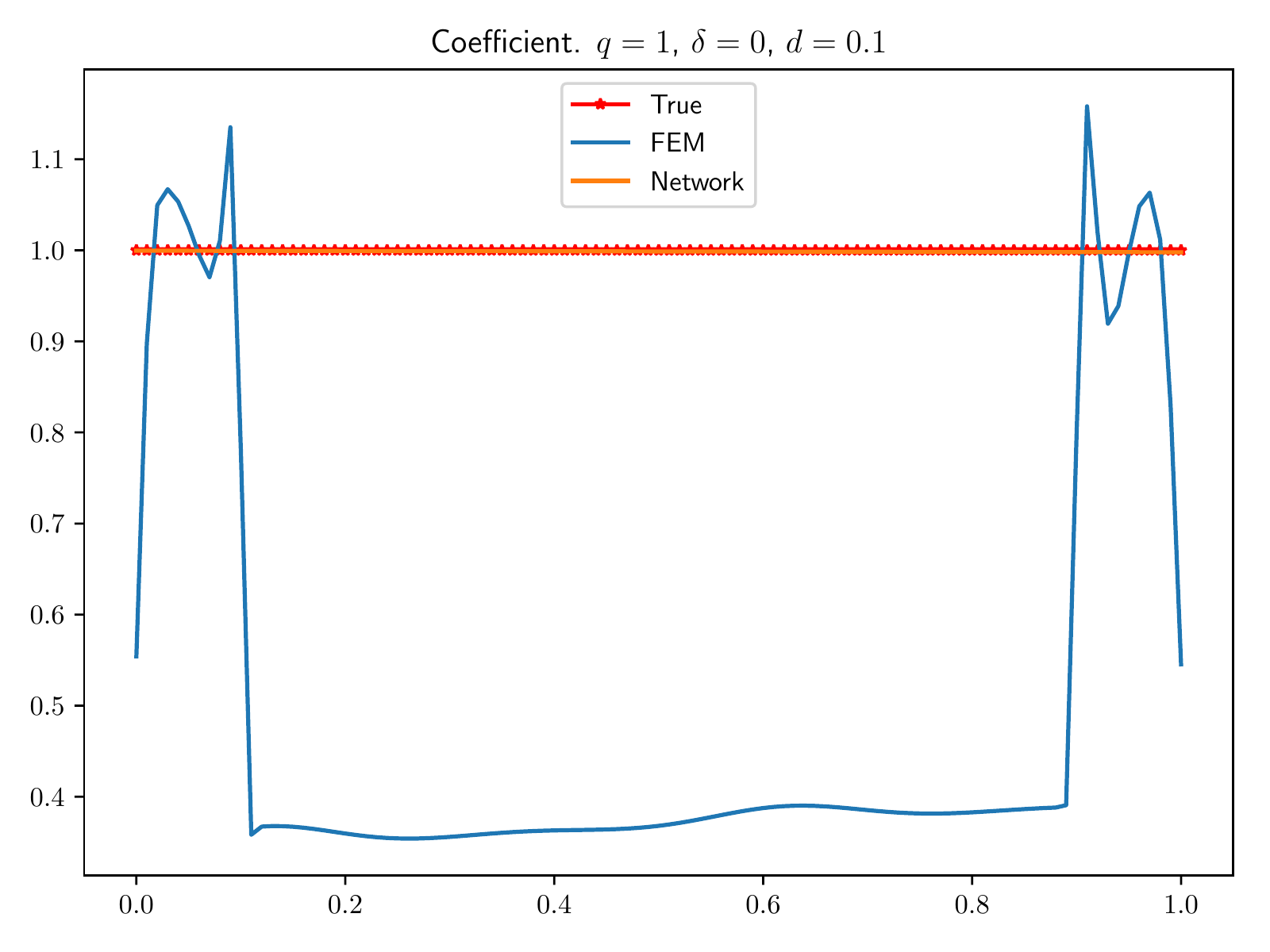}
\caption{The optimized coefficients without noise.}
\end{subfigure}
\begin{subfigure}[t]{0.45\textwidth}
\centering
\includegraphics[width=\textwidth]{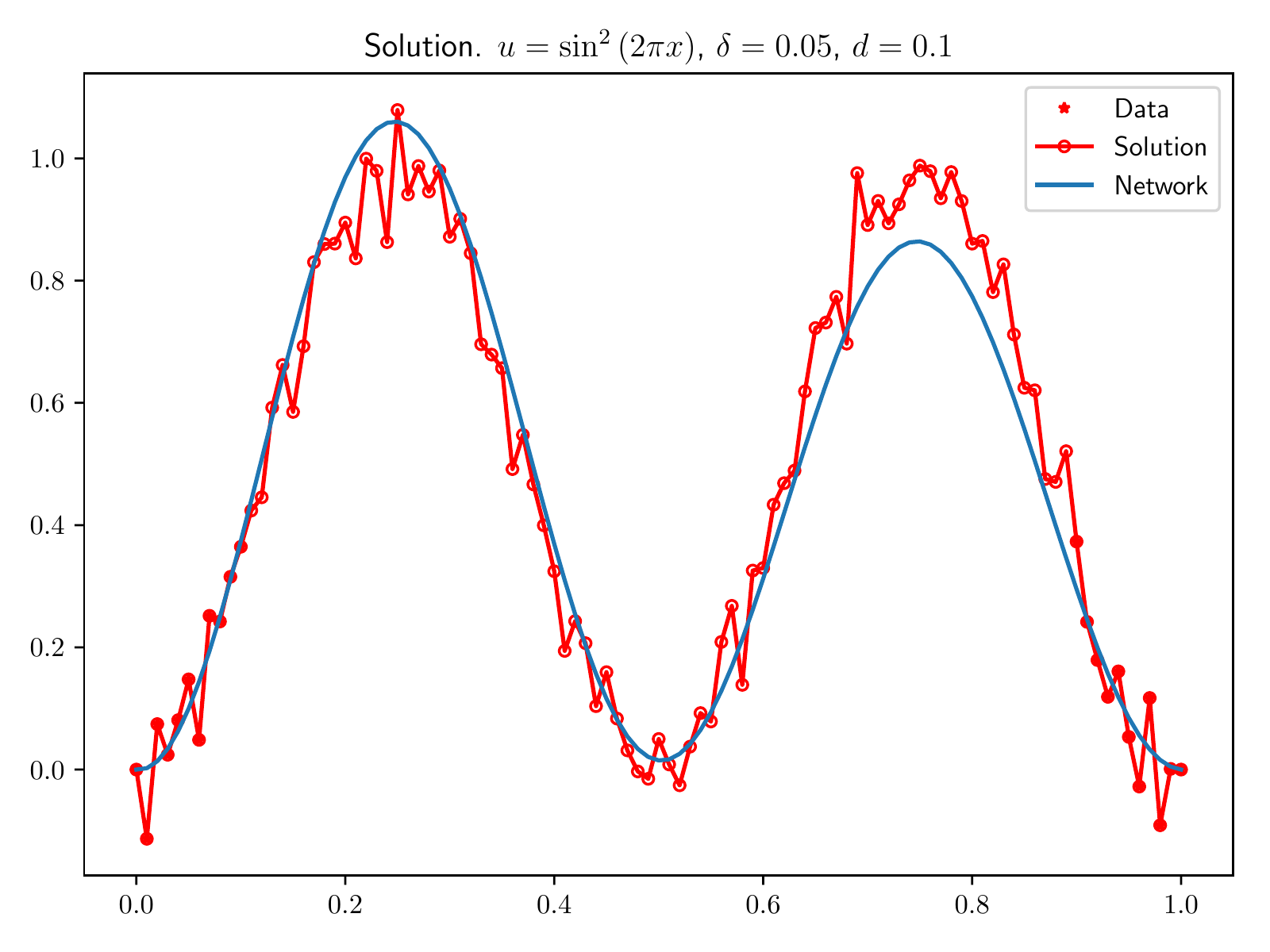}
\caption{Network solution with the optimized coefficient and 5\% noise level.}
\end{subfigure}
\begin{subfigure}[t]{0.45\textwidth}
\centering
\includegraphics[width=\textwidth]{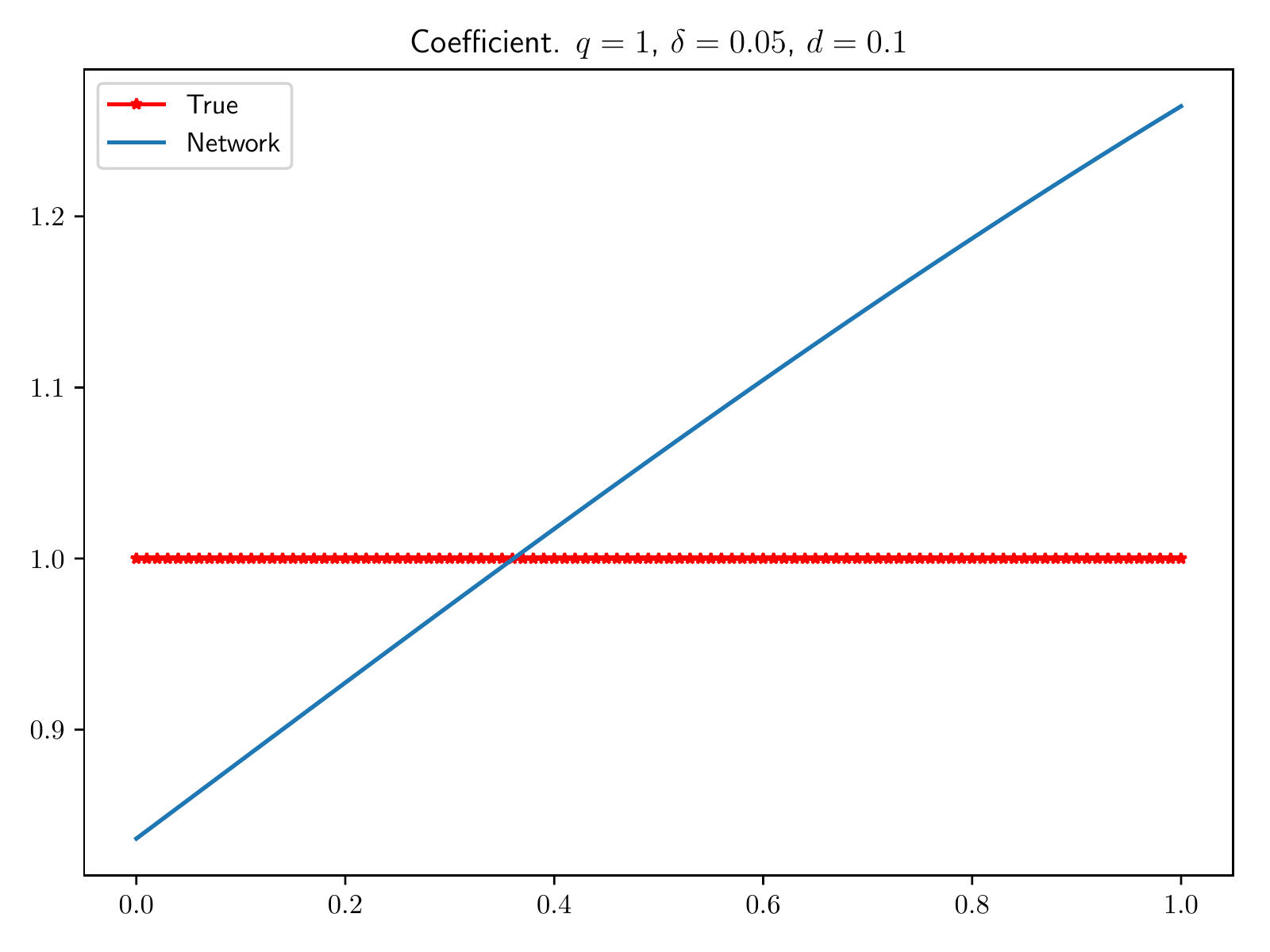}
\caption{The optimized network coefficient and 5\% noise level.}
\end{subfigure}
\begin{subfigure}[t]{0.45\textwidth}
\centering
\includegraphics[width=\textwidth]{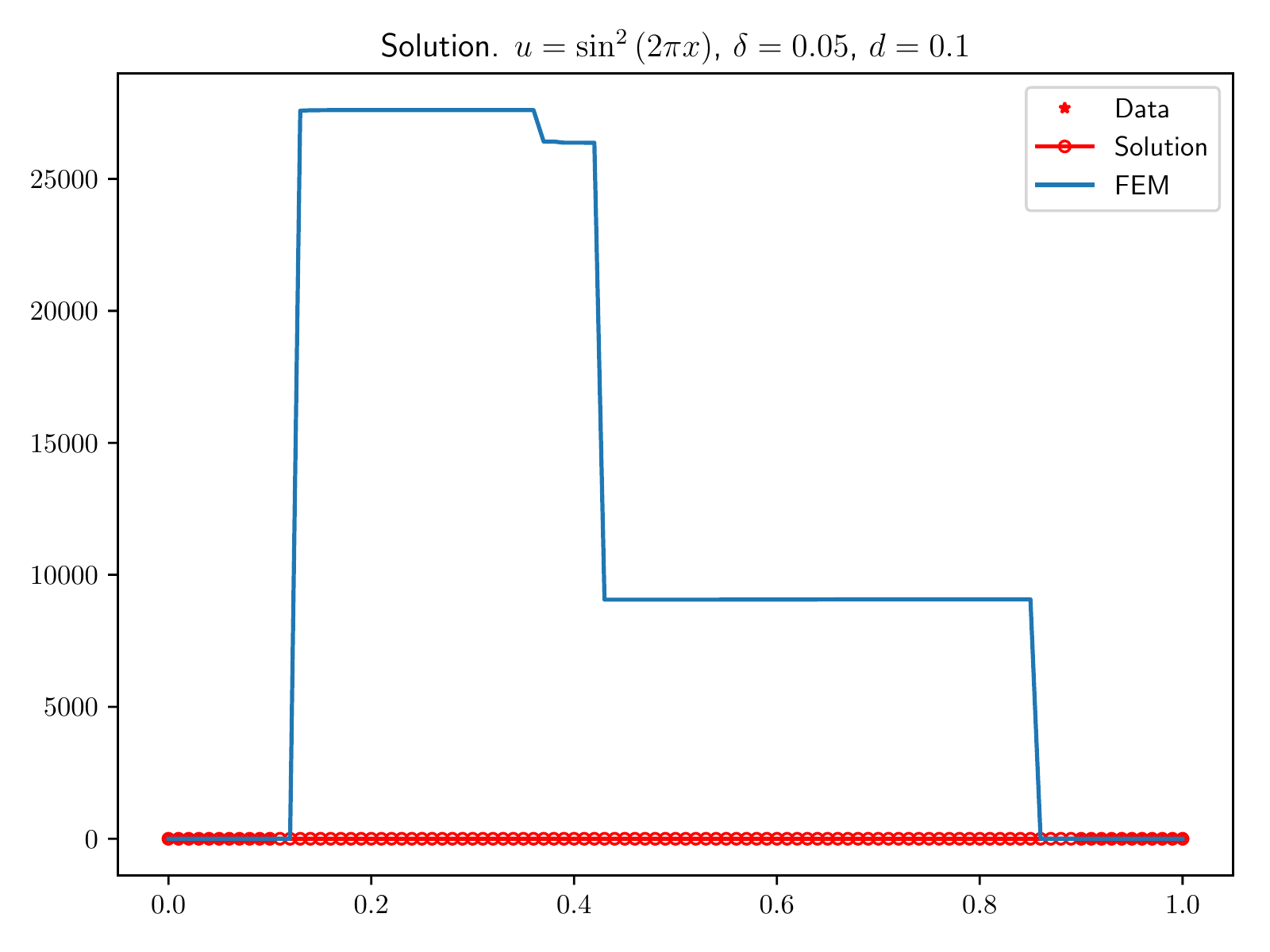}
\caption{FEM solution with the optimized coefficient and 5\% noise level.}
\end{subfigure}
\begin{subfigure}[t]{0.45\textwidth}
\centering
\includegraphics[width=\textwidth]{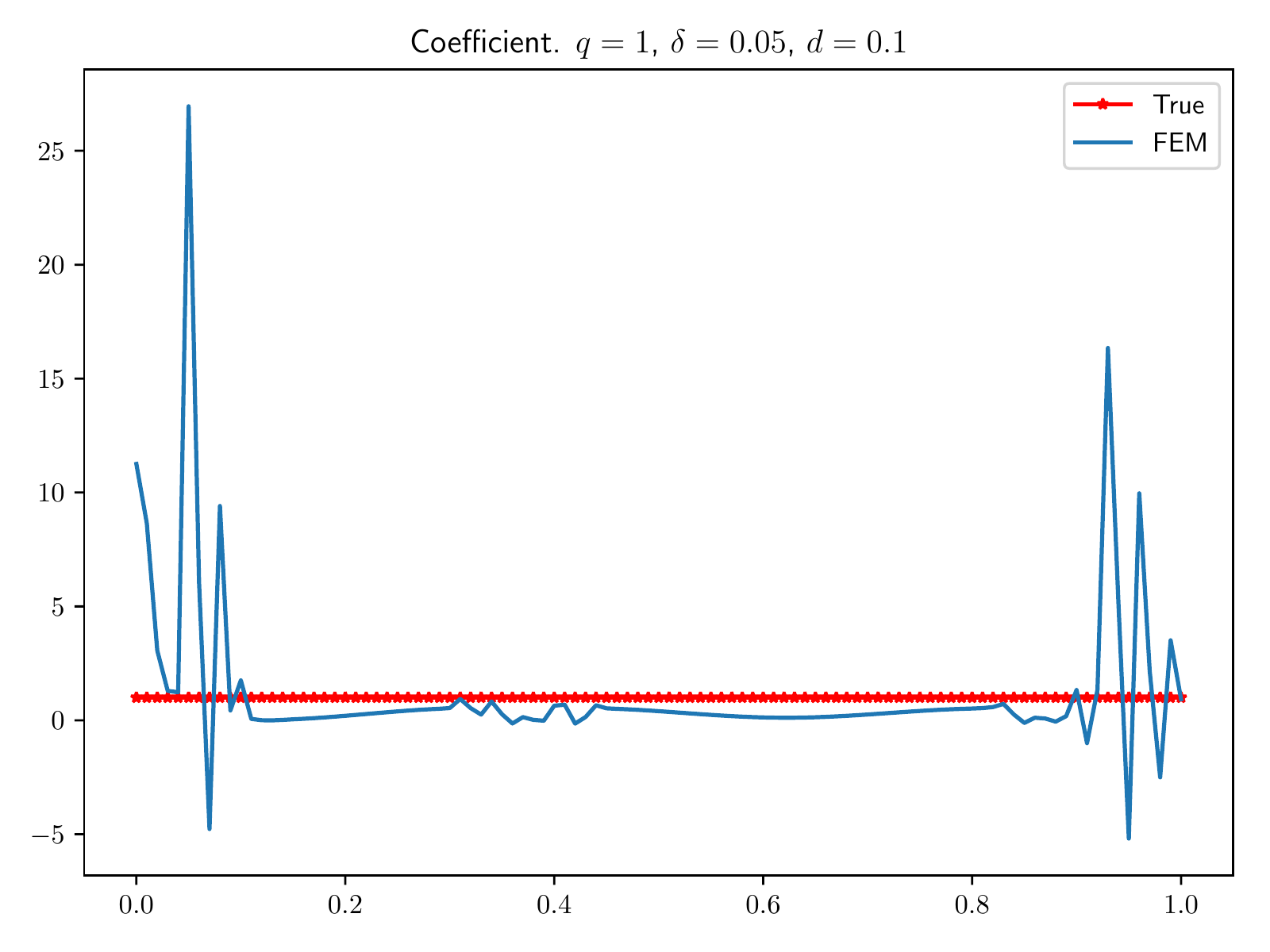}
\caption{The optimized FEM coefficient and 5\% noise level.}
\end{subfigure}
\caption{Comparison between FEM and a neural network with a constant coefficient and $d=0.1$.}
\label{heat1dqconstinc3figs}
\end{figure}

We repeat the examples in section~\ref{secheat1d} for the constant coefficient case, $q=1$, under the assumption that we only have access to measurement data at a distance $d$ from the boundaries. In this case, the error functional \eqref{errfunc} we want to minimize is reduced to
\begin{equation}
J(u, q) = \frac{1}{2} \int_0^d |u - \hat{u}|^2 dx + \frac{1}{2} \int_{1-d}^1 |u - \hat{u}|^2 dx.
\label{incompleteerrfunc}
\end{equation}
We let, as before, the neural network representing the unknown coefficient have a single hidden layer with 3 neurons and the sigmoid activation function. The result can seen in Figures~\ref{heat1dqconstinc1figs}--\ref{heat1dqconstinc3figs} for $d=0.4$, $0.2$, and $0.1$, respectively. We can clearly see that the neural network is able to reconstruct the coefficient and solution in the whole domain despite the data being known only close to the boundaries.

\subsection{Remark on multiple discontinuities and multiscale coefficients}
The neural network augmentation without explicit regularization works well when there are only a few discontinuities of the same size. As the number of discontinuities grow, or the scales of the coefficient start to differ too much, the low capacity networks are no longer sufficient to approximate the coefficient. To approximate more complicated coefficients, the number of network parameters needs to be increased. However, without explicit regularization, we have been unable to obtain good results in these cases. Multiple discontinuities, highly oscillatory and multiscale coefficients require more advanced networks and/or explicit regularization and will be the topic of future research.

\section{Summary and conclusions}
We have shown how to augment classical inverse problems with artificial neural networks. We used the finite element method to discretize and solve the forward problem, and automatic differentiation to derive and solve the adjoint problem related to the error functional. The total gradient of the error functional could then be computed exactly by combining the automatic adjoint for the finite element part, and backpropagation/automatic differentiation for the neural network part. By the chain rule, the discretization of the forward and backward problems are factored from the neural network prior in the computation of the gradient of the error functional. With the gradient of the error functional given, any gradient based optimization method can be used.

The neural network acts as a prior for the coefficient to be estimated from noisy data. As neural networks are global, smooth, universal approximators, they do not require explicit regularization of the error functional to recover both smooth solutions and coefficients. The convergence of classical optimization methods, with the neural network augmentation, is independent of both the mesh and geometry in contrast to standard finite elements. As the number of optimization parameters is independent of the mesh, large scale inverse problems can be computed with efficient optimizers, such as BFGS, as long as the number of network parameters is rather small.

The implicit regularization by low capacity network priors allows a smooth reconstruction of discontinuous coefficients as long as there are not too many discontinuities. Since neural networks are globally defined, they allow the reconstruction of the coefficient in the whole domain even when the measurement data is available only close to the boundaries.

For multiple discontinuities and multiscale coefficients, more research is needed on explicit regularization and the choice of neural network design.

\section{Acknowledgments}
We would like to thank the anonymous reviewers for their thorough reviews and constructive suggestions on how to improve the quality of the paper.

The authors were partially supported by a grant from the G{\"o}ran Gustafsson Foundation for Research in Natural Sciences and Medicine.

\appendix
\section{Optimal regularization of the 1D Poisson equation} \label{optreg}
We return to the problem in section~\ref{secheat1d} to discuss optimal regularization for a fair comparison with FEM. We are to minimize the generalized Tikhonov functional \eqref{genregfunc} which is restated here for convenience,
\begin{equation}
\tilde{J} = \frac{1}{2} \int |u - \hat{u}|^2 dx + \frac{\alpha^{\delta}}{2} \int |q - q_*|^2 dx.
\label{genregfuncapp}
\end{equation}
In this case we have a complete description of the noise which is uniformly distributed with a value of $\delta r$ added to each measurement, where $r \in \mathcal{N}(0, 1)$ and $\delta = 0.05$. Moreover, since the problem is artificially constructed we know the exact value of $q_*$ in \eqref{genregfuncapp}. This allows us to compute the optimal value of $\alpha^{\delta}$ using the Morozov discrepancy principle for the various coefficients to be estimated. The Morozov discrepancy principle amounts to finding the unique zero of the monotone function
\begin{equation}
f(\alpha^{\delta}) = \int |u - \hat{u}|^2 dx - ||\delta||
\label{morozov}
\end{equation}
where $||\delta||$ denotes the noise level. The computation of $u$ requires that the inverse coefficient problem is solved by minimizing \eqref{genregfuncapp}. Each evaluation of \eqref{morozov} thus require the solution of the forward problem and the backward problem until convergence in the coefficient $q$ for a given $\alpha$. If \eqref{morozov} is computed numerically by an iterative method, this procedure is repeated in each iteration making the computation of an optimal regularization extremely expensive. For the 1D Poisson problem, however, we can perform the computations to obtain the optimal regularization parameters for $\delta = 0.05$ as shown in Table~\ref{optregtable}.
\begin{table}[htp]
\centering
\def\arraystretch{1.2}
\begin{tabular}{|c|c|c|}
\hline
$q_*$ & $\alpha^{\delta}$ (FEM) & $\alpha^{\delta}$ (Network) \\
\hline
$1$ & 10.3 & 0.29 \\
$1 + x$ & 18.6 & 0.29 \\
$1 + x^2$ & 17.3 & 0.29 \\
$1 + 0.5\sin(2 \pi x)$ & 12.0 & 0.29 \\
\hline
\end{tabular}
\caption{Optimal regularization parameters by the Morozov discrepancy principle for 5\% noise level and known coefficients $q_*$.}
\label{optregtable}
\end{table}

The results in section~\ref{secheat1d} are repeated here where we have used optimal regularization. It can clearly be seen that FEM outperforms neural networks this time. However, since optimal regularization is rare we propose that neural networks are used to compute the estimated coefficient $q_*$ in \eqref{genregfuncapp} which can be used in a generalized Tikohonov regularization.

\begin{figure}[htp]
\centering
\begin{subfigure}[t]{0.49\textwidth}
\centering
\includegraphics[width=\textwidth]{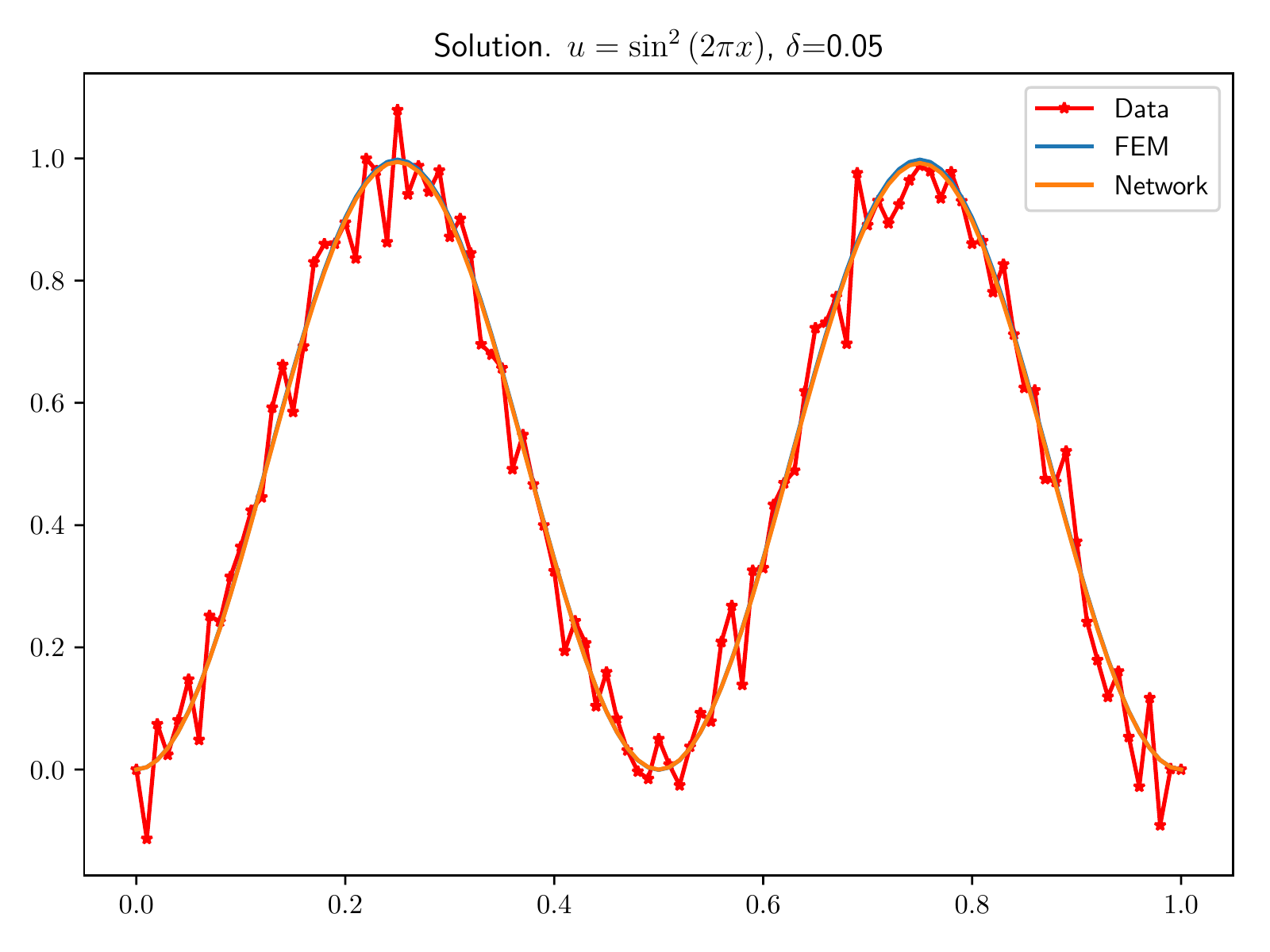}
\caption{Solutions with the optimized coefficients.}
\end{subfigure}
\begin{subfigure}[t]{0.49\textwidth}
\centering
\includegraphics[width=\textwidth]{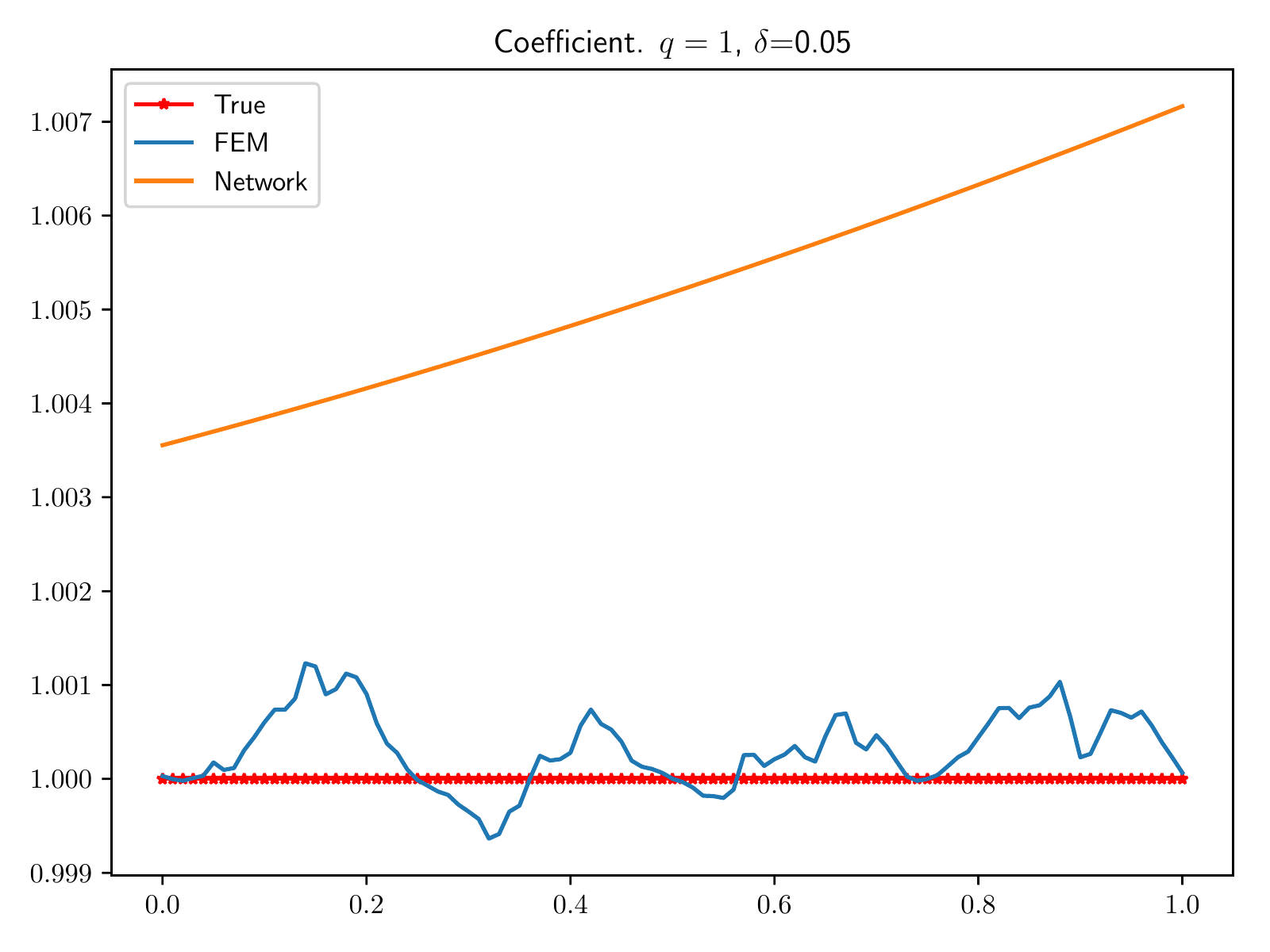}
\caption{The optimized coefficients.}
\end{subfigure}
\caption{Comparison between optimal FEM and a neural network with constant coefficient $\hat{q} = 1$ and 5\% noise level.}
\label{heat1dqconstoptimfigs}
\end{figure}
\begin{figure}[htp]
\centering
\begin{subfigure}[t]{0.49\textwidth}
\centering
\includegraphics[width=\textwidth]{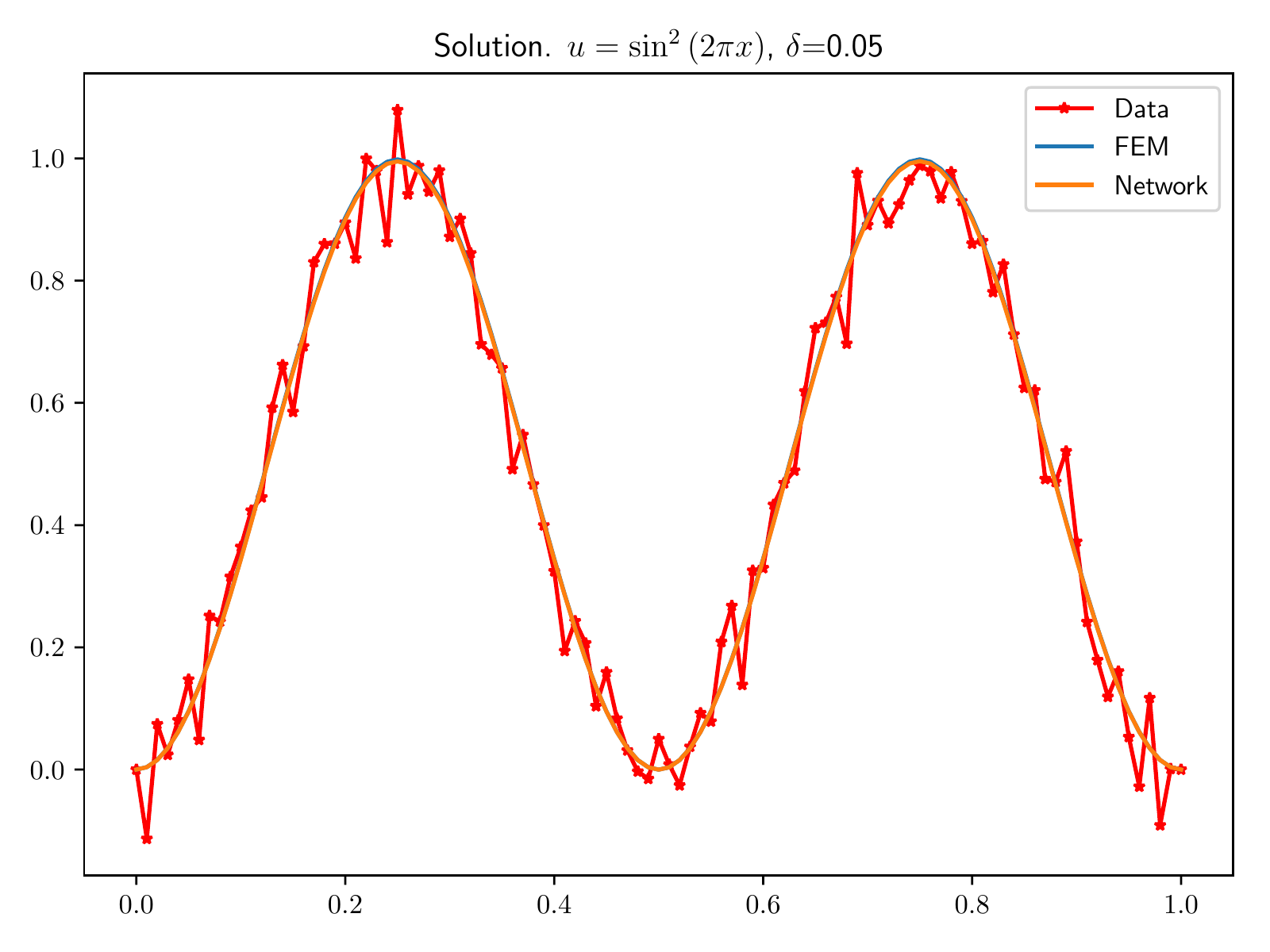}
\caption{Solutions with the optimized coefficients.}
\end{subfigure}
\begin{subfigure}[t]{0.49\textwidth}
\centering
\includegraphics[width=\textwidth]{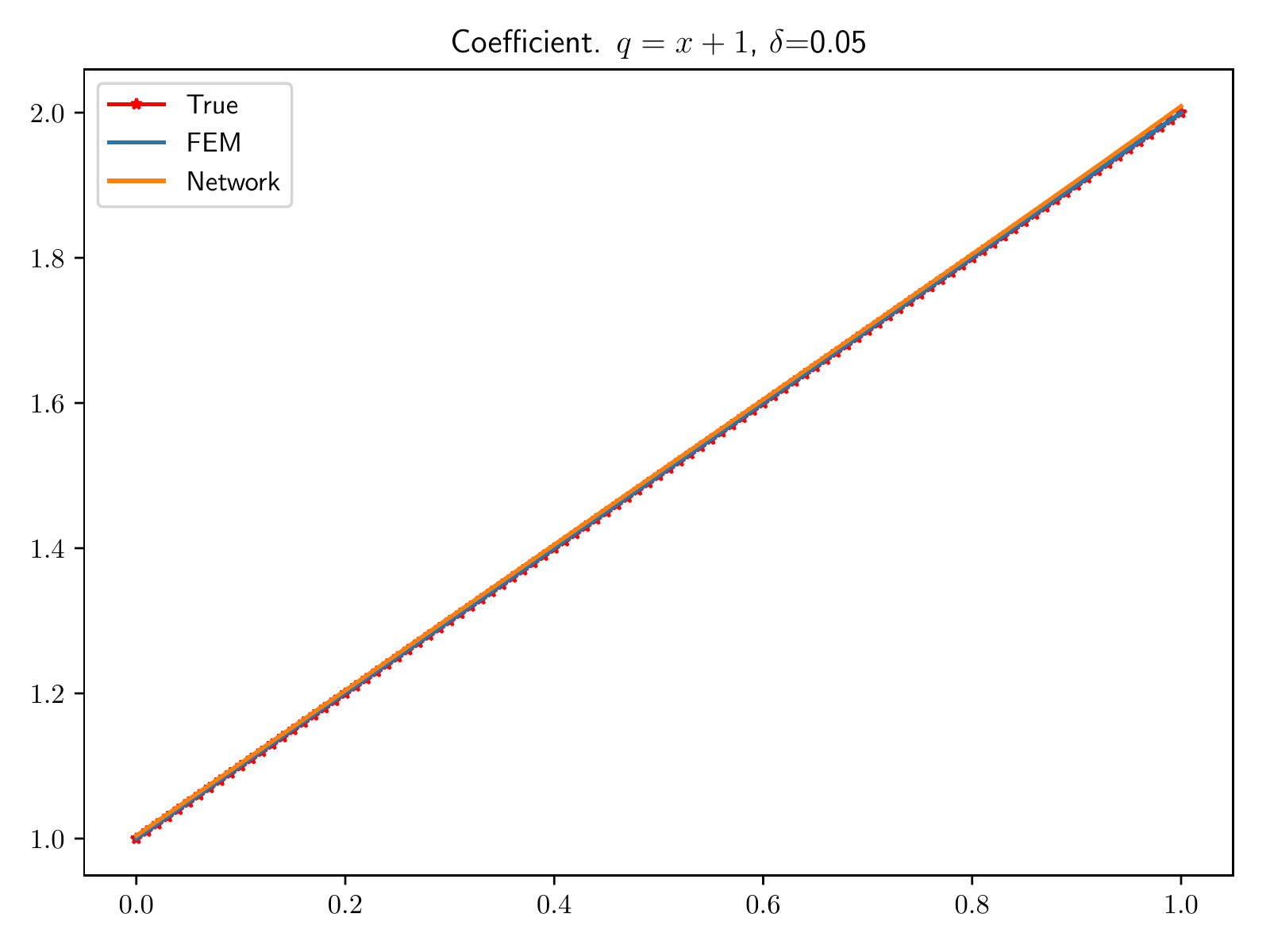}
\caption{The optimized coefficients.}
\end{subfigure}
\caption{Comparison between optimal FEM and a neural network with linear coefficient $\hat{q} = 1 + x$ and 5\% noise level.}
\label{heat1dqlineoptimfigs}
\end{figure}
\begin{figure}[htp]
\centering
\begin{subfigure}[t]{0.49\textwidth}
\centering
\includegraphics[width=\textwidth]{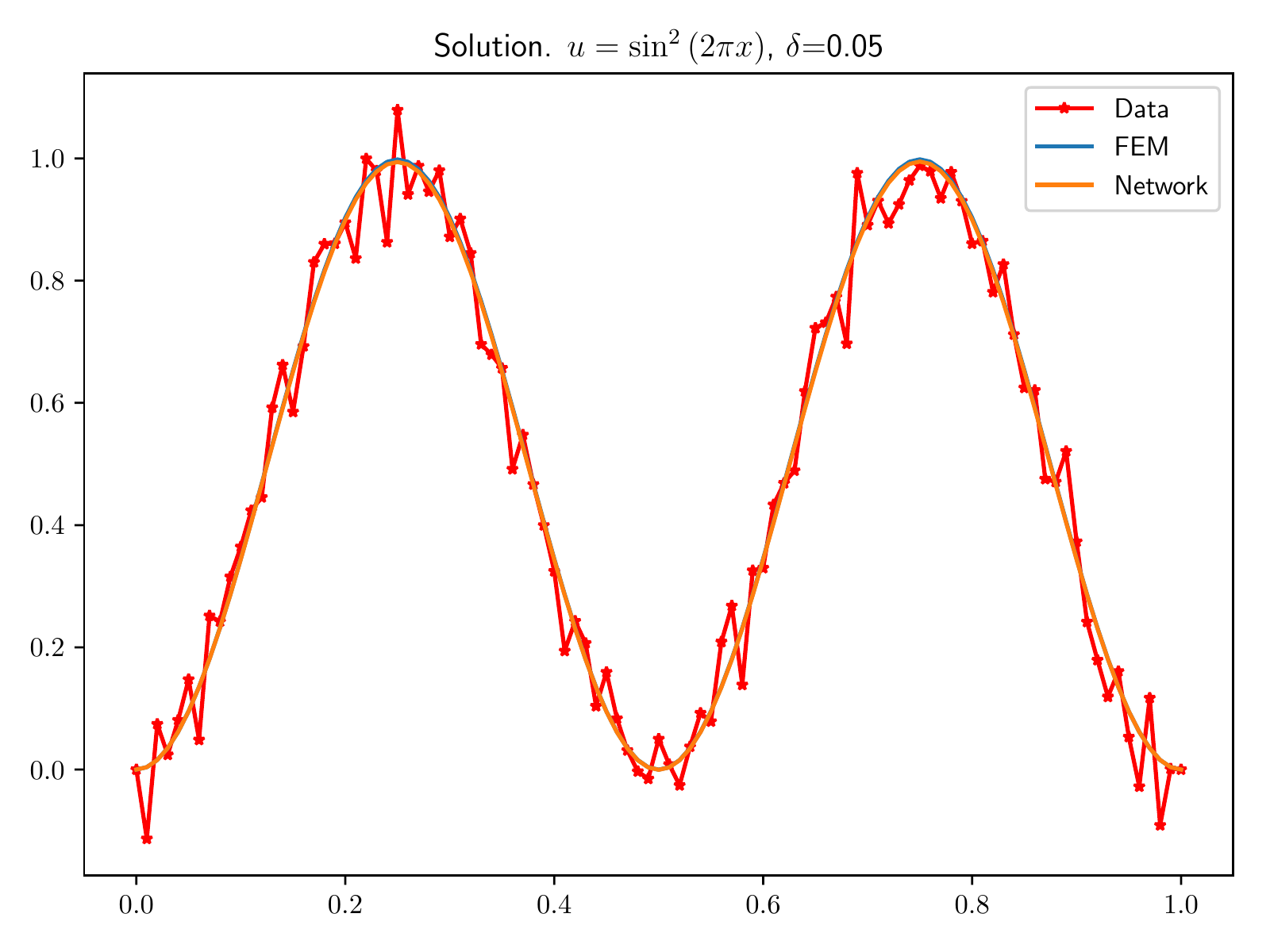}
\caption{Solutions with the optimized coefficients.}
\end{subfigure}
\begin{subfigure}[t]{0.49\textwidth}
\centering
\includegraphics[width=\textwidth]{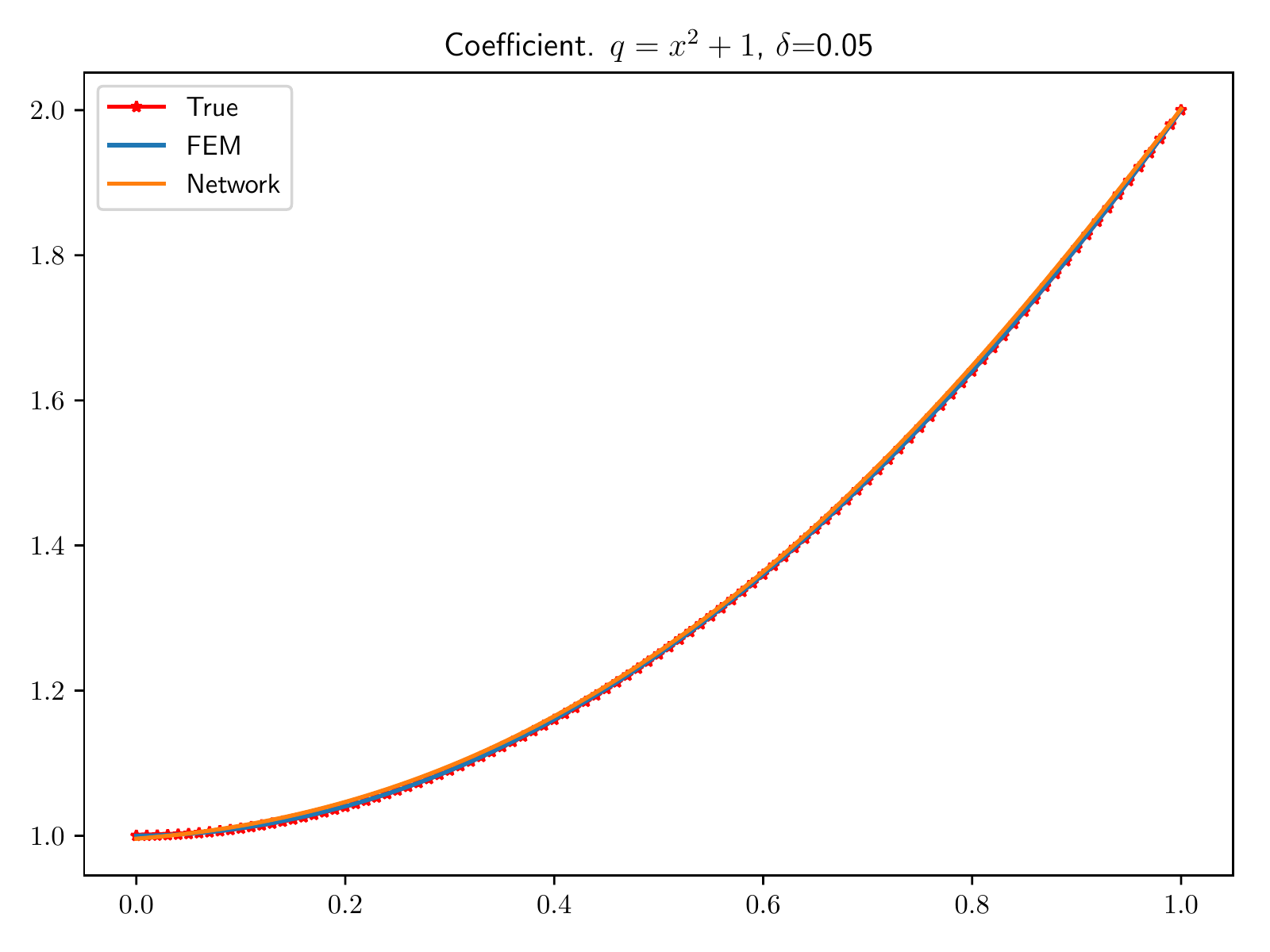}
\caption{The optimized coefficients.}
\end{subfigure}
\caption{Comparison between optimal FEM and a neural network with quadratic coefficient $\hat{q} = 1 + x^2$ and 5\% noise level.}
\label{heat1dqquadoptimfigs}
\end{figure}
\begin{figure}[htp]
\centering
\begin{subfigure}[t]{0.49\textwidth}
\centering
\includegraphics[width=\textwidth]{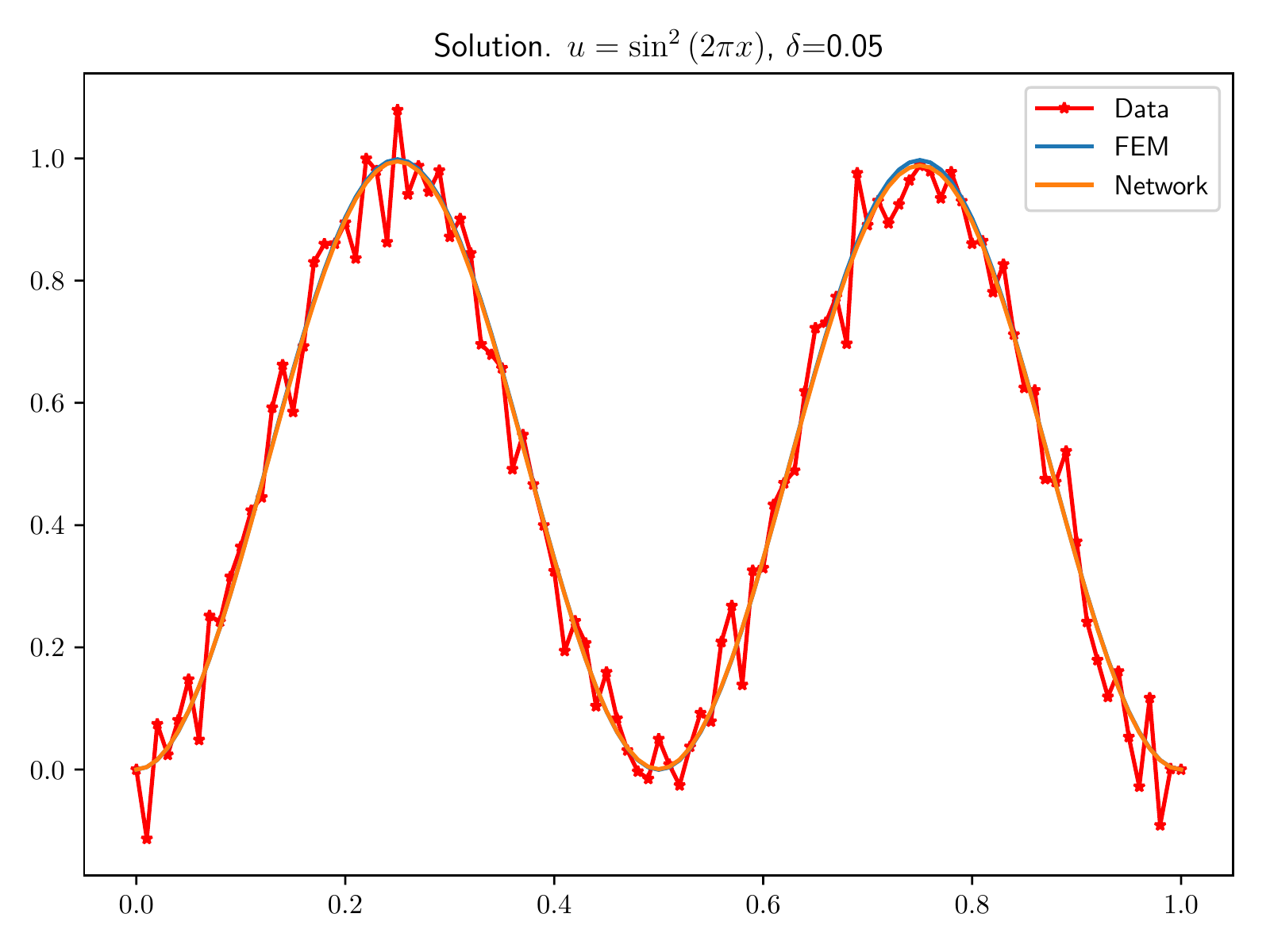}
\caption{Solutions with the optimized coefficients.}
\end{subfigure}
\begin{subfigure}[t]{0.49\textwidth}
\centering
\includegraphics[width=\textwidth]{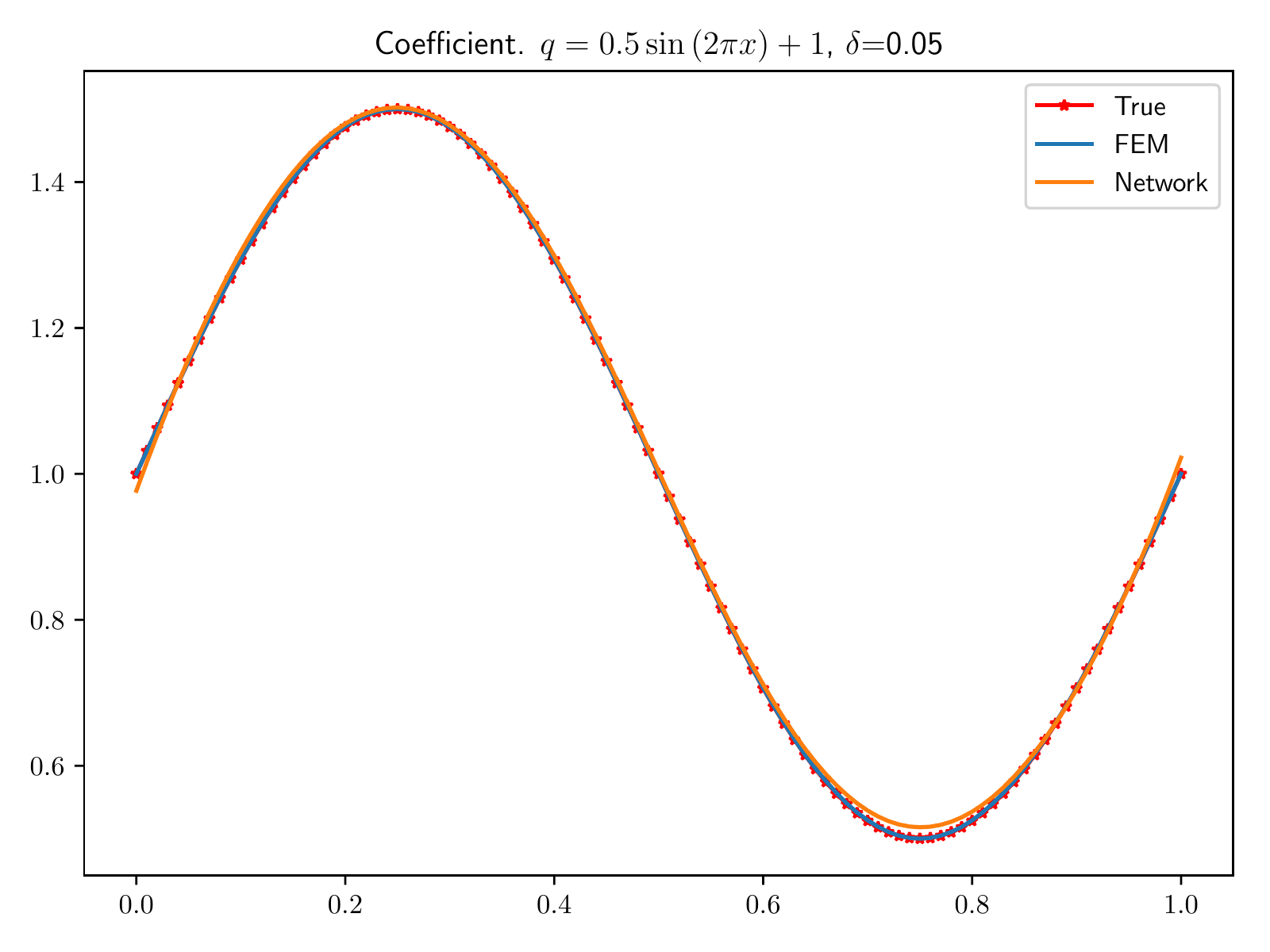}
\caption{The optimized coefficients.}
\end{subfigure}
\caption{Comparison between optimal FEM and a neural network with sine coefficient $\hat{q} = 1 + 0.5\sin(2\pi x)$ and 5\% noise level.}
\label{heat1dqsineoptimfigs}
\end{figure}

\def\arraystretch{1.2}
\begin{table}[htp]
\centering
\begin{tabular}{|c|c|c|c|c|c|c|c|c|}
\hline
& \multicolumn{4}{|c|}{$q=1$} \\
\hline
& \#it & Time (s) & $||u - \hat{u}||$ & $||q-\hat{q}||$ \\
\hline
Network & 12 & 1 & 4.06e-03 & 5.34e-03 \\
FEM & 40 & 4 & 1.26e-03 & 5.00e-04 \\
\hline
\hline
& \multicolumn{4}{|c|}{$q=x + 1$} \\
\hline
& \#it & Time (s) & $||u - \hat{u}||$ & $||q-\hat{q}||$ \\
\hline
Network & 354 & 46 & 2.90e-03 & 5.03e-03 \\
FEM & 36 & 8 & 9.30e-04 & 2.06e-04 \\
\hline
\hline
& \multicolumn{4}{|c|}{$q=x^{2} + 1$} \\
\hline
& \#it & Time (s) & $||u - \hat{u}||$ & $||q-\hat{q}||$ \\
\hline
Network & 281 & 78 & 3.26e-03 & 5.43e-03 \\
FEM & 37 & 12 & 9.99e-04 & 2.44e-04 \\
\hline
\hline
& \multicolumn{4}{|c|}{$q=0.5 \sin{\left (2 \pi x \right )} + 1$} \\
\hline
& \#it & Time (s) & $||u - \hat{u}||$ & $||q-\hat{q}||$ \\
\hline
Network & 473 & 190 & 4.48e-03 & 8.09e-03 \\
FEM & 38 & 19 & 1.38e-03 & 4.82e-04 \\
\hline
\end{tabular}
\caption{Performance comparison between neural networks and FEM with optimal regularization and 5\% noise level.}
\end{table}

\bibliographystyle{abbrv}
\bibliography{citings}

\begin{thebibliography}{10}

\bibitem{AlnaesBlechta2015a}
M.~S. Aln{\ae}s, J.~Blechta, J.~Hake, A.~Johansson, B.~Kehlet, A.~Logg,
  C.~Richardson, J.~Ring, M.~E. Rognes, and G.~N. Wells.
\newblock The {FEniCS} project version 1.5.
\newblock {\em Archive of Numerical Software}, 3(100), 2015.

\bibitem{coeffinversebook}
L.~Beilina and M.~V. Klibanov.
\newblock {\em Approximate Global Convergence and Adaptivity for Coefficient
  Inverse Problems}.
\newblock Springer-Verlag New York, 2012.

\bibitem{unified}
J.~Berg and K.~Nystr{\"o}m.
\newblock A unified deep artificial neural network approach to partial
  differential equations in complex geometries.
\newblock {\em ArXiv e-prints}, Nov. 2017.

\bibitem{2014arXiv14071517B}
T.~{Bui-Thanh} and M.~{Girolami}.
\newblock Solving large-scale {PDE}-constrained {B}ayesian inverse problems
  with {R}iemann manifold {H}amiltonian {M}onte {C}arlo.
\newblock {\em arXiv pre-print 1407.1517}, July 2014.

\bibitem{lbfgsb}
R.~H. Byrd, P.~Lu, J.~Nocedal, and C.~Zhu.
\newblock A limited memory algorithm for bound constrained optimization.
\newblock {\em SIAM Journal on Scientific Computing}, 16(5):1190--1208, 1995.

\bibitem{stepguide}
G.~Chavent.
\newblock {\em Nonlinear Least Squares for Inverse Problems: Theoretical
  Foundations and Step-by-Step Guide for Applications}.
\newblock Scientific Computation. Springer Netherlands, 2010.

\bibitem{2016arXiv160507811C}
J.~{Cockayne}, C.~{Oates}, T.~{Sullivan}, and M.~{Girolami}.
\newblock Probabilistic meshless methods for partial differential equations and
  {B}ayesian inverse problems.
\newblock {\em ArXiv}, May 2016.

\bibitem{2017arXiv170203673C}
J.~{Cockayne}, C.~{Oates}, T.~{Sullivan}, and M.~{Girolami}.
\newblock {B}ayesian probabilistic numerical methods.
\newblock {\em ArXiv e-prints}, stat.ME 1702.03673, Feb. 2017.

\bibitem{dolfinadjoint}
P.~E. Farrell, D.~A. Ham, S.~W. Funke, and M.~E. Rognes.
\newblock Automated derivation of the adjoint of high-level transient finite
  element programs.
\newblock {\em SIAM Journal on Scientific Computing}, 35(4):C369--C393, 2013.

\bibitem{bfgs}
R.~Fletcher.
\newblock {\em Practical methods of optimization}.
\newblock Wiley, 2nd edition, 1987.

\bibitem{feminversetrap}
D.~N. H\'{a}o and T.~N.~T. Quyen.
\newblock Finite element methods for coefficient identification in an elliptic
  equation.
\newblock {\em Applicable Analysis}, 93(7):1533--1566, 2014.

\bibitem{StochasticNewton}
P.~Hennig.
\newblock Fast probabilistic optimization from noisy gradients.
\newblock In {\em International Conference on Machine Learning (ICML)}, 2013.

\bibitem{hennig13_quasi_newton_method}
P.~Hennig and M.~Kiefel.
\newblock Quasi-{N}ewton methods -- a new direction.
\newblock {\em Journal of Machine Learning Research}, 14:834--865, Mar. 2013.

\bibitem{HenOsbGirRSPA2015}
P.~Hennig, M.~A. Osborne, and M.~Girolami.
\newblock Probabilistic numerics and uncertainty in computations.
\newblock {\em Proceedings of the Royal Society of London A: Mathematical,
  Physical and Engineering Sciences}, 471(2179), 2015.

\bibitem{dropout1}
G.~E. {Hinton}, N.~{Srivastava}, A.~{Krizhevsky}, I.~{Sutskever}, and R.~R.
  {Salakhutdinov}.
\newblock Improving neural networks by preventing co-adaptation of feature
  detectors.
\newblock {\em ArXiv e-prints}, July 2012.

\bibitem{pdeoptbook}
M.~Hinze, R.~Pinnau, M.~Ulbrich, and S.~Ulbrich.
\newblock {\em Optimization with PDE Constraints}.
\newblock Mathematical Modelling: Theory and Applications. Springer
  Netherlands, 2009.

\bibitem{Hornik1990551}
K.~Hornik, M.~Stinchcombe, and H.~White.
\newblock Universal approximation of an unknown mapping and its derivatives
  using multilayer feedforward networks.
\newblock {\em Neural Networks}, 3(5):551--560, 1990.

\bibitem{scipy}
E.~Jones, T.~Oliphant, P.~Peterson, et~al.
\newblock {SciPy}: Open source scientific tools for {Python}.
\newblock \url{http://www.scipy.org}, 2001--.
\newblock Accessed 2017-12-12.

\bibitem{inversesurvey}
S.~I. Kabanikhin.
\newblock Definitions and examples of inverse and ill-posed problems.
\newblock {\em Journal of Inverse and Ill-posed Problems}, 16(4):317--357,
  1966.

\bibitem{drumshape}
M.~Kac.
\newblock Can one hear the shape of a drum?
\newblock {\em The American Mathematical Monthly}, 73(4):1--23, 1966.

\bibitem{feminverseperkkele}
T.~K\"{a}rkk\"{a}inen, P.~Neittaanm\"{a}ki, and A.~Niemist\"{o}.
\newblock Numerical methods for nonlinear inverse problems.
\newblock {\em Journal of Computational and Applied Mathematics},
  74(1):231--244, 1996.

\bibitem{Kohn1988}
R.~V. Kohn and B.~D. Lowe.
\newblock A variational method for parameter identification.
\newblock {\em {ESAIM}: Mathematical Modelling and Numerical Analysis},
  22(1):119--158, 1988.

\bibitem{efficientbackprop}
Y.~LeCun, L.~Bottou, G.~B. Orr, and K.~R. M{\"u}ller.
\newblock Efficient backprop.
\newblock In {\em Neural Networks: Tricks of the Trade}, pages 9--50. Springer,
  1998.

\bibitem{levenberg}
K.~Levenberg.
\newblock A method for the solution of certain non-linear problems in least
  squares.
\newblock {\em Quarterly of Applied Mathematics}, 2(2):164--168, 1944.

\bibitem{Li1996327}
X.~Li.
\newblock Simultaneous approximations of multivariate functions and their
  derivatives by neural networks with one hidden layer.
\newblock {\em Neurocomputing}, 12(4):327--343, 1996.

\bibitem{lbfgs}
D.~C. Liu and J.~Nocedal.
\newblock On the limited memory {BFGS} method for large scale optimization.
\newblock {\em Mathematical Programming}, 45(1):503--528, Aug. 1989.

\bibitem{deathstar}
A.~Logg.
\newblock Mesh generation in {FEniCS}.
\newblock
  \url{http://www.logg.org/anders/2016/06/02/mesh-generation-in-fenics}, 2016.
\newblock Accessed 2017-12-12.

\bibitem{LoggMardalEtAl2012a}
A.~Logg, K.-A. Mardal, G.~N. Wells, et~al.
\newblock {\em Automated Solution of Differential Equations by the Finite
  Element Method}.
\newblock Springer, 2012.

\bibitem{arquardt}
D.~W. Marquardt.
\newblock An algorithm for least-squares estimation of nonlinear parameters.
\newblock {\em Journal of the Society for Industrial and Applied Mathematics},
  11(2):431--441, 1963.

\bibitem{feminversemodel}
N.~Petra and G.~Stadler.
\newblock Model variational inverse problems governed by partial differential
  equations.
\newblock Technical report, The Institute for Computational Engineering and
  Sciences, The University of Texas at Austin, 2011.

\bibitem{gausspde1}
M.~Raissi, P.~Perdikaris, and G.~E. Karniadakis.
\newblock Machine learning of linear differential equations using {G}aussian
  processes.
\newblock {\em Journal of Computational Physics}, 348:683--693, 2017.

\bibitem{gausspde2}
M.~Raissi, P.~Perdikaris, and G.~E. Karniadakis.
\newblock Numerical {G}aussian processes for time-dependent and nonlinear
  partial differential equations.
\newblock {\em SIAM Journal on Scientific Computing}, 40(1):A172--A198, 2018.

\bibitem{inversemeshdep}
T.~{Schwedes}, S.~W. {Funke}, and D.~A. {Ham}.
\newblock An iteration count estimate for a mesh-dependent steepest descent
  method based on finite elements and {R}iesz inner product representation.
\newblock {\em ArXiv e-prints}, June 2016.

\bibitem{meshindependencebook}
T.~Schwedes, D.~A. Ham, S.~W. Funke, and M.~D. Piggott.
\newblock {\em Mesh dependence in PDE-constrained optimisation}, pages 53--78.
\newblock Springer International Publishing, 2017.

\bibitem{dropout2}
N.~Srivastava, G.~Hinton, A.~Krizhevsky, I.~Sutskever, and R.~Salakhutdinov.
\newblock Dropout: A simple way to prevent neural networks from overfitting.
\newblock {\em Journal of Machine Learning Research}, 15:1929--1958, 2014.

\bibitem{tikhonov}
A.~Tikhonov and V.~Arsenin.
\newblock {\em Solutions of ill-posed problems}.
\newblock Scripta series in mathematics. Winston, 1977.

\bibitem{feminverseelliptic}
J.~Zou.
\newblock Numerical methods for elliptic inverse problems.
\newblock {\em International Journal of Computer Mathematics}, 70(2):211--232,
  1998.

\end{thebibliography}

\end{document}